\documentclass[10pt,twocolumn,letterpaper,tikz]{article}

\usepackage[pagenumbers]{cvpr} % To force page numbers, e.g. for an arXiv version

\usepackage{graphicx}
\usepackage{amsmath}
\usepackage{amssymb}
\usepackage{booktabs}

\newcommand*{\myfont}{\fontfamily{LinuxLibertineT-OsF}\selectfont}

\usepackage{siunitx}
\usepackage{adjustbox}
\usepackage{tikz}
\usepackage{arydshln}
\usepackage{float}

\definecolor{cvprblue}{rgb}{0.21,0.49,0.74}
\usepackage[pagebackref,breaklinks,colorlinks,allcolors=cvprblue]{hyperref}
\usepackage{multirow}

\usepackage[vlined,linesnumbered,ruled]{algorithm2e}
\SetAlFnt{\footnotesize}
\SetAlCapFnt{\footnotesize}
\SetAlCapNameFnt{\footnotesize}
\usepackage{algorithmic}
\algsetup{linenosize=\tiny}
\usepackage[accsupp]{axessibility}

\usepackage{siunitx}
\usepackage{adjustbox}
\usepackage{tikz}
\usepackage{arydshln}
\usepackage{float}
\usepackage{graphicx}

\definecolor{gold}{rgb}{1.0, 0.84, 0.0}
\definecolor{silver}{rgb}{0.75, 0.75, 0.75}
\definecolor{bronze}{rgb}{0.8, 0.5, 0.2}

\newcommand{\tikzcircle}[2][black,fill=none]{%
    \tikz[baseline=-0.75ex]\draw[#1] (0,0) circle (#2);%
}

\newcommand{\goldmedal}{\tikzcircle[gold,fill=gold]{3pt}}
\newcommand{\silvermedal}{\tikzcircle[silver,fill=silver]{3pt}}
\newcommand{\bronzemedal}{\tikzcircle[bronze,fill=bronze]{3pt}}

\usepackage{bbm}

\usepackage{listings}
\definecolor{code_blue}{HTML}{609AD1}
\definecolor{code_dkgreen}{rgb}{0,0.6,0}
\definecolor{code_gray}{rgb}{0.5,0.5,0.5}
\definecolor{code_mauve}{rgb}{0.58,0,0.82}
\def\paperID{10641} % *** Enter the Paper ID here
\def\confName{CVPR}
\def\confYear{2025}

\title{SemanticDraw: Towards Real-Time Interactive Content Creation\\
from Image Diffusion Models}

\author{
  \href{https://jaerinlee.com}{Jaerin Lee}$^{1}$ \hskip1.6em \href{https://dqj5182.github.io}{Daniel Sungho Jung}$^{2,3*}$ \hskip1.6em \href{https://github.com/dlrkdrjs97}{Kanggeon Lee}$^{1}$ \hskip1.6em \href{https://cv.snu.ac.kr/index.php/kmlee}{Kyoung Mu Lee}$^{1,2,3}$ \\
   $^{1}$ASRI, Department of ECE, $^{2}$Interdisciplinary Program in Artificial Intelligence,  \\ 
   $^{3}$SNU-LG AI Research Center,
   Seoul National University, Korea \\
   {\tt\small \{ironjr,dqj5182,dlrkdrjs97,kyoungmu\}@snu.ac.kr} 
}

\begin{document}
\twocolumn[{
\maketitle
\begin{center}
    \vspace{-1.2em}
    \includegraphics[width=.9\linewidth]{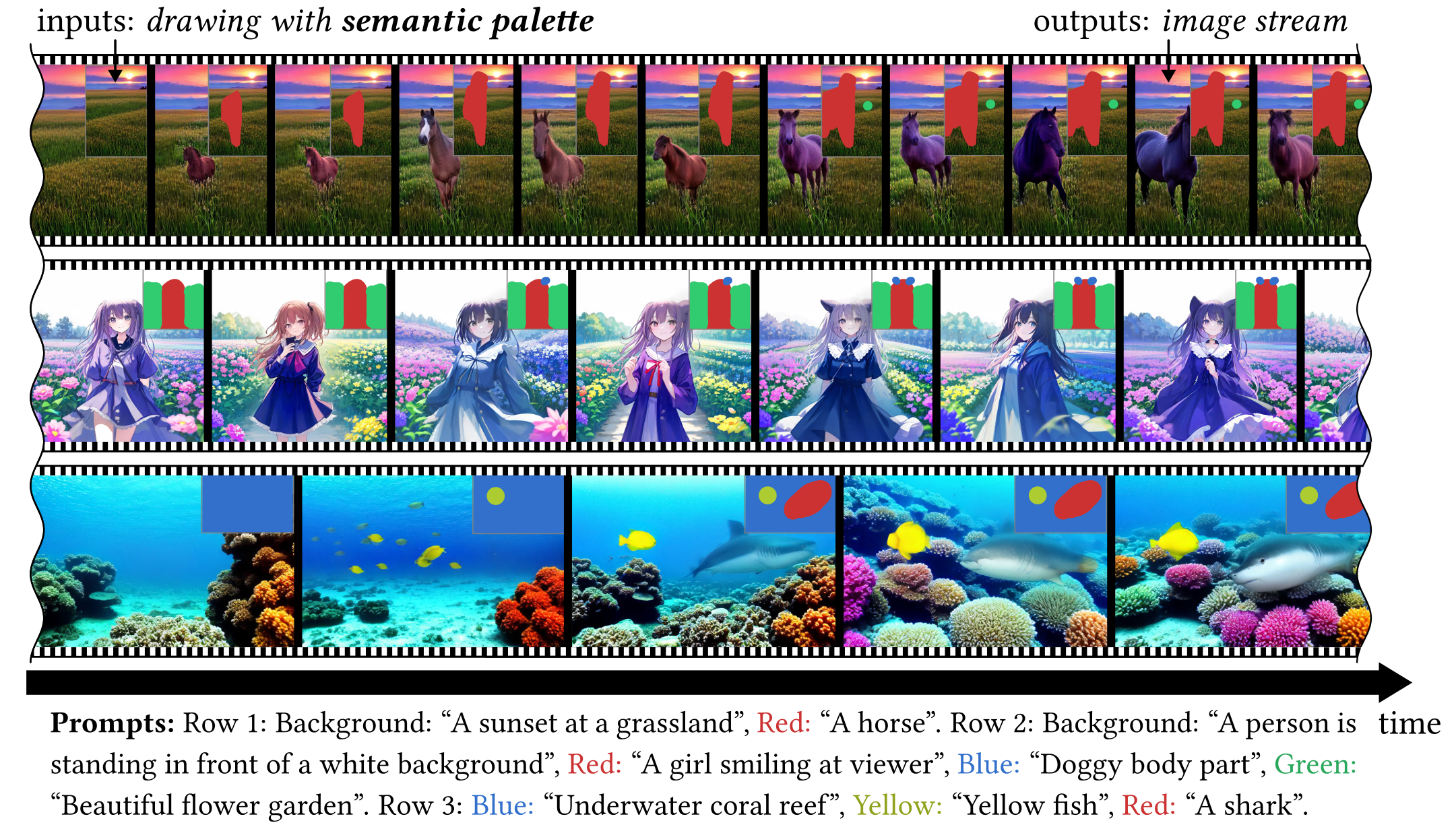}
    \vspace{-.5em}
  \captionof{figure}{%
    Overview.
    Our \textsc{SemanticDraw} is a sub-second~(0.64 seconds) solution for region-based text-to-image generation.
    This streaming architecture enables an interactive application framework, dubbed \emph{semantic palette}, where image is generated in near instant interactivity based on online user commands of hand-drawn semantic masks.
}
  \label{fig:figure_one}
    \vspace{-.2em}
\end{center}

}]

%%%%%%%%% ABSTRACT
\begin{abstract}
We introduce SemanticDraw, a new paradigm of interactive content creation where high-quality images are generated in near real-time from given multiple hand-drawn regions, each encoding prescribed semantic meaning.
In order to maximize the productivity of content creators and to fully realize their artistic imagination, it requires both quick interactive interfaces and fine-grained regional controls in their tools.
Despite astonishing generation quality from recent diffusion models, we find that existing approaches for regional controllability are very slow~(52 seconds for 512 $\times$ 512 image) while not compatible with acceleration methods such as LCM, blocking their huge potential in interactive content creation.
From this observation, we build our solution for interactive content creation in two steps:
(1) we establish compatibility between region-based controls and acceleration techniques for diffusion models, maintaining high fidelity of multi-prompt image generation with $\times 10$ reduced number of inference steps,
(2) we increase the generation throughput with our new \emph{multi-prompt stream batch} pipeline, enabling low-latency generation from multiple, region-based text prompts on a single RTX 2080 Ti GPU.
Our proposed framework is generalizable to any existing diffusion models and acceleration schedulers, allowing sub-second~(0.64 seconds) image content creation application upon well-established image diffusion models.
The code is \textnormal{\url{https://github.com/ironjr/semantic-draw}}.

\vspace{-5em}
\end{abstract}

%%%%%%%%% BODY TEXT
\section{Introduction}
\label{sec:1_intro}
Recent massive advancements and widespread adoptions of generative AI~\cite{zhang2023adding,ramesh2021zero,rombach2022high,ramesh2022hierarchical,achiam2023gpt,saharia2022photorealistic} are fundamentally transforming the landscape of content creation, demonstrating huge potential for improving efficiency of production processes and expanding the boundaries of creativity.
Especially, diffusion models~\cite{rombach2022high} are gaining significant attention in generative AI for image content creation because of their ability to produce realistic, high-resolution images.
Nevertheless, in the perspective of content creators, a pure generative quality is not the only point of consideration~\cite{mitra2024dmvcc}.
Diffusion models for content creators should require efficient, interactive tools that can swiftly translate their artistic imaginations into refined outputs, supporting a more responsive and iterative creative process with fine-grained controllability under straightforward control panels as illustrated in Figure~\ref{fig:figure_one} and~\ref{fig:screenshot}.
These goals should all be satisfied simultaneously.

The academic community had several attempts to address these criteria in isolated areas, but has yet to tackle them comprehensively.
On one hand, there is a line of works dealing with acceleration of the inference speed~\cite{song2020denoising, song2023consistency,luo2023latent,luo2023lcm, lin2024sdxl, ren2024hyper, chadebec2024flash} of diffusion models.
Acceleration schedulers including DDIM~\cite{song2020denoising}, latent consistency models~(LCM)~\cite{song2023consistency,luo2023latent,luo2023lcm}, SDXL-Lightning~\cite{lin2024sdxl}, Hyper-SD~\cite{ren2024hyper}, and Flash Diffusion~\cite{chadebec2024flash} reduced the number of required inference steps from several thousand to a few tens and then down to 4.
Focusing on the throughput directly, StreamDiffusion~\cite{kodaira2023streamdiffusion} reformed diffusion models into a pipelined architecture, enabling streamed generation and real-time video styling.
On the other hand, methods to enhance the controllability~\cite{zhang2023adding, ye2023ip, avrahami2023spatext, bar2023multidiffusion} of the generative framework were also heavily sought.
ControlNet~\cite{zhang2023adding} and IP-Adapter~\cite{ye2023ip} enabled image-based conditioning of the pre-trained diffusion models.
SpaText~\cite{avrahami2023spatext} and MultiDiffusion~\cite{bar2023multidiffusion} achieved image generation from multiple region-based texts, allowing more fine-grained controls over the generation process from localized text prompts.
Those two areas of research have developed largely independently.
This suggests a straightforward approach for fast yet controllable generation: simply combining achievements from both, \textit{e.g.}, acceleration technique such as LCM~\cite{luo2023lcm} can serve a pair of a noise schedule sequence and fine-tuned model weights.

\begin{figure}[t]
\newcommand{\figurewidth}{.96\linewidth}
\newcommand{\h}{.384\linewidth}
\newcommand{\hh}{2.5mm}
\newcommand{\vv}{\vspace*{-0.20mm}}
\definecolor{p1color}{HTML}{16C232}
\definecolor{p2color}{HTML}{F92F6C}
\definecolor{p3color}{HTML}{92C6EC}
\definecolor{p4color}{HTML}{FECAC0}
\definecolor{p5color}{HTML}{AC6AEB}
\definecolor{p6color}{HTML}{2692F3}
\definecolor{p7color}{HTML}{F89E12}
\definecolor{p8color}{HTML}{92C62C}
  \centering
{\myfont
    \makebox[\hh]{\rotatebox[origin=l]{90}{\makebox[\h][c]{\hspace{-0.\linewidth}\footnotesize{{MultiDiffusion~\cite{bar2023multidiffusion} ({\color{p2color} \textbf{51m 39s}})}}}}}\hspace{0.5mm}%
    \includegraphics[height=\h,width=\figurewidth]{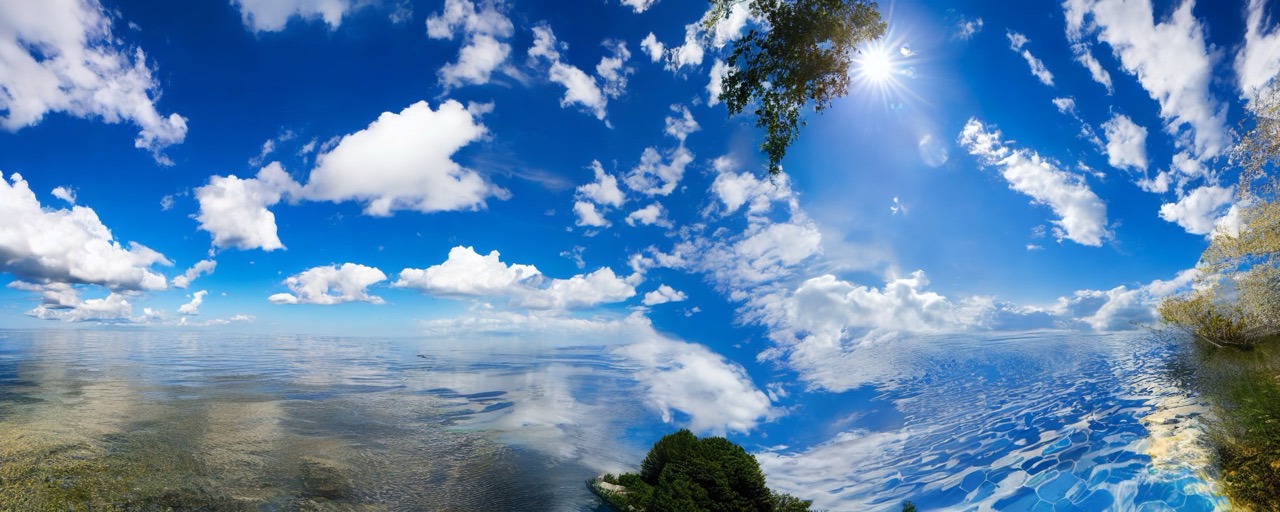}\vv\\
    \makebox[\hh]{\rotatebox[origin=l]{90}{\makebox[\h][c]{\hspace{-0.\linewidth}\footnotesize{MD~\cite{bar2023multidiffusion}+LCM~\cite{podellsdxl} ({\color{p2color} \textbf{4m 47s}})}}}}\hspace{0.5mm}%
    \includegraphics[height=\h,width=\figurewidth]{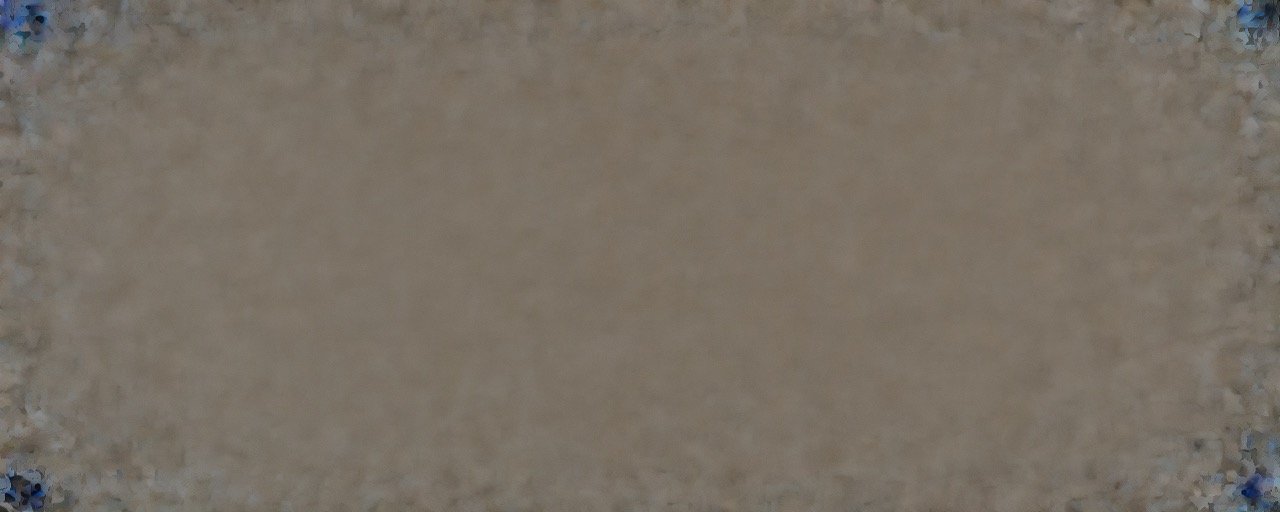}\\
    \makebox[\hh]{\rotatebox[origin=l]{90}{\makebox[\h][c]{\hspace{-0.\linewidth}\footnotesize{\textbf{Ours ({\color{p1color}59s})}}}}}\hspace{0.5mm}%
    \includegraphics[height=\h,width=\figurewidth]{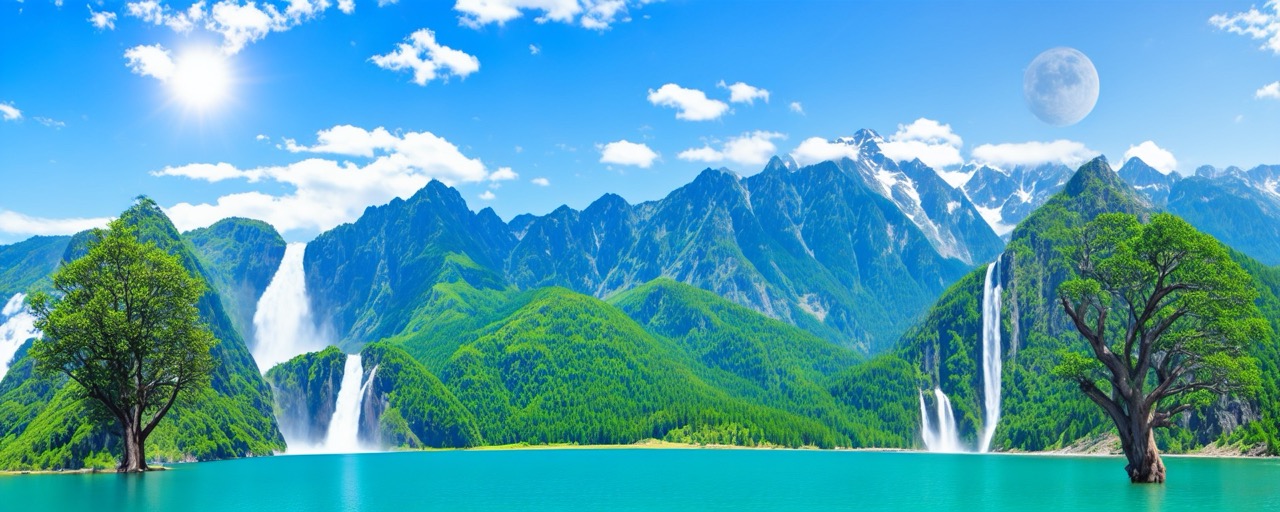}\vv\\
    \makebox[\hh]{\rotatebox[origin=l]{90}{\makebox[\h][c]{\hspace{-0.\linewidth}\footnotesize{Prompt and Mask}}}}\hspace{0.5mm}%
    \includegraphics[width=\figurewidth]{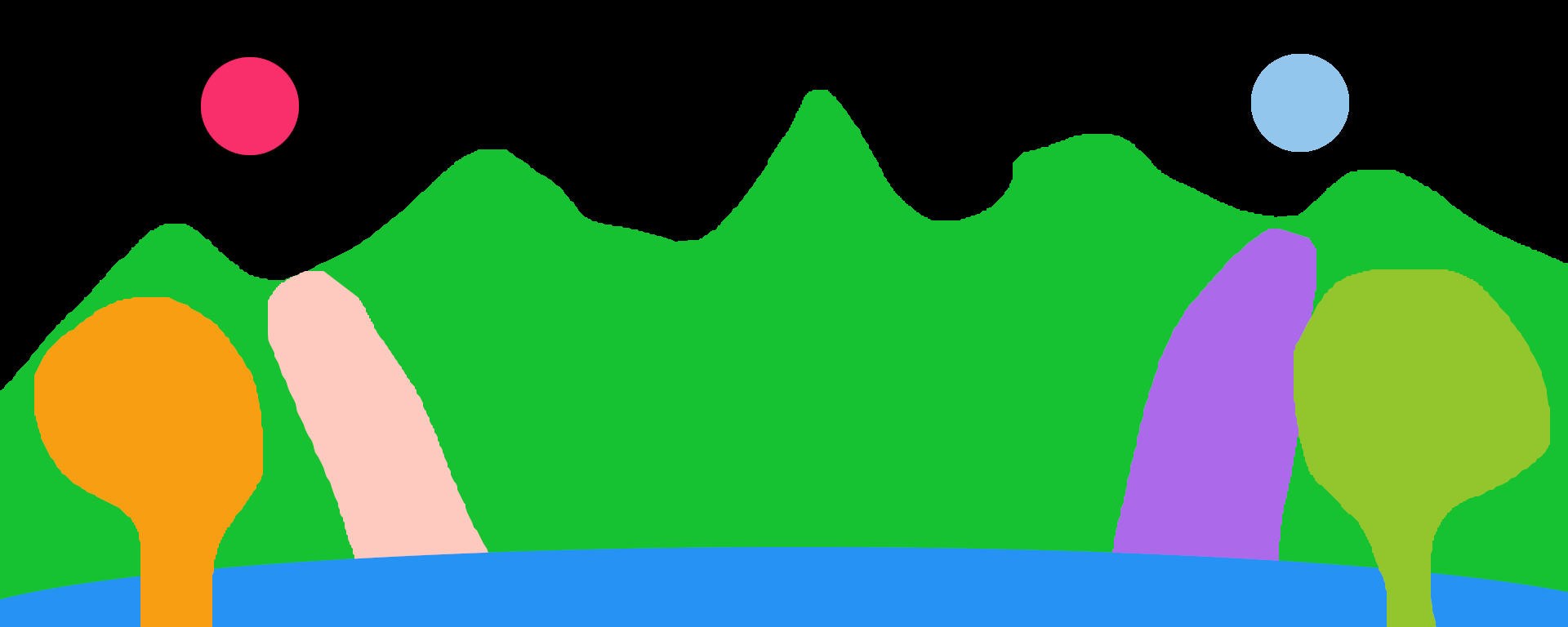}\\[0.3em]
}
{\myfont
  \hfill
  \parbox{.99\linewidth}{\footnotesize{%
  \textbf{Text prompt:} Background: \textit{``Clear deep blue sky''}, \,
  {\color{p1color} Green}: \textit{``Summer mountains''}, \,
  {\color{p2color} Red}: \textit{``The Sun''}, \,
  {\color{p3color} Pale Blue}: \textit{``The Moon''}, \,
  {\color{p4color} Light Orange}: \textit{``A giant waterfall''}, \,
  {\color{p5color} Purple}: \textit{``A giant waterfall''}, \,
  {\color{p6color} Blue}: \textit{``Clean deep blue lake''}, \,
  {\color{p7color} Orange}: \textit{``A large tree''}, \,
  {\color{p8color} Light Green}: \textit{``A large tree''}
  }}
}
  \captionof{figure}{%
  Example of large-size region-based text-to-image synthesis inspired by Korean traditional art, \textit{Irworobongdo}.
  Our \textsc{SemanticDraw} can synthesize high-resolution images from multiple, locally assigned text prompts with $\times 52.5$ faster speed of convergence.
  The size of the image is $768 \times 1920$ and we use 9 text prompt-mask pairs including the background.
  The time is measured with a RTX 2080 Ti GPU.
  Note that time takes longer than regular sized images~(\textit{e.g.,} 512~$\times$512) due to panoramic shape.
  }
  \label{fig:figure_one_problem}
  \vspace{-1em}
\end{figure}

However, directly combining multiple works together does not work as intended.
Figure~\ref{fig:figure_one_problem} illustrates an example where diffusion models fail when extended to complex real-world scenarios.
Here, inspired from the famous yet complex artwork of Korean royal folding screen, \textit{Irworobongdo} (``Painting of the Sun, Moon, and the Five Peaks'')\footnote{\url{https://g.co/arts/9DESwLeAtdtaHkGv9}}, we generate an image of size $768 \times 1920$ from nine regionally assigned text prompts as defined by a user under Figure~\ref{fig:figure_one_problem}.
At this scale, previous state-of-the-art~(SOTA) region-based controlling pipeline~\cite{bar2023multidiffusion} fails to match the designated mask regions and text prompts despite its extremely slow and, hence, cautious reverse diffusion process.
Applying a famous acceleration method LCM~\cite{luo2023lcm} on the diffusion model~\cite{bar2023multidiffusion} does not solve high-latency problem, producing noisy output in the second row in Figure~\ref{fig:figure_one_problem}.
This proves that the problem of controllability and acceleration cannot be scaled to real-world scenarios when simply combining the existing diffusion models and acceleration methods, due to their poor compatibility.

\begin{figure*}[tb]
\newcommand{\tboxheight}{0.45em}
\newcommand{\figwidth}{0.164\linewidth}
\newcommand{\figheight}{0.248\linewidth}
\newcommand{\h}{20mm}
\newcommand{\vv}{\vspace*{-0.00mm}}
\definecolor{p1color}{HTML}{F89E12}
\definecolor{p2color}{HTML}{F92F6C}
\definecolor{greenforgood}{HTML}{16C232}
\definecolor{redforbad}{HTML}{F92F6C}
  \centering
{\myfont
  \makebox[\h][c]{\hspace{-0.\linewidth}\footnotesize{%
  Background: \textit{``A photo of a Greek temple''}, \,
  {\color{p1color} Yellow}: \textit{``A photo of God Zeus with arms open''}, \,
  {\color{p2color} Red}: \textit{``A tiny sitting eagle''}
  }}\vv\\
}
    \subfloat[%
    Prompt
    \label{fig:problem2:prompt}
    ]{\includegraphics[width=\figwidth,height=\figheight]{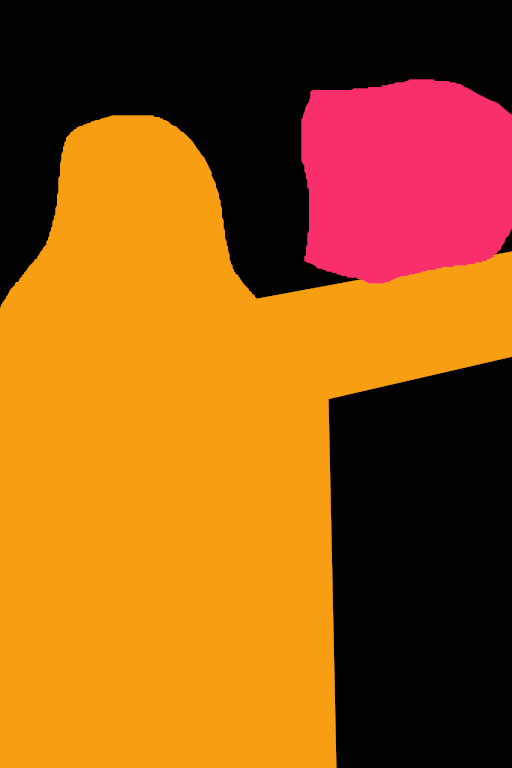}}
    \hfill
    \subfloat[%
    MD 50steps {\color{redforbad} \textbf{52s}}
    \label{fig:problem2:md}
    ]{\includegraphics[width=\figwidth,height=\figheight]{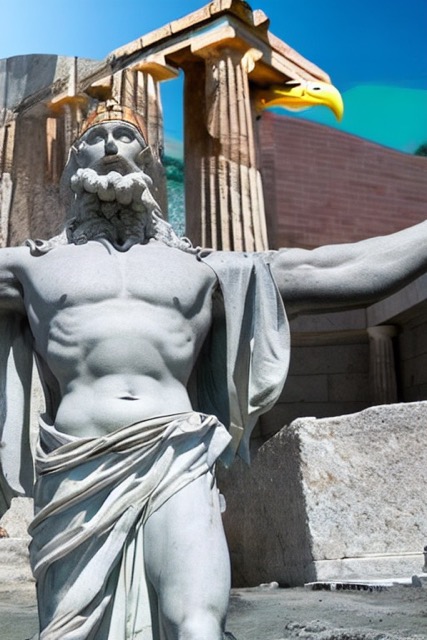}}
    \hfill
    \subfloat[%
    \scriptsize MD+LCM 5steps {\color{greenforgood} \textbf{5s}}
    \label{fig:problem2:mdlcm}
    ]{\includegraphics[width=\figwidth,height=\figheight]{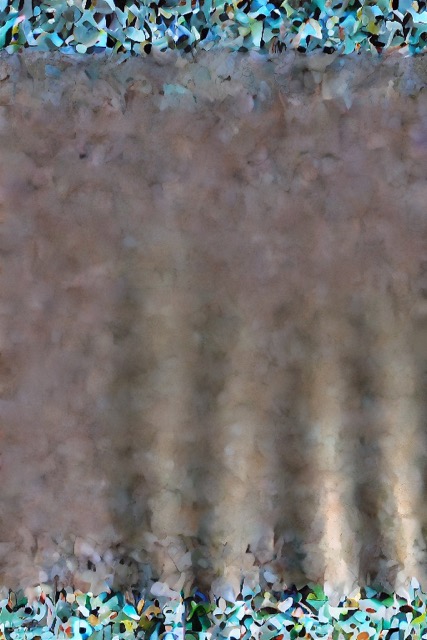}}
    \hfill
    \subfloat[%
    \scriptsize \textbf{+PreAvg} 5steps {\color{greenforgood} \textbf{5s}}
    \label{fig:problem2:preavg}
    ]{\includegraphics[width=\figwidth,height=\figheight]{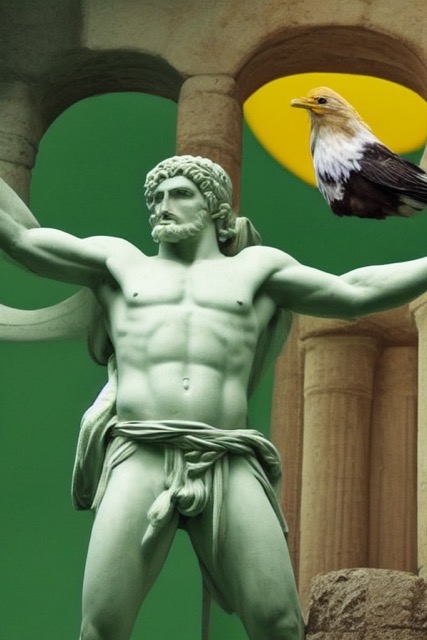}}
    \hfill
    \subfloat[%
    \scriptsize \textbf{+Bstrap} 5steps {\color{greenforgood} \textbf{5s}}
    \label{fig:problem2:bstrap}
    ]{\includegraphics[width=\figwidth,height=\figheight]{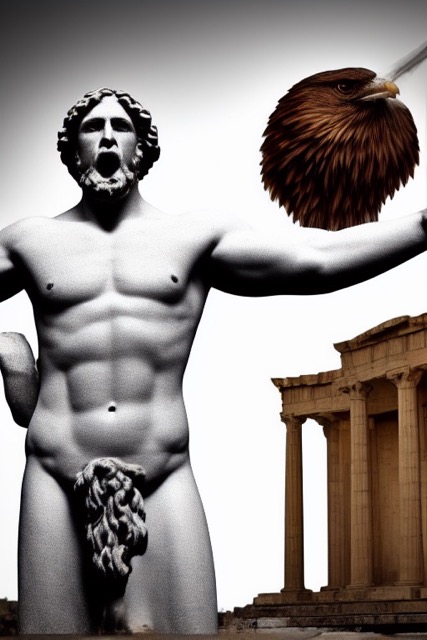}}
    \hfill
    \subfloat[%
    \scriptsize \textbf{+QMask(Ours)}5steps {\color{greenforgood} \textbf{5s}}
    \label{fig:problem2:qmask}
    ]{\includegraphics[width=\figwidth,height=\figheight]{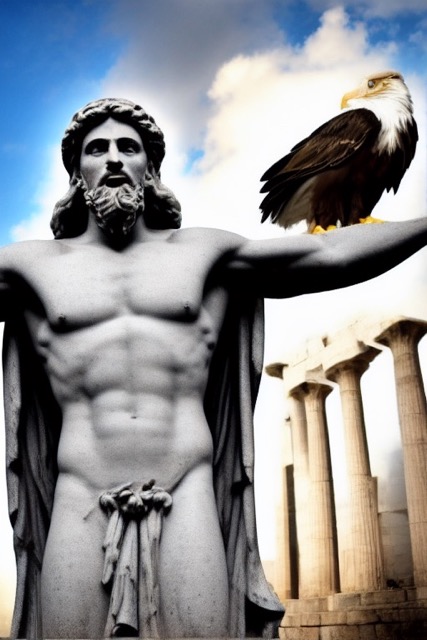}}
\\[-.5em]
  \caption{%
  Our \textsc{SemanticDraw} enables fast region-based text-to-image generation by stable acceleration of MultiDiffusion~\cite{bar2023multidiffusion}.
  PreAvg, Bstrap, and QMask stand for the \emph{latent pre-averaging}, \emph{mask-centering bootstrapping}, and \emph{quantized masks}, our first three proposed strategies.
  Each method used in (d), (e), (f) contains the method used in the previous image.
  The images are single tiles of size $768 \times 512\,$.
  }
  \label{fig:problem2}
\vspace{-.5em}
\end{figure*}

Our goal is to build a real-time pipeline for image content creation, ready for interactive user applications.
The system should be operated at least in near real-time, while maintaining stability of fine-grained regional controls.
In the end, we propose \textsc{SemanticDraw} which solves the problems from existing methods as shown in Figure~\ref{fig:problem2}.
Elaborated in Section~\ref{sec:3_method:stabilizing}, we establish a stable pipeline for accelerated image synthesis with fine-grained controls, given through multiple, locally assigned text prompts.
Building upon the rapid development from both acceleration schedulers~\cite{luo2023latent,luo2023lcm,lin2024sdxl,ren2024hyper,chadebec2024flash} and network architectures~\cite{rombach2022high,podellsdxl,sauer2024sd3} for diffusion models, we propose the first method to allow the acceleration schedulers to be compatible with region-based controllable diffusion models.
We achieve up to $\times 50$ speed-up of the multi-prompt generation while maintaining or even surpassing the image fidelity of the original algorithm~\cite{bar2023multidiffusion}.

Even after resolving the compatibility problem between the acceleration and controllability modules, generation throughput remains to be a main obstacle to interactive application.
To this end, as illustrated in Section~\ref{sec:3_method:streaming}, we restructure our multi-prompt reverse diffusion process into a pipelined architecture~\cite{kodaira2023streamdiffusion}, which we call the \emph{multi-prompt stream batch} architecture.
By bundling multi-prompt latents at different timesteps as a batched sequence of requests for image generation, we can perform the multi-prompt text-to-image synthesis endlessly by repeating a single, batched reverse diffusion.
The result is a sub-second interactive image generation framework, achieving 1.57 FPS in a single 2080 Ti GPU.
This high, stable throughput from~\textsc{SemanticDraw} allows a novel type of application for image content creation, named~\emph{semantic palette}, in which we can draw semantic masks in real-time to create an endless stream of images as in Figure~\ref{fig:figure_one} and~\ref{fig:screenshot}.
Our model-agnostic and acceleration-agnostic design allows the framework to be suitable for any existing diffusion pipelines~\cite{rombach2022high,podellsdxl,sauer2024sd3}.
We highly recommend readers to try our technical demo application of~\emph{semantic palette} in our Supplementary Material for better understanding.

\section{Related Work}
\label{sec:2_survey}
\paragraph{Accelerating Inference from Diffusion Models.}

Diffusion model~\cite{ho2020denoising,song2020denoising,rombach2022high} is a branch of generative models that sample target data distributions, \textit{e.g.}, images, videos, sounds, \textit{etc.}, by iteratively reducing randomness from pure noise.
Its earliest form~\cite{sohldickstein2015diffusion,ho2020denoising,song2020denoising} traded off inference efficiency against sample diversity and quality.
These have required thousands of iterations to generate a single image, raising a need for acceleration of inference to gain practicality.
Majority of works~\cite{song2020denoising, lu2022dpm, lu2022dpmp} achieved speed through reformulating the reverse diffusion process.
DDIM~\cite{song2020denoising} utilized a non-Markovian graphical model, and DPM-Solvers~\cite{lu2022dpm,lu2022dpmp} interpreted the generation process as Euler's method for solving ordinary differential equations, cutting the required number of inference steps from thousands down to 20.
Later, Consistency Models~\cite{song2023consistency} exploited identity map boundary condition, and Flow Matching~\cite{lipman2023flowmatching} adopted optimal transport for efficient sampling.
These became the foundations of the most recent accelerated schedulers, including latent consistency model (LCM)~\cite{luo2023latent,luo2023lcm}, SDXL-Lightning~\cite{lin2024sdxl}, Hyer-SD~\cite{ren2024hyper} and Flash Diffusion~\cite{chadebec2024flash}, which are utilized by large-scale latent diffusion models~\cite{rombach2022high,podellsdxl,esser2024scaling} in form of low-rank adaptations (LoRA)~\cite{hu2021lora}, a weight modifier upon baseline diffusion models.
Alternatively, StreamDiffusion~\cite{kodaira2023streamdiffusion} introduced a novel pipelined architecture for video-to-video transfer, video stylization, and streamed image generation from a latent consistency model~\cite{luo2023latent}.
Our \textit{multi-prompt stream batch} pipeline for interactive semantic drawing extends this philosophy with multi-prompt-based generation.

\paragraph{Controlling Generation from Diffusion Models.}

Enhanced controllability of diffusion models is another intensely investigated field of study.
There are five major subgroups: (1) modifying from intermediate latent vectors, (2) modifying from inpainting masks, (3) attaching separate conditional branches, (4) connecting a subset of prompt tokens to positions in an image, and (5) enabling finer-grained generation from multiple, region-based prompts.
The first group including ILVR~\cite{choi2021ilvr}, RePaint~\cite{lugmayr2022repaint}, and SDEdit~\cite{meng2022sdedit} attempt to hijack the intermediate latent variables in the reverse process.
SSI~\cite{li2022ssi} accelerates this group of methods by utilizing locality of edit command.
LazyDiffusion~\cite{nitzan2024lazydiffusion} take advantage of the transformer architecture to progressively edit and generate images within few seconds of latency, whereas our method builds upon arbitrary architecture and achieves sub-second generation time.
The second major group utilizes the in-painting functionality~\cite{nichol2022glide} of diffusion models for editting~\cite{lugmayr2022repaint,xie2023smartbrush,nichol2022glide,avrahami2022blended,ju2024brushnet}.
After diffusion models have become massively publicized as image generation~\cite{rombach2022high,autoamtic1111-stable-diffusion-webui} and editing~\cite{meng2022sdedit,kim2022diffusionclip,kawar2023imagic,mokady2023null,su2022dual,hertz2022prompt,liu2023more,yang2023paint} tools, the demand for easier, modularized controls on behalf of professional creators has increased.
In the third group, ControlNet~\cite{zhang2023adding} and IP-Adapter~\cite{ye2023ip} introduce simple yet effective way to append image conditioning feature to existing pre-trained diffusion models.
Our method applies orthogonally with the control methods in this group.
Various text-conditioning~\cite{hertz2022prompt,kawar2023imagic,kim2022diffusionclip,liu2023more,mokady2023null} and image-conditioning~\cite{su2022dual,yang2023paint,guo2023animatediff} methods can also be placed in this group.
The fourth group, including GLIGEN~\cite{li2023gligen} and InstanceDiffusion~\cite{wang2024instancediff} attach add-on modules to the diffusion model that focus on increasing the positional accuracy of a \textit{single} prompt.
Alternatively, we are mainly interested in a scenario where image diffusion models continuously create new images from multiple, dynamically moving, regionally assigned text prompts.
This is most related to the final group~\cite{avrahami2023spatext,alvaro2023mixtureofdiffusers,bar2023multidiffusion,qi2024lrdiff} which focus on controlling the semantic composition of the generated images.

\paragraph{Content Creation from Regional Text Prompts.}

The last group mentioned above provides a way to flexibly integrate multiple regionally assigned text prompts into a single image.
SpaText~\cite{avrahami2023spatext} achieves generation from multiple spatially localized text prompts by utilizing CLIP-based spatio-temporal representation.
Differential Diffusion~\cite{levin2023differential} and Mixture of Diffusers~\cite{alvaro2023mixtureofdiffusers} similarly operate on mask-based generation but differs in their approaches to overlapping regions and noise addition.
MultiDiffusion~\cite{bar2023multidiffusion} and more recent LRDiff~\cite{qi2024lrdiff} present simple yet effective way to generate from multiple different semantic masks: to iteratively decompose and recompose the latent images according to different regional prompts during reverse diffusion process.
This simple formulation works not only with irregular-shaped regions, but also with irregular-sized canvases.
However, as mentioned in Section~\ref{sec:1_intro} and depicted in Figure~\ref{fig:figure_one_problem}, this breakthrough has not been developed in aware of modern acceleration methods, reducing their practical attraction in this era of rapid diffusion models.
Starting from the following section, we will establish the compatibility between these type of pipeline architecture with accelerated samplers.
This opens a new type of semantic drawing application, \textsc{SemanticDraw}, where users draw images interactively with brush-type tools that paints semantic meanings as shown in Section~\ref{sec:6_discussion}.

\begin{figure*}[tb]
\definecolor{preavg}{HTML}{2692F3}
\definecolor{bootstrap}{HTML}{F89E12}
  \centering
  \begin{subcaptionbox}{%
  Bootstrapping strategy overview.
    \label{fig:pipeline:stabilize}
  }[0.60\linewidth]
  {
    \includegraphics[width=\linewidth]{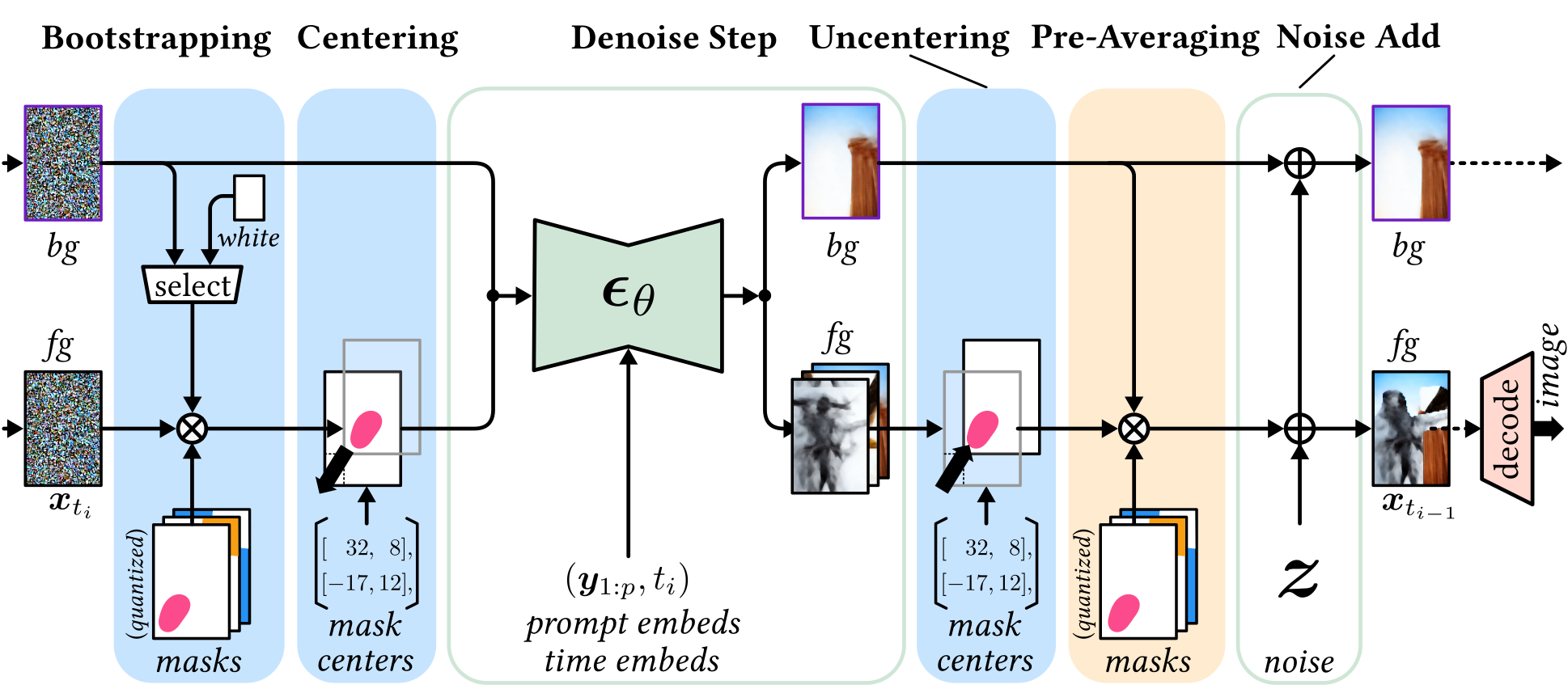}
  }
  \end{subcaptionbox}
\hfill
  \begin{subcaptionbox}{
  Multi-prompt stream batch architecture.
    \label{fig:pipeline:schematic}
  }[0.38\linewidth]
  {
    \includegraphics[width=\linewidth]{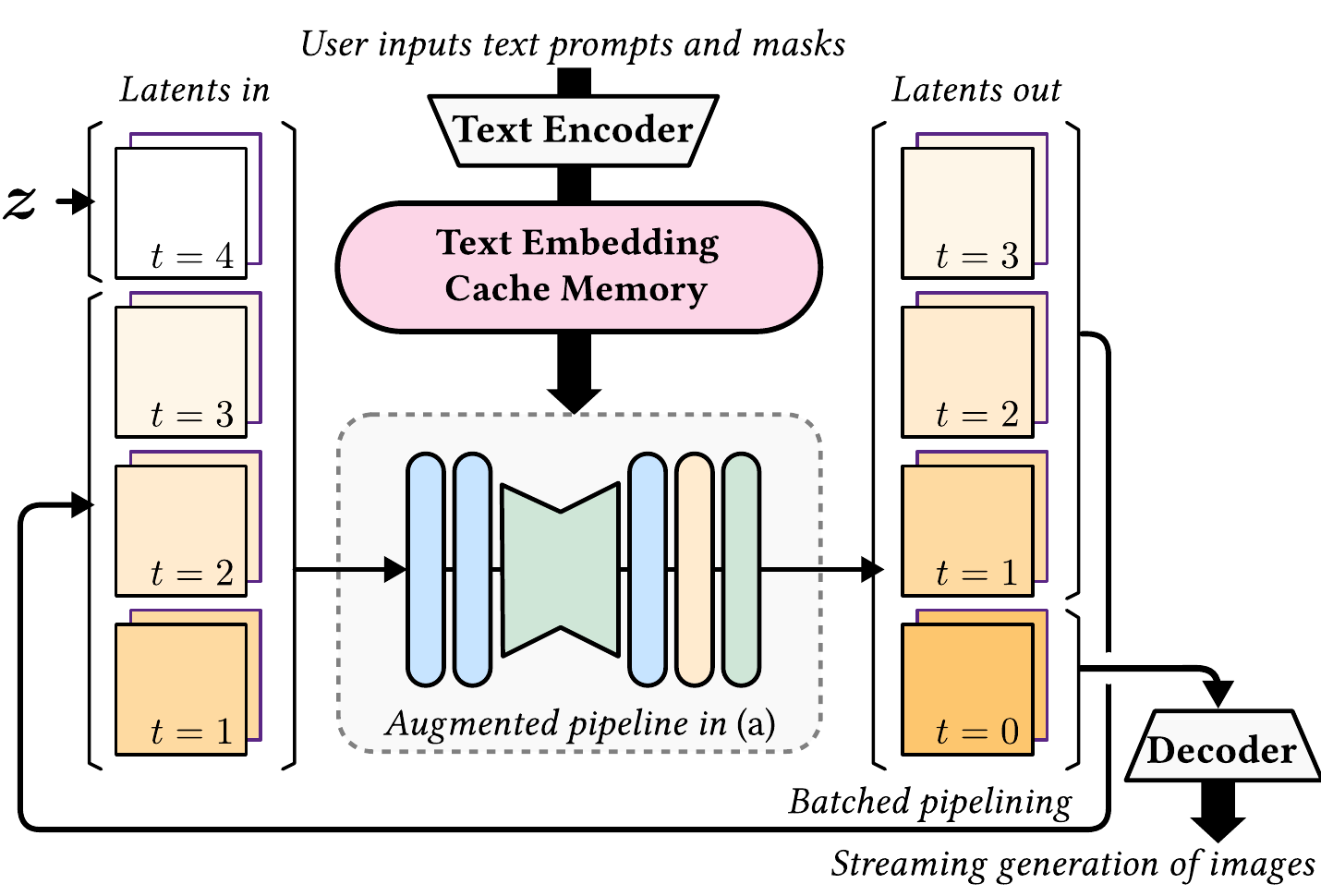}
  }
  \end{subcaptionbox}
  \caption{%
  \textsc{SemanticDraw} pipeline.
  Our acceleration technique for region-based multi-prompt generation consists of three strategies.
  Figure 4a summarizes the first two of three: (1) {\color{bootstrap} latent pre-averaging} and (2) {\color{preavg} mask-centering bootstrapping}.
  In Figure 4b, we devise \textit{multi-prompt stream batch} pipeline that aggregates foreground and background latents from different time steps to maximize the throughput of generation, enabling near real-time content creation.
  Further, text embeddings are cached for interactive brush-like interface, elaborated in Section~\ref{sec:6_discussion}.
  Our method can be applied to arbitrary diffusion pipelines.
  We also provide the full algorithm in the Supplementary Material.
  }
  \label{fig:pipeline}
\end{figure*}

\section{Method}
\label{sec:3_method}
\subsection{Preliminary}
\label{sec:3_method:preliminary}

A latent diffusion model~(LDM)~\cite{rombach2022high} $\boldsymbol{\epsilon}_{\theta}$ is an additive Gaussian noise estimator defined over a latent space.
The model $\boldsymbol{\epsilon}_{\theta}$ receives a combination of a noisy latent $\boldsymbol{x}\,$, a text prompt embedding $\boldsymbol{y}$, and a timestep $t \in [0, T]$.
It outputs an estimation of the noise $\boldsymbol{\epsilon}$ that was mixed with the true latent $\boldsymbol{x}_{0}\,$.
At inference, the diffusion model $\boldsymbol{\epsilon}_{\theta}$ is consulted multiple times to {estimate} a latent $\hat{\boldsymbol{x}}_{0} \approx \boldsymbol{x}_{0}$, which correlates to the information described in the conditional input~$\boldsymbol{y}\,$, starting from a pure noise $\boldsymbol{x}_{T} \sim \mathcal{N}(0, 1)^{HWD}\,$.
Each of the recursive calls to the reverse diffusion process can be expressed as a summation of a denoising term and a noise-adding term to the intermediate latent:
\begin{equation}
    \label{eq:high-level_algorithm_single}
    \boldsymbol{x}_{t_{i - 1}} = \textsc{Step} ( \boldsymbol{x}_{t_{i}}, \boldsymbol{y}, i, \boldsymbol{\epsilon} ; \boldsymbol{\epsilon}_{\theta}, \alpha, \boldsymbol{t} )\,,
\end{equation}
where, we denote $i$ as the index of the current time step $t_{i}\,$.
Note that the newly added noise~$\boldsymbol{\epsilon}$ depends on the type of scheduler.

Although this abstract form embraces almost every generation algorithm of diffusion models~\cite{ho2020denoising,song2020denoising,lu2022dpm}, it does not consider practical scenarios of our interest: (1) when the desired shape ($H' \times W'$) of the latent $\hat{\boldsymbol{x}}'_{0}$ is different from that of the training set ($H \times W$) or (2) multiple text prompts $\boldsymbol{y}_{1}, \ldots, \boldsymbol{y}_{p}$ correlate to different regions of the generated images.
MultiDiffusion~\cite{bar2023multidiffusion} is one of the pioneers to deal with this problem.
Their main idea is to aggregate~(\textsc{AggrStep}) multiple overlapping tiles of intermediate latents with simple averaging.
That is, for every sampling step $t_{i}\,$, perform:
\begin{align}
    \label{eq:multidiffusion}
    \boldsymbol{x}'_{t_{i - 1}} &= \textsc{AggrStep} ( \boldsymbol{x}'_{t_{i}}, \boldsymbol{y}, i, \mathcal{W} ; \textsc{Step} ) \\
    \label{eq:multidiffusion_detail}
    &= \frac{\sum_{\boldsymbol{w} \in \mathcal{W}} \textsc{Step} ( \texttt{crop} ( \boldsymbol{w} \odot \boldsymbol{x}'_{t_{i}} ), \boldsymbol{y}_{\boldsymbol{w}}, i, \boldsymbol{\epsilon})}{\sum_{\boldsymbol{w} \in \mathcal{W}} \boldsymbol{w} }\,,
\end{align}
where $\odot$ is an element-wise multiplication, $\boldsymbol{w} \in \mathcal{W} \subset \{ 0, 1 \}^{H'  W'}$ is a binary mask for each latent tile, $\boldsymbol{y}_{\boldsymbol{w}}$ is a conditional embedding corresponding to the tile $\boldsymbol{w}\,$, and $\texttt{crop}$ is a cropping operation to chop down large $\boldsymbol{x}'_{t_{i}}$ into tiles of same size as training image latents.

\subsection{Acceleration-Compatible Regional Controls}
\label{sec:3_method:stabilizing}

Our objective is to build an accelerated solution to image generation from multiple regionally assigned text prompts.
Unfortunately, simply replacing the Stable Diffusion (SD) model~\cite{rombach2022high,podellsdxl,esser2024scaling} with an acceleration module, such as Latent Consistency Model (LCM)~\cite{luo2023latent} or SDXL-Lightning~\cite{lin2024sdxl}, \textit{etc.}, and updating the default DDIM sampler~\cite{song2020denoising} with the corresponding accelerated sampler~\cite{luo2023latent,karras2022elucidating} does not work in general.
This incompatibility greatly limits potential applications of \emph{both} acceleration~\cite{luo2023latent,lin2024sdxl,ren2024hyper,chadebec2024flash} and region-based control techniques~\cite{avrahami2023spatext,bar2023multidiffusion}.
We discuss each of the causes and seek for faster and stronger alternatives.
In summary, our stabilization trick consists of three strategies: (1) \emph{latent pre-averaging}, (2) \emph{mask-centering bootstrapping}, and (3) \emph{quantized masks}.

\paragraph{Step 1: Achieving Compatibility through Latent Pre-Averaging.}

The primary reason for the blurry image of the second row of Figure~\ref{fig:figure_one_problem} is that the previous algorithm~\cite{bar2023multidiffusion} is not aware of different types of underlying reverse diffusion step functions \textsc{Step}.
While the reverse diffusion algorithms can be categorized into two types: (1) additional noise at each step~\cite{lu2022dpm, luo2023latent}, (2) no additional noise at each step~\cite{song2020denoising, bar2023multidiffusion}, the previous SOTA region-based controllable method~\cite{bar2023multidiffusion} falls into the latter.
Hence, applying the averaging aggregation of the method~\cite{bar2023multidiffusion} cancels the prompt-wise added noises in~\textsc{Step}, which leads to overly smooth latents.
We can avoid this problem with a simple workaround.
First, we split the \textsc{Step} function into a deterministic denoising part~(\textsc{Denoise}) and an optional noise addition:
\begin{align}
    \label{eq:high-level_algorithm_single_split}
    \boldsymbol{x}_{t_{i - 1}} &= \tilde{\boldsymbol{x}}_{t_{i - 1}} + \eta_{t_{i - 1}} \boldsymbol{\epsilon} \\
    \label{eq:high-level_algorithm_single_split_detail}
    &= \textsc{Denoise} ( \boldsymbol{x}_{t_{i}}, \boldsymbol{y}, i ; \boldsymbol{\epsilon}_{\theta}, \alpha, \boldsymbol{t} ) + \eta_{t_{i - 1}} \boldsymbol{\epsilon} \,,
\end{align}
where $\eta_{t}$ is an algorithm-dependent parameter.
The averaging of equation~\eqref{eq:multidiffusion_detail} is then applied to the output of the denoising part $\tilde{\boldsymbol{x}}_{t_{i - 1}}\,$, instead of the output of the full step $\boldsymbol{x}_{t_{i - 1}}\,$.
Note that the noise is added after aggregation step.
\begin{equation}
    \label{eq:stable_multidiffusion}
    \boldsymbol{x}'_{t_{i - 1}} = \textsc{AggrStep} ( \boldsymbol{x}'_{t_{i}}, \boldsymbol{y}, i, \mathcal{W} ; \textsc{Denoise} ) + \eta_{t_{i - 1}} \boldsymbol{\epsilon}\,.
\end{equation}
As it can be seen in Figure~\ref{fig:problem2:preavg}, this change alleviates the compatibility issue with acceleration methods like LCM.

\paragraph{Step 2: Mask-Centering Bootstrapping for Few-Step Generation.}

The second cause of the incompatibility lies in the bootstrapping stage of the previous method~\cite{bar2023multidiffusion}.
MultiDiffusion~\cite{bar2023multidiffusion} introduced bootstrapping stages that replace the background latents with random colors in the first 40\% of total steps.
This is performed to cut out the generated regions outside of object masks, which claims to enhance mask-fidelity.
In original form, the perturbation introduced by the bootstrapping cancels out during long inference steps.
However, as we decrease the number of timesteps in ten-fold from $n = 50$ steps to $n = 4$ or $5$ steps, the number of bootstrapping stage is reduced down to $n = 2\,$.
Regrettably, this magnifies the effect of perturbation introduced by the random color latents in the bootstrapping phase, and results in leakage of mixed colors onto the final image as shown in Figure~\ref{fig:problem2}.
Instead, we propose to use a mixture of white background and aggregation of contents co-generated from other regional prompts~(blue in Figure~\ref{fig:pipeline:stabilize}), which alleviates the problem and allows compatibility with the accelerated generation as seen in Figure~\ref{fig:problem2:bstrap}.

Furthermore, we empirically found that first two steps of reverse diffusion process determine the overall structure of generated images when sampling with accelerated schedulers.
Even after the first step, the network formulates the rough structure of the objects being created.
The problem is that diffusion models exhibit a strong tendency to generate screen-centered objects than off-centered ones, following image datasets~\cite{schuhmann2021laion} they are trained for.
After the first step, the object for every mask is generated at the center of the screen, not at the center of the mask.
Off-centered objects are often masked out by the pre-averaging step (yellow in Figure~\ref{fig:pipeline:stabilize}).
The final results often neglect small, off-centered regional prompts, and the large objects are often unnaturally cut, lacking harmonization within the image.
To prevent this, we propose \emph{mask centering} strategy~(pink in Figure~\ref{fig:pipeline:stabilize}) to exploit the \textit{center-bias} of the diffusion model.
Especially, for the first two steps of generation, we shift the intermediate latents from each prompt to the center of the frame before being handled by the noise estimator $\boldsymbol{\epsilon}_{\theta}\,$.
The result of Step 2 can be seen in Figure~\ref{fig:problem2:bstrap}.

\begin{figure}[tb]
  \includegraphics[width=.95\linewidth]{figures/bootstrapping/qmask.png}
  \vspace{-.8em}
  \caption{%
  Quantized mask.
  As the last of three stabilizing techniques, binary masks are blurred and quantized by scheduler noise level to trade off between mask-fidelity and overall harmonization.
  See Supplementary Material for utilizing this trade-off.
  }
  \label{fig:maskquant:schematic}
\vspace{-1.5em}
\end{figure}

\paragraph{Step 3: Quantized Mask for Seamless Generation.}

Another problem from the reduced number of inference steps is that \emph{harmonization} of the generated content becomes more difficult.
As Figure~\ref{fig:problem2:bstrap} shows, all the objects appear to be salient and their abrupt boundaries are visible between regions.
This is because the number of later sampling steps that contribute to the harmonization is now insufficient.
In contrast, the baseline with long reverse diffusion steps~\ref{fig:problem2:md} effectively smooth out the mask boundaries by consecutively adding noises and blurring them.
To mitigate this issue, we develop an alternative way to seamlessly amalgamate generated regions: \emph{quantized masks}, shown in Figure~\ref{fig:maskquant:schematic}.
Given a binary mask, we obtain a smoothened mask by applying Gaussian blur.
Then, we quantize the real-numbered mask by the noise levels of the diffusion sampler.
As Figure~\ref{fig:pipeline:stabilize} illustrates, for each denoising step, we use a mask with corresponding noise level.
Since the noise levels monotonically decrease throughout iterations, the coverage of a mask gradually increases along with each sampling step, gradually mixing the boundary regions.
The final result can be seen from Figure~\ref{fig:problem2:qmask}.
This relaxation of semantic masks also provides intuitive interpretation of \emph{brushes}, one of the most widely used tool in professional graphics editing software.
We will revisit this interpretation in Section~\ref{sec:6_discussion}.

\subsection{Optimization for Throughput}
\label{sec:3_method:streaming}

As mentioned in Section~\ref{sec:1_intro}, achieving real-time response is important for practical end-user application.
Inspired by StreamDiffusion~\cite{kodaira2023streamdiffusion}, we reconstruct our region-based text-to-image synthesis framework into a pipelined architecture to maximize the throughput of image generation.

\paragraph{Multi-Prompt Stream Batch Architecture.}

Figure~\ref{fig:pipeline:schematic} illustrates the architecture and the interfaces of our pipeline.
Instead of the typical mini-batched use of diffusion model with synchronized timesteps, the noise estimator is fed with a new input image every timestep along with the last processed batch of images.
In other words, each image in a mini-batch has different timestep.
This architecture hides the latency caused by multi-step algorithm of reverse diffusion.
Restructuring our stabilized framework in~\ref{fig:pipeline:stabilize} takes several steps.
The quantized masks, the background images, the noises, and the prompt embeddings differ along each timesteps and should be saved separately.
Instead of a single image, we change the architecture to process a mini-batch of images of different prompts and masks to the U-Net at every timestep, as depicted in Figure~\ref{fig:pipeline:schematic}.
We call this the \textit{multi-prompt stream batch} architecture.
To further reduce the latency, we added asynchronous pre-calculation step applied only when a user command changes the configuration of the text prompts and masks.
This allows interactive brush-like interfaces elaborated in Section~\ref{sec:6_discussion}.

\paragraph{Optimizing Throughput.}

Additional increase of throughput can be achieved by using a compressed autoencoder such as Tiny AutoEncoder~\cite{tinyvae2023}.
Detailed analysis on the effect of throughput optimization is in Table~\ref{tab:exp_speed}.

\section{Experiment}
\label{sec:5_exp}
We provide comprehensive evaluation of our \textsc{SemanticDraw} using various types of acceleration modules and samplers.
We compare our experiments based on the public checkpoints of Stable Diffusion 1.5~\cite{rombach2022high}, SDXL~\cite{podellsdxl}, and SD3~\cite{sauer2024sd3}.
However, we note that our method can be applied to any community-trained models using DreamBooth~\cite{ruiz2023dreambooth}.
More results can be found in Section~S2 of our Supplementary Materials.

\subsection{Quality of Generation}
\label{sec:5_exp:stabilize}

\paragraph{Generation from Multiple Region-Based Prompts.}

We first demonstrate the stability and speed of our algorithm for image generation from multiple regionally assigned text prompts.
The evaluation is based on COCO validation dataset~\cite{lin2014microsoft}, where we generate images from the image captions as background prompts and object masks with categories as foreground prompts.
The public latent diffusion models~\cite{rombach2022high,podellsdxl,sauer2024sd3} are trained for specific range of image sizes, and reportedly fail when given image sizes are small.
Since COCO datasets consists of relatively small images compared to the default size the models were trained for, we rescale the object masks with nearest neighbor interpolation to the default size of each model.
This is $512 \times 512$ for SD1.5~\cite{rombach2022high} and $1024 \times 1024$ for SDXL~\cite{podellsdxl} and SD3~\cite{sauer2024sd3}.
To compare the image fidelity, we use Fréchet Inception Distance (FID)~\cite{heusel2017gans} and Inception Score (IS)~\cite{salimans2016is}.
We also use CLIP scores~\cite{hessel2021clipscore} to compare the text prompt fidelity.
We separate the foreground score (CLIP$_{\text{fg}}$), which is obtained by taking the average CLIP score between each generated image and corresponding set of foreground object categories, from the background score (CLIP$_{\text{bg}}$), which is a measured between images and their corresponding COCO captions.
Tables~\ref{tab:quality-region-lcm} through~\ref{tab:quality-region-flash} summarizes the results.

%-------------------------------------------------------------------------
\begin{table}[t]
  \caption{%
  Comparison of generation from region-based prompts between DDIM~\cite{song2020denoising}~(default) and LCM~\cite{luo2023latent} sampler.
  }
  \label{tab:quality-region-lcm}
    \vspace{-.5em}
  \centering
  \resizebox{\linewidth}{!}{%
  \begin{tabular}{@{}ll|ccccc@{}}
    \toprule
    Method & Sampler & FID $\downarrow$ & IS $\uparrow$ & CLIP$_{\text{fg}}$ $\uparrow$ & CLIP$_{\text{bg}}$ $\uparrow$ & Time (s) $\downarrow$ \\
    \midrule\midrule
    \textbf{SD1.5} ($512 \times 512$)\\
    MultiDiffusion (Ref.) & DDIM~\cite{song2020denoising} & 70.93 \goldmedal & 16.24 \goldmedal & 24.09 \silvermedal & 27.55 \goldmedal & 14.1 \bronzemedal \\
    \midrule
    MultiDiffusion~(MD) & LCM~\cite{luo2023latent} & 270.55 \bronzemedal & 2.653 \bronzemedal & 22.53 \bronzemedal & 19.63 \bronzemedal & 1.7 \silvermedal \\
    \textbf{SemanticDraw~(Ours)} & LCM~\cite{luo2023latent} & 93.93 \silvermedal & 14.12 \silvermedal & 24.14 \goldmedal & 24.00 \silvermedal & 1.3 \goldmedal \\
  \bottomrule
  \end{tabular}%
  }
\vspace{-.5em}
\end{table}

\begin{table}[t]
  \caption{%
  Comparison of generation from region-based prompts between DDIM~\cite{song2020denoising}~(default) and Hyper-SD~\cite{ren2024hyper} sampler.
  }
  \label{tab:quality-region-hypersd}
    \vspace{-.5em}
  \centering
  \resizebox{\linewidth}{!}{%
  \begin{tabular}{@{}ll|ccccc@{}}
    \toprule
    Method & Sampler & FID $\downarrow$ & IS $\uparrow$ & CLIP$_{\text{fg}}$ $\uparrow$ & CLIP$_{\text{bg}}$ $\uparrow$ & Time (s) $\downarrow$ \\
    \midrule\midrule
    \textbf{SD1.5} ($512 \times 512$)\\
    MultiDiffusion (Ref.) & DDIM~\cite{song2020denoising} & 70.93 \goldmedal & 16.24 \goldmedal & 24.09 \silvermedal & 27.55 \goldmedal & 14.1 \bronzemedal \\
    \midrule
    MultiDiffusion~(MD) & Hyper-SD~\cite{ren2024hyper} & 168.34 \bronzemedal & 10.12 \bronzemedal & 20.08 \bronzemedal & 15.90 \bronzemedal & 1.7 \silvermedal \\
    \textbf{SemanticDraw~(Ours)} & Hyper-SD~\cite{ren2024hyper} & 98.60 \silvermedal & 14.90 \silvermedal & 24.48 \goldmedal & 23.31 \silvermedal & 1.3 \goldmedal \\
  \bottomrule
  \end{tabular}%
  }
\vspace{-.5em}
\end{table}

\begin{table}[t]
  \caption{%
  Comparison of generation from region-based prompts between DDIM~\cite{song2020denoising}~(default) and Euler Discrete~\cite{karras2022elucidating} sampler.
  }
  \label{tab:quality-region-lightning}
    \vspace{-.5em}
  \centering
  \resizebox{\linewidth}{!}{%
  \begin{tabular}{@{}ll|ccccc@{}}
    \toprule
    Method & Sampler & FID $\downarrow$ & IS $\uparrow$ & CLIP$_{\text{fg}}$ $\uparrow$ & CLIP$_{\text{bg}}$ $\uparrow$ & Time (s) $\downarrow$ \\
    \midrule\midrule
    \textbf{SDXL} ($1024 \times 1024$) \\
    MultiDiffusion (Ref.) & DDIM~\cite{song2020denoising} & 73.77 \goldmedal & 16.31 \goldmedal & 24.16 \silvermedal & 28.11 \goldmedal & 50.6 \bronzemedal \\
    \midrule
    MultiDiffusion~(MD) & EulerDiscrete~\cite{karras2022elucidating} & 572.95 \bronzemedal & 1.328 \bronzemedal & 21.02 \bronzemedal & 17.36 \bronzemedal & 4.3 \silvermedal \\
    \textbf{SemanticDraw~(Ours)} & EulerDiscrete~\cite{karras2022elucidating} & 84.27 \silvermedal & 15.04 \silvermedal & 24.19 \goldmedal & 24.22 \silvermedal & 3.6 \goldmedal \\
  \bottomrule
  \end{tabular}%
  }
\vspace{-.5em}
\end{table}

\begin{table}[t]
  \caption{%
  Comparison of generation from region-based prompts between Flow Match Euler Discrete~\cite{esser2024scaling}~(default) and Flash Flow Match Euler Discrete~\cite{chadebec2024flash} sampler.
  }
  \label{tab:quality-region-flash}
    \vspace{-.5em}
  \centering
  \resizebox{\linewidth}{!}{%
  \begin{tabular}{@{}ll|ccccc@{}}
    \toprule
    Method & Sampler & FID $\downarrow$ & IS $\uparrow$ & CLIP$_{\text{fg}}$ $\uparrow$ & CLIP$_{\text{bg}}$ $\uparrow$ & Time (s) $\downarrow$ \\
    \midrule\midrule
    \textbf{SD3} ($1024 \times 1024$) \\
    MultiDiffusion (Ref.) & FlowMatch~\cite{esser2024scaling} & 166.42 \silvermedal & 8.517 \silvermedal & 20.66 \silvermedal & 16.39 \silvermedal & 46.3 \bronzemedal \\
    \midrule
    MultiDiffusion~(MD)  & FlashFlowMatch~\cite{chadebec2024flash} & 209.36 \bronzemedal & 5.347 \bronzemedal & 19.83 \bronzemedal & 14.48 \bronzemedal & 4.0 \silvermedal \\
    \textbf{SemanticDraw~(Ours)} & FlashFlowMatch~\cite{chadebec2024flash} & 79.2 \goldmedal & 17.41 \goldmedal & 23.59 \goldmedal & 27.83 \goldmedal & 3.2 \goldmedal \\
  \bottomrule
  \end{tabular}%
  }
  \vspace{-1em}
\end{table}
%-------------------------------------------------------------------------

\begin{figure}[tb]
\newcommand{\h}{\hspace{0.1em}}
\newcommand{\figwidth}{0.243\linewidth}
\newcommand{\figheighta}{0.162\linewidth}
\newcommand{\figheightb}{0.243\linewidth}
\newcommand{\figheight}{0.3645\linewidth}
\newcommand{\hh}{20mm}
\newcommand{\vv}{\vspace*{-0.00mm}}
\definecolor{p1color}{HTML}{F89E12}
\definecolor{p2color}{HTML}{F92F6C}
\definecolor{p3color}{HTML}{2692F3}
  \centering
{\myfont
  \makebox[\hh][c]{\hspace{-0.\linewidth}\scriptsize{%
  Background: \textit{``Plain wall''}, \,
  {\color{p1color} Yellow}: \textit{``A desk''}, \,
  {\color{p2color} Red}: \textit{``A flower vase''}, \,
  {\color{p3color} Blue}: \textit{``A window''}
  }}\vv\\
}
  \hfill
    \subfloat{\includegraphics[width=\figwidth,height=\figheighta]{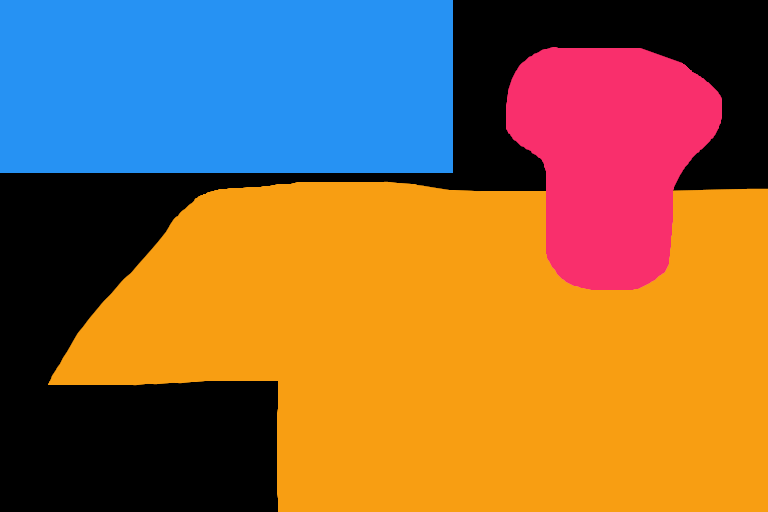}}\h
    \subfloat{\includegraphics[width=\figwidth,height=\figheighta]{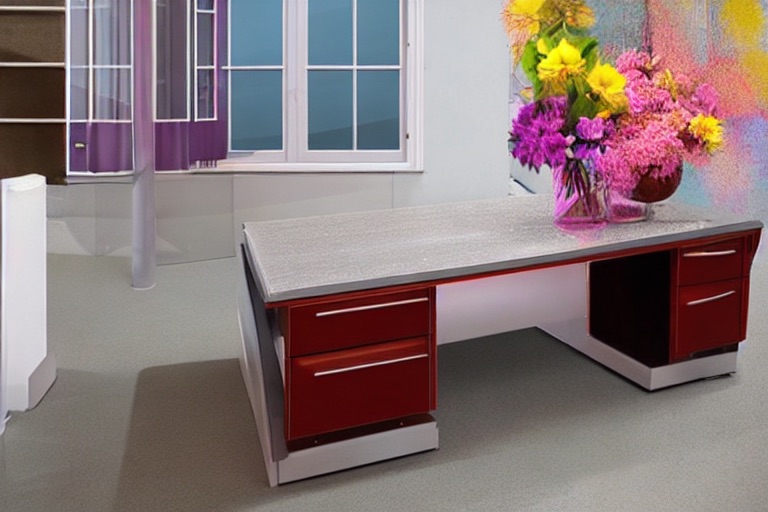}}\h
    \subfloat{\includegraphics[width=\figwidth,height=\figheighta]{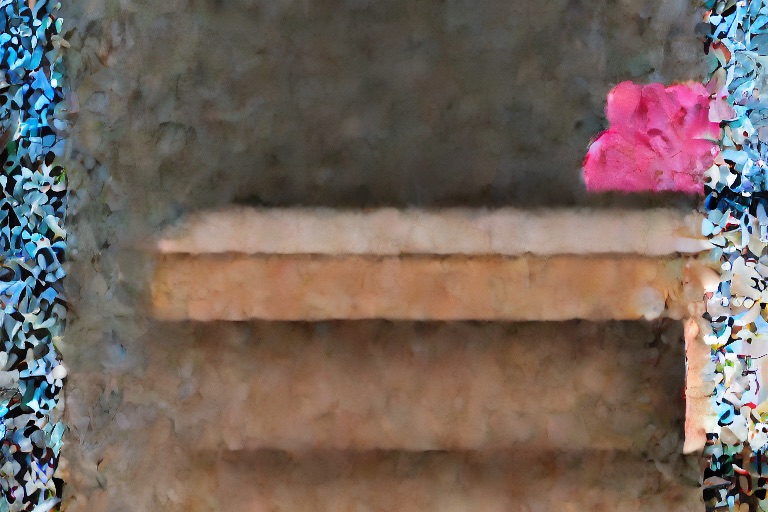}}\h
    \subfloat{\includegraphics[width=\figwidth,height=\figheighta]{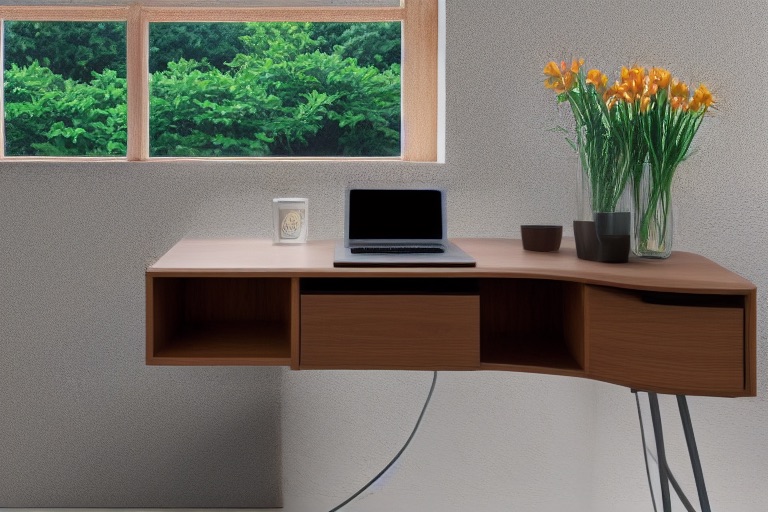}}
  \hfill
    \\
{\myfont
  \makebox[\hh][c]{\hspace{-0.\linewidth}\scriptsize{%
  Background: \textit{``A photo of backyard''}, \,
  {\color{p1color} Yellow}: \textit{``A yellow bird''}, \,
  {\color{p2color} Red}: \textit{``A red bird''}
  }}\vv\\
}
  \hfill
    \subfloat{\includegraphics[width=\figwidth,height=\figheightb]{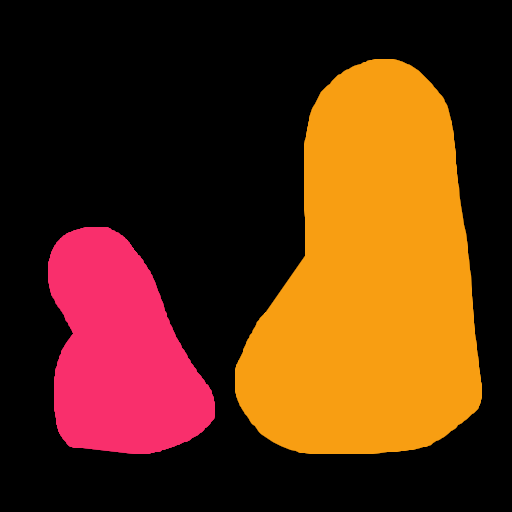}}\h
    \subfloat{\includegraphics[width=\figwidth,height=\figheightb]{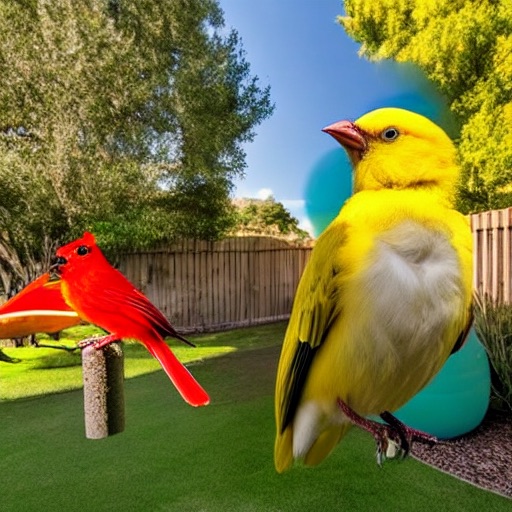}}\h
    \subfloat{\includegraphics[width=\figwidth,height=\figheightb]{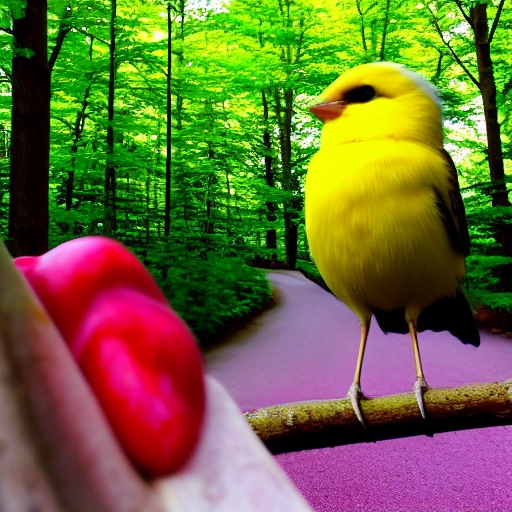}}\h
    \subfloat{\includegraphics[width=\figwidth,height=\figheightb]{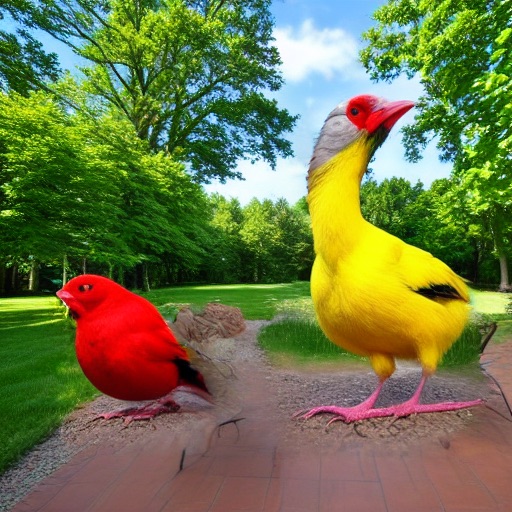}}
  \hfill
    \\
{\myfont
  \makebox[\hh][c]{\hspace{-0.\linewidth}\scriptsize{%
  Background: \textit{``A floor''}, \,
  {\color{p1color} Yellow}: \textit{``A box''}, \,
  {\color{p2color} Red}: \textit{``A tiny head of a cat''}
  }}\vv\\
}
  \hfill
    \subfloat{\includegraphics[width=\figwidth,height=\figheightb]{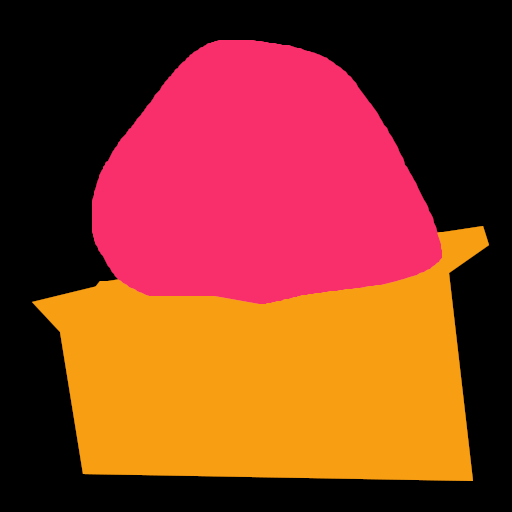}}\h
    \subfloat{\includegraphics[width=\figwidth,height=\figheightb]{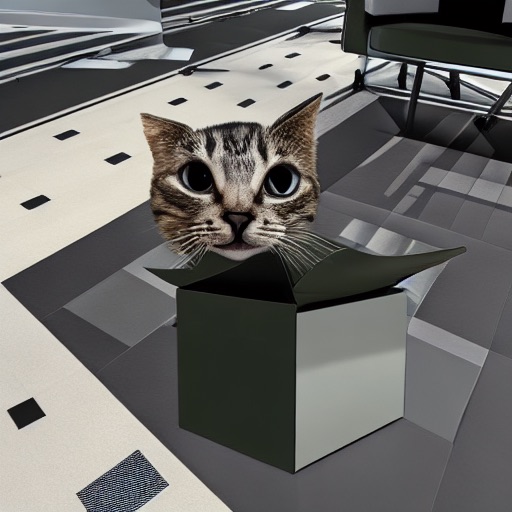}}\h
    \subfloat{\includegraphics[width=\figwidth,height=\figheightb]{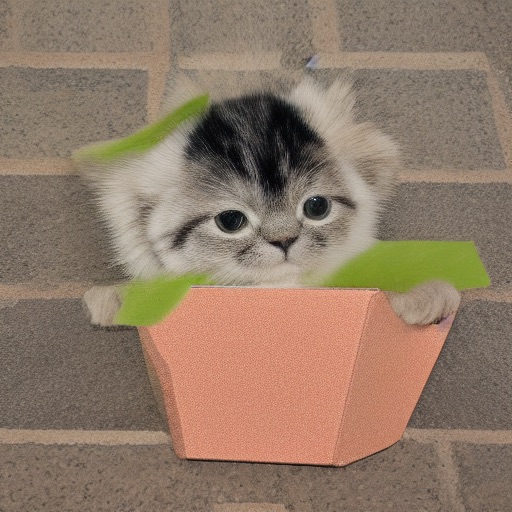}}\h
    \subfloat{\includegraphics[width=\figwidth,height=\figheightb]{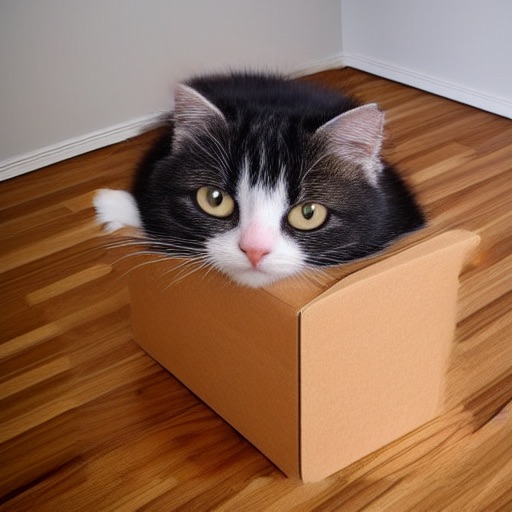}}
  \hfill
    \\
{\myfont
\resizebox{.22\linewidth}{0.4\baselineskip}{%
  \makebox[\hh][c]{\hspace{-0.\linewidth}\scriptsize{%
  Background: \textit{``A photo''}, \,
  {\color{p1color} Yellow}: \textit{``A smiling girl''}, \,
  {\color{p2color} Red}: \textit{``A cool beret hat''}, \,
  {\color{p3color} Blue}: \textit{``Sky at noon''}
  }}
}\vv\\
}
  \hfill
    \addtocounter{subfigure}{-12}
    \subfloat[Prompt]{\includegraphics[width=\figwidth,height=\figheight]{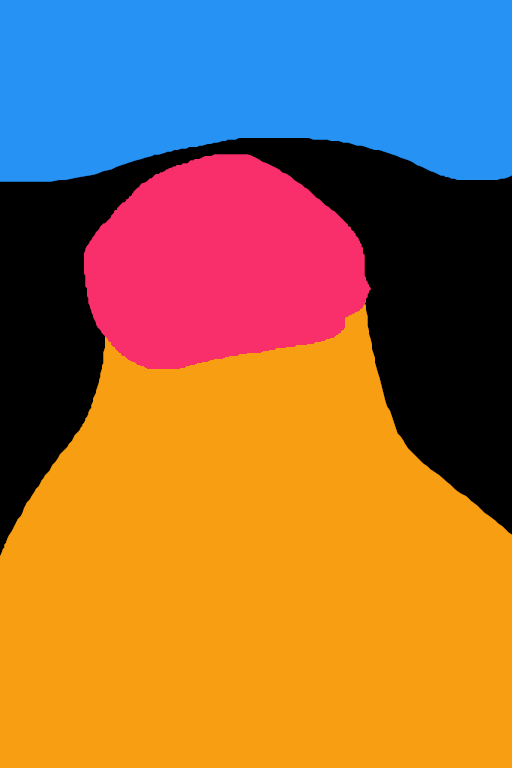}}\h
    \subfloat[MD, 50 steps]{\includegraphics[width=\figwidth,height=\figheight]{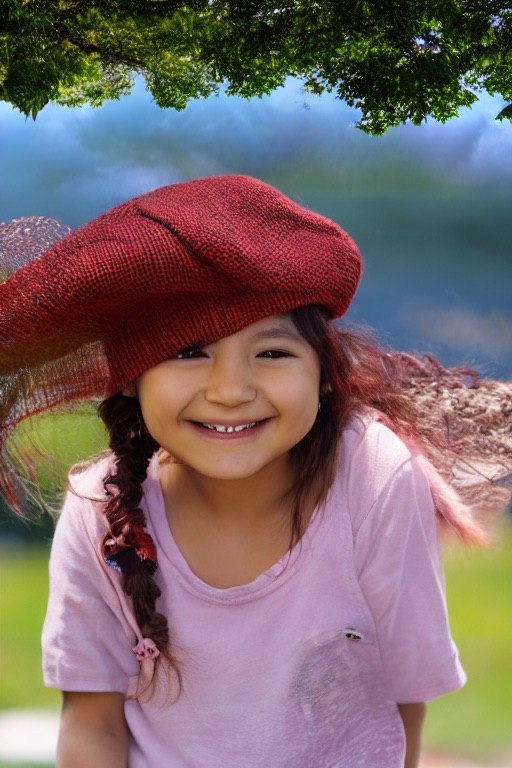}}\h
    \subfloat[MD+LCM,\\\phantom{(c) }5 steps]{\includegraphics[width=\figwidth,height=\figheight]{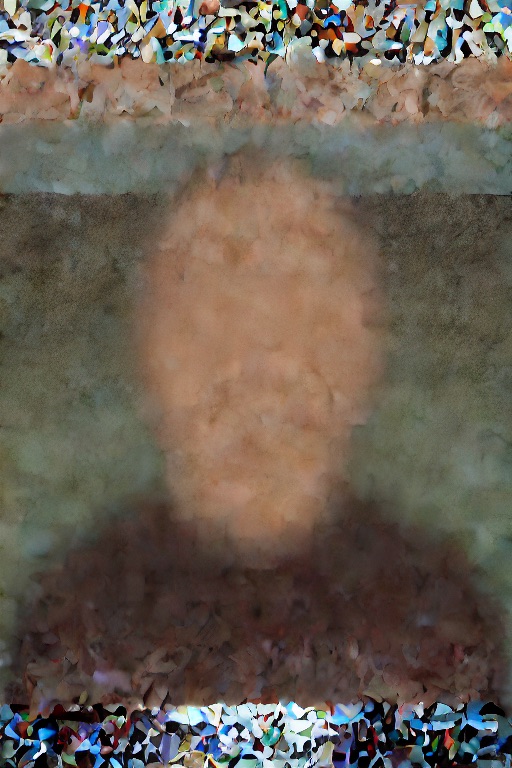}}\h
    \subfloat[{\textbf{Ours, 5 steps}}]{\includegraphics[width=\figwidth,height=\figheight]{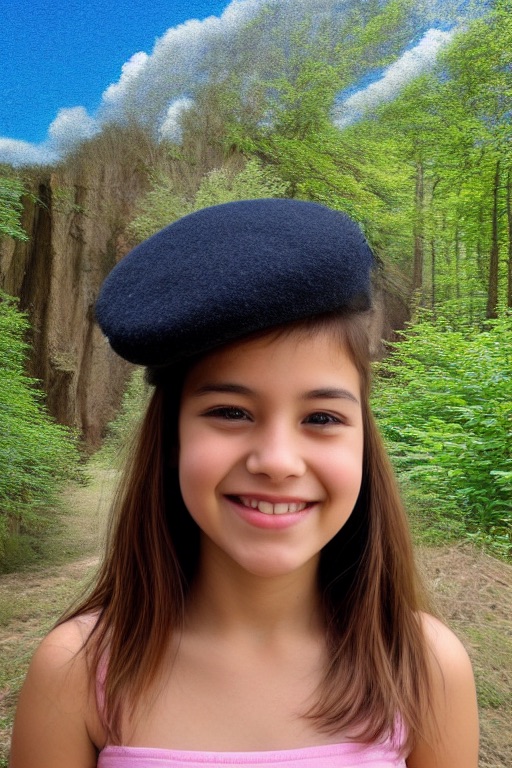}}
  \hfill
\\[-.5em]
  \caption{%
  Region-based text-to-image synthesis results.
  Our stabilization methods accelerate MultiDiffusion~\cite{bar2023multidiffusion} up to $\times$10 while preserving quality.
  }
  \label{fig:exp_region}
\vspace{-1em}
\end{figure}

We implement MultiDiffusion~\cite{bar2023multidiffusion} for SDXL~\cite{podellsdxl} and SD3~\cite{sauer2024sd3} simply by changing the pipelines, accelerator LoRAs~\cite{hu2021lora}, and schedulers, from the official implementation.
Even though schedulers with higher numbers of iterations generally produce better quality images~\cite{song2020denoising}, the tables show that our accelerated pipeline achieves comparable quality with more than $\times 10$ reduction of time.
These results demonstrate that our method provides universal acceleration under different types of diffusion pipelines (SD1.5~\cite{luo2023lcm} , SDXL~\cite{podellsdxl}, SD3~\cite{esser2024scaling}), noise schedulers (DDIM~\cite{song2020denoising}, LCM~\cite{luo2023latent}, Euler Discrete~\cite{karras2022elucidating}, Flow Match Euler Discrete~\cite{esser2024scaling}), and acceleration methods (LCM~\cite{luo2023latent}, Lightning~\cite{lin2024sdxl}, Hyper-SD~\cite{ren2024hyper}, Flash Diffusion~\cite{chadebec2024flash}), without compromising the visual quality.
Figure~\ref{fig:exp_region} shows a random subset of generation from the experiments in Table~\ref{tab:quality-region-lcm}.
Comparable visual quality from our method is consistent to the quantitative comparisons.

\paragraph{Stabilized Acceleration of Region-Based Generation.}

Next, we evaluate the effectiveness of each stabilization step introduced in Section~\ref{sec:3_method:stabilizing}.
Figure~\ref{fig:problem2} and Table~\ref{tab:abl_bstrap} summarize the result on region-based text-to-image generation from the same setup as Table~\ref{tab:quality-region-lcm}.
Applying each strategy consistently boosts both perceptual quality, measured by FID score~\cite{heusel2017gans}, and text prompt-fidelity, measured by the two CLIP scores~\cite{hessel2021clipscore}.
This reveals that our techniques help alleviating the incompatibility as intended.

\paragraph{Throughput Maximization.}

Table~\ref{tab:exp_speed} compares the effect of throughput optimization.
We have already achieved $\times 9.7$ speed-up by establishing the compatibility with acceleration modules.
This is further enhanced though our \emph{multi-prompt stream batch} architecture.
With low-memory autoencoder~\cite{tinyvae2023} to trade quality off for speed, we could finally achieve 1.57 FPS (0.64 seconds per frame).
This near real-time, sub-second generation speed is a necessary step towards practical applications of generative models.

\paragraph{User Study.}

Finally, we conduct a user study on the methods of Table~\ref{tab:quality-region-lcm}.
Its result, summarized in Table~\ref{tab:user-study}, shows that our method greatly increases generation quality with multiple region-based text prompts.

%-------------------------------------------------------------------------
\begin{table}[tb]
  \caption{%
  Ablation on the effectiveness of our stabilization techniques on the fidelity of region-based generation.
  }
  \label{tab:abl_bstrap}
  \centering
  \vspace{-.5em}
  \resizebox{\linewidth}{!}{%
  \begin{tabular}{@{}lccc@{}}
    \toprule
    Method & FID $\downarrow$ & CLIP$_{\text{fg}}$ $\uparrow$ & CLIP$_{\text{bg}}$ $\uparrow$ \\
    \midrule
    No stabilization & 270.55 & 22.53 & 19.63 \\
    + Latent pre-averaging & 80.64 & 22.80 & 26.95 \\
    + Mask-centering bootstrapping & 79.54 & 23.06 & \textbf{26.72} \\
    + Quantized masks ($\sigma = 4$) & \textbf{78.21} & \textbf{23.08} & \textbf{26.72} \\
  \bottomrule
  \end{tabular}%
  }
\vspace{-.5em}
\end{table}

\begin{table}[tb]
  \caption{%
  Ablations on throughput optimization techniques, measured with a single RTX 2080 Ti.
  Images of $512 \times 512$ are generated from three prompt-mask pairs.
  }
  \label{tab:exp_speed}
  \vspace{-.5em}
  \centering
  \resizebox{\linewidth}{!}{%
  \begin{tabular}{@{}lcc@{}}
    \toprule
    Method & Throughput (FPS) & Relative Speedup \\
    \midrule
    Baseline~\cite{bar2023multidiffusion} & 0.0189 & $\times$1.0 \\
    + Stable Acceleration & 0.183 & $\times$9.7 \\
    + Multi-Prompt Stream Batch & 1.38 & $\times$73.0 \\
    + Tiny AutoEncoder~\cite{tinyvae2023} & \textbf{1.57} & \textbf{$\times$83.1} \\
  \bottomrule
  \end{tabular}%
  }
\vspace{-.5em}
\end{table}
%-------------------------------------------------------------------------
\begin{table}[t] 
\caption{%
  User preference regarding quality of generation.
  }
  \label{tab:user-study}
  \vspace{-0.5em} 
  \centering
  \resizebox{0.75\linewidth}{!}{%
  \begin{tabular}{@{}l|ccc@{}}
    \toprule
    Method & \textbf{Ours} & MD~\cite{bar2023multidiffusion} & MD~\cite{bar2023multidiffusion}+LCM~\cite{luo2023latent}  \\
    \midrule
    Preference & 90.9\% & 9.1\% & 0.0\% \\
  \bottomrule
  \end{tabular}%
  }
  \vspace{-.5em}
\end{table}
%-------------------------------------------------------------------------

\section{Semantic Draw}
\label{sec:6_discussion}
Our real-time interface of \textsc{SemanticDraw} opens up a new paradigm of user-interactive application for image generation.
We discuss the key features of the application.

\paragraph{Concept.}

Responsive region-based text-to-image synthesis enabled by our streaming pipeline allows users to edit their prompt masks similarly to drawing.
Since the standard text encoding by large text encoders (e.g., CLIP) accounts for approximately 40\% of our sub-second runtime (1.57 FPS), caching and reusing these encodings when only mask modifications occur hides this latency and provides \emph{even faster} feedback to users.
This allows them to iteratively refine their commands according to the generated image.
In scenarios where users change text prompts, standard text processing occurs while still maintaining the original sub-second runtime.
This enables users to \emph{paint} with \emph{text prompts} just like they can paint a drawing with colored brushes, hence the name: \textsc{SemanticDraw}.

\begin{figure}[tb]
  \centering
  \begin{subcaptionbox}{
    {Semantic Draw}.
    \label{fig:screenshot:palette}
  }[0.485\linewidth]
  {
    \includegraphics[width=\linewidth]{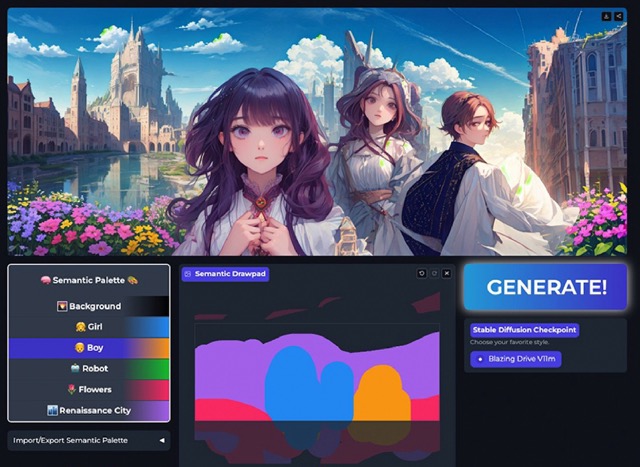}
  }
  \end{subcaptionbox}
  \hfill
  \begin{subcaptionbox}{
    {Streaming Semantic Draw}.
    \label{fig:screenshot:stream}
  }[0.46\linewidth]
  {
    \includegraphics[width=\linewidth]{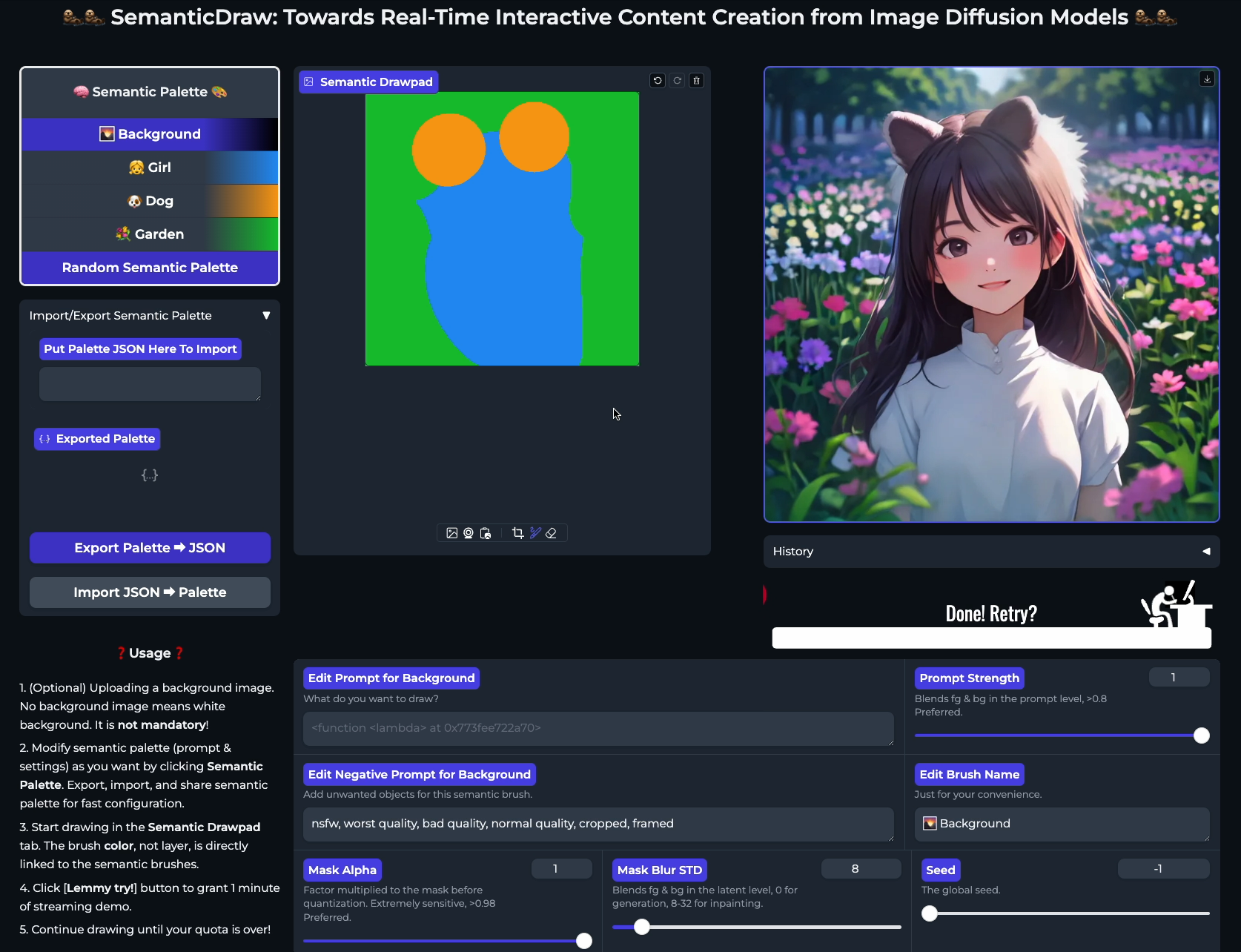}
  }
  \end{subcaptionbox}
  \caption{%
  Screenshot of the sample applications of \textsc{SemanticDraw}.
  After registering prompts and optional background image, the users can create images in real-time by drawing with text prompts.
  We invite the readers to play with the application.
  }
  \label{fig:screenshot}
\vspace{-.5em}
\end{figure}

\paragraph{Sample Application Design.}

We briefly describe our minimal demo application in Figure~\ref{fig:screenshot}.
The application consists of a front-end user interface and a back-end server that runs \textsc{SemanticDraw}.
Each user input is either a modification of the background image, the text prompts, the masks, and the tweakable options for the text prompts and the masks such as mix ratios and blur strengths.
When commanding major changes requiring preprocessing stages, such as a change of prompts or the background, the back-end pipeline is flushed and reinitialized with the newly given context.
Otherwise, the pipeline is repeatedly called to obtain a stream of generated images.
The user first selects the background image and creates a \textit{palette of semantic masks} by entering a pair of positive and negative text prompts.
The user can then draw masks corresponding to the created palette with a familiar brush tool, a shape tool, or a paint tool.
The application automatically generates a stream of synthesized images according to user inputs.
We gently invite the readers to play with our technical demo provided with the official code\footnote{\url{https://github.com/ironjr/semantic-draw}}.

\section{Conclusion}
\label{sec:7_concl}
We proposed \textsc{SemanticDraw}, a new type image content creation where users interactively draw with a brush tool that paints semantic masks to endlessly and continuously create images.
Enabling this application required high generation throughput and well-established compatibility between regional control pipelines and acceleration schedulers.
We devised multi-prompt regional control pipeline that is both scheduler-agnostic and model-agnostic in order to maximize the compatibility.
We further proposed \emph{multi-prompt stream batch} architecture to build a near real-time, highly interactive image content creation system for professional usage.
Our \textsc{SemanticDraw} achieves up to $\times 50$ faster generation of large scale images than the baseline, bringing the latency of multi-prompt irregular-sized generation down to a practically meaningful bounds.

%%%%%%%
\section*{Acknowledgment}
This work was supported in part by the IITP grants [No.2021-0-01343, Artificial Intelligence Graduate School Program (Seoul National University), No. 2021-0-02068, and  No.2023-0-00156], and the NOTIE grant (No. RS-2024-00432410) by the Korean government.

%%%%%%%%% REFERENCES
{
    \small
    \bibliographystyle{ieeenat_fullname}
    \bibliography{main}
}

% ---------------------------------------------------------------
\maketitlesupplementary

\setcounter{section}{0}
\setcounter{table}{0}
\setcounter{figure}{0}
\renewcommand{\thesection}{S\arabic{section}}   
\renewcommand{\thetable}{S\arabic{table}}   
\renewcommand{\thefigure}{S\arabic{figure}}

\renewcommand\thesection{S\arabic{section}}
\renewcommand\thesubsection{\thesection.\arabic{subsection}}
\renewcommand\thefigure{S\arabic{figure}}
\renewcommand\thetable{S\arabic{table}}
\renewcommand\thealgocf{S\arabic{algocf}}
\renewcommand\theequation{S\arabic{equation}}
% ---------------------------------------------------------------

\begin{abstract}
Section~\ref{sec:a_algorithm} shows implementation details of our acceleration methods.
In Section~\ref{sec:b_exp_qual}, additional visual results are shown.
Finally, we provide our demo application as we have promised in our main manuscript.
Our formulation introduces new controllable hyperparameters that users may interact in order to create images that respect their intentions.
Section~\ref{sec:e_app_instruction} demonstrates how our new tool can be used in image content creation.
\end{abstract}

\section{Implementation Details}
\label{sec:a_algorithm}
We begin by providing additional implementation details.

\subsection{Acceleration-Compatible Regional Controls}

\definecolor{importantcolor}{HTML}{2692F3}

Algorithm~\ref{alg:multidiffusion} compares between the the baseline MultiDiffusion~\cite{bar2023multidiffusion} and our stabilized sampling from multiple regionally assigned text prompts introduced in Section~3.2 of the main manuscript.
As we have discussed in Section~3 of the main manuscript, improper placing of the aggregation step and strong interference of its bootstrapping strategy limit the ability to generate visually pleasing images under modern fast inference algorithms~\cite{song2023consistency,luo2023latent,luo2023lcm,lin2024sdxl,ren2024hyper,chadebec2024flash}.
Therefore, we instead focus on changing the bootstrapping stage of line 9-13 and the diffusion update stage of line 14-15 of Algorithm~\ref{alg:multidiffusion} in order to establish compatibility to accelerating diffusion samplers.

The resulting Algorithm~\ref{alg:single} developed in Section~3.2 of the main manuscript achieves this.
The differences between our approach from the baseline inference algorithm are marked with {\color{importantcolor} blue}.
First, in line 10, we change the bootstrapping background color to white.
Having extremely low number of sampling steps (4-5), this bootstrapping background is easily leaked through the final image as seen in Figure~3 of the main manuscript.
We notice that white backgrounds are common in public image datasets on which the diffusion models are trained.
Therefore, changing random background images into white backgrounds alleviate this leakage problem.

Diffusion models have a strong tendency to generate objects at the center of the frame.
This positional bias makes generation from small, off-centered masks difficult especially in the accelerated sampling, where the final structure of generated images are determined at the first two inference steps.
By masking with off-centered masks, the objects under generation are unnaturally cut, leading to defective generations.
Lines 13-14 of Algorithm~\ref{alg:single} are our \emph{mask-centering} stage for bootstrapping to alleviate this problem.
In the first few steps of generation, for each mask-designated object, intermediate latents are masked then shifted to the center of the object bounding box.
This operation enforces the denoising network to focus on each foreground object located at the center of the screen.
Lines 17-19 of Algorithm~\ref{alg:single} undo this centering operation done in lines 13-14.
The separately estimated foreground objects are aggregated into the single scene by shifting them back to their original absolute positions.

Finally, a single reverse diffusion step in line 14 of Algorithm~\ref{alg:multidiffusion} is split into the denoising part in line 16 of Algorithm~\ref{alg:single} and the noise addition part in line 24 of Algorithm~\ref{alg:single}.
As we have discussed with visual example in Figure~3c in the main manuscript, this simple augmentation of the original MultiDiffusion~\cite{bar2023multidiffusion} stabilizes the algorithm to work with fast inference techniques such as LCM-LoRA~\cite{luo2023latent,luo2023lcm}, SDXL-Lightning~\cite{lin2024sdxl}, Hyper-SD~\cite{ren2024hyper}, and Flash Diffusion~\cite{chadebec2024flash}.
Also refer to panorama generation in Figure~\ref{fig:appx:panorama1} where this wrongly placed aggregation after \textsc{Step} operation causing extremely blurry generation under accelerating schedulers~\cite{luo2023latent,luo2023lcm}.
The readers may also consult our submitted code for the implementation of Algorithm~\ref{alg:single}.

\begin{algorithm*}[!htbp]
\definecolor{commentcolor}{HTML}{74985D}
\newcommand\mycommfont[1]{\scriptsize\ttfamily\textcolor{commentcolor}{#1}}\SetCommentSty{mycommfont}
\DontPrintSemicolon
   \caption{\small Baseline~\cite{bar2023multidiffusion}.}
   \label{alg:multidiffusion}
  \SetAlgoLined
  \KwIn{%
  a diffusion model $\boldsymbol{\epsilon}_{\theta}\,$,
  a latent autoencoder $(\texttt{enc}, \texttt{dec})\,$,
  prompt embeddings $\boldsymbol{y}_{1:p}\,$,
  masks $\boldsymbol{w}_{1:p}\,$,
  timesteps $\boldsymbol{t} = t_{1:n}\,$,
  the output size $(H', W')\,$,
  the tile size $(H, W)\,$,
  an inference algorithm $\textsc{Step}\,$,
  a noise schedule $\alpha\,$,
  the number of bootstrapping steps $n_{\text{bstrap}}\,$.
  }
  \KwOut{%
  An image $\boldsymbol{I}$ of designated size $(8H', 8W')$ generated from multiple text-mask pairs.
  }
  $\boldsymbol{x}'_{t_{n}} \sim \mathcal{N}(0, 1)^{H' \times W' \times D}$ \tcp*{sample the initial latent}
  $\{ \mathcal{T}_{1}, \ldots, \mathcal{T}_{m} \} \subset \{ (h_{\text{t}}, h_{\text{b}}, w_{\text{l}}, w_{\text{r}}): 0 \leq h_{\text{t}} < h_{\text{b}} \leq H', 0 \leq w_{\text{l}} < w_{\text{r}} \leq W' \}$ \\ \tcp*{get a set of overlapping tiles}
  \For{$i \gets n$ \KwTo $1$}{%
    $\tilde{\boldsymbol{x}} \gets \boldsymbol{0} \in \mathbb{R}^{H' \times W' \times D}$ \tcp*{placeholder for the next step latent}
    $\tilde{\boldsymbol{w}} \gets \boldsymbol{0} \in \mathbb{R}^{H' \times W'}$ \tcp*{placeholder for the next step mask weights}
    \For{$j \gets 1$ \KwTo $m$}{%
      $\bar{\boldsymbol{x}}_{1:p} \gets \texttt{repeat}(\texttt{crop}(\boldsymbol{x}_{t_{i}}, \mathcal{T}_{j}), p)$ \tcp*{get a cropped intermediate latent tile}
      $\bar{\boldsymbol{w}}_{1:p} \gets \texttt{crop}(\boldsymbol{w}_{1:p}, \mathcal{T}_{j})$ \tcp*{get cropped mask tiles}
      \If{$i \leq n_{\text{bstrap}}$}{%
        $\boldsymbol{x}_{\text{bg}} \gets \texttt{enc}(c \boldsymbol{1})\,$, where $c \sim \mathcal{U}(0, 1)^{3}$ \tcp*{get a uniform color background}
        $\boldsymbol{x}_{\text{bg}} \gets \sqrt{\alpha(t_{i})} \boldsymbol{x}_{\text{bg}} \sqrt{1 - \alpha(t_{i})} \boldsymbol{\epsilon}\,$, where $\boldsymbol{\epsilon} \sim \mathcal{N}(0, 1)^{H \times W \times D}$ \tcp*{add noise to the background for mixing}
        $\bar{\boldsymbol{x}}_{1:p} \gets \bar{\boldsymbol{w}}_{1:p} \odot \bar{\boldsymbol{x}}_{1:p} + (\boldsymbol{1} - \bar{\boldsymbol{w}}_{1:p}) \odot \boldsymbol{x}_{\text{bg}}$ \tcp*{bootstrap by treating as multiple single-instance images}
      }
      $\bar{\boldsymbol{x}}_{1:p} \gets \textsc{Step} ( \bar{\boldsymbol{x}}_{1:p}, \boldsymbol{y}_{1:p}, i ; \boldsymbol{\epsilon}_{\theta}, \alpha, \boldsymbol{t} )$ \tcp*{prompt-wise batched diffusion update}
      $\tilde{\boldsymbol{x}}[\mathcal{T}_{j}] \gets \tilde{\boldsymbol{x}}[\mathcal{T}_{j}] + \sum_{k = 1}^{p} \bar{\boldsymbol{w}}_{k} \odot \bar{\boldsymbol{x}}_{k}$ \tcp*{aggregation by averaging}
      $\tilde{\boldsymbol{w}}[\mathcal{T}_{j}] \gets \tilde{\boldsymbol{w}}[\mathcal{T}_{j}] + \sum_{k = 1}^{p} \bar{\boldsymbol{w}}_{k}$ \tcp*{total weights for normalization}
    }
    $\boldsymbol{x}_{t_{i - 1}} \gets \tilde{\boldsymbol{x}} \odot \tilde{\boldsymbol{w}}^{-1}$ \tcp*{reverse diffusion step}
  }
  $\boldsymbol{I} \gets \texttt{dec}(\boldsymbol{x}_{t_{1}})$ \tcp*{decode latents to get an image}
\end{algorithm*}
\begin{algorithm*}[!htbp] %[!htbp]
\definecolor{importantcolor}{HTML}{2692F3}
\definecolor{commentcolor}{HTML}{74985D}
\newcommand\mycommfont[1]{\scriptsize\ttfamily\textcolor{commentcolor}{#1}}
\SetCommentSty{mycommfont}
\DontPrintSemicolon
   \caption{\small \textsc{SemanticDraw} pipeline of Section~3.2.}
   \label{alg:single}
  \SetAlgoLined
  \KwIn{%
  a diffusion model $\boldsymbol{\epsilon}_{\theta}\,$,
  a latent autoencoder $(\texttt{enc}, \texttt{dec})\,$,
  prompt embeddings $\boldsymbol{y}_{1:p}\,$,
  {\color{importantcolor} quantized masks $\boldsymbol{w}^{(t_{1:n})}_{1:p}\,$},
  timesteps $\boldsymbol{t} = t_{1:n}\,$,
  the output size $(H', W')\,$,
  a noise schedule $\alpha$ {\color{importantcolor} and $\eta\,$},
  the tile size $(H, W)\,$,
  {\color{importantcolor} an inference algorithm $\textsc{StepExceptNoise}\,$},
  the number of bootstrapping steps $n_{\text{bstrap}}\,$.
  }
  \KwOut{%
  An image $\boldsymbol{I}$ of designated size $(8H', 8W')$ generated from multiple text-mask pairs.
  }
  $\boldsymbol{x}'_{t_{n}} \sim \mathcal{N}(0, 1)^{H' \times W' \times D}$ \\
  $\{ \mathcal{T}_{1}, \ldots, \mathcal{T}_{m} \} \subset \{ (h_{\text{t}}, h_{\text{b}}, w_{\text{l}}, w_{\text{r}}): 0 \leq h_{\text{t}} < h_{\text{b}} \leq H', 0 \leq w_{\text{l}} < w_{\text{r}} \leq W' \}$ \\
  \For{$i \gets n$ \KwTo $1$}{%
    $\tilde{\boldsymbol{x}} \gets \boldsymbol{0} \in \mathbb{R}^{H' \times W' \times D}$ \\
    $\tilde{\boldsymbol{w}} \gets \boldsymbol{0} \in \mathbb{R}^{H' \times W'}$ \\
    \For{$j \gets 1$ \KwTo $m$}{%
      $\bar{\boldsymbol{x}}_{1:p} \gets \texttt{repeat}(\texttt{crop}(\boldsymbol{x}_{t_{i}}, \mathcal{T}_{j}), p)$ \\
      {\color{importantcolor} $\bar{\boldsymbol{w}}_{1:p}^{(t_{i})} \gets \texttt{crop}(\boldsymbol{w}_{1:p}^{(t_{i})}, \mathcal{T}_{j})$} \tcp*{use different quantized masks for each timestep}
      \If{$i \leq n_{\text{bstrap}}$}{%
        {\color{importantcolor} $\boldsymbol{x}_{\text{bg}} \gets \texttt{enc}(\boldsymbol{1})\,$} \tcp*{get a white color background}
        $\boldsymbol{x}_{\text{bg}} \gets \sqrt{\alpha(t_{i})} \boldsymbol{x}_{\text{bg}} \sqrt{1 - \alpha(t_{i})} \boldsymbol{\epsilon}\,$, where $\boldsymbol{\epsilon} \sim \mathcal{N}(0, 1)^{H \times W \times D}$ \\
        $\bar{\boldsymbol{x}}_{1:p} \gets \bar{\boldsymbol{w}}_{1:p} \odot \bar{\boldsymbol{x}}_{1:p} + (\boldsymbol{1} - \bar{\boldsymbol{w}}_{1:p}) \odot \boldsymbol{x}_{\text{bg}}$ \\
        {\color{importantcolor} $\boldsymbol{u}_{1:p} \gets \texttt{get\_bounding\_box\_centers}(\bar{\boldsymbol{w}}_{1:p}) \in \mathbb{R}^{p \times 2}$} \tcp*{get the bounding box center of each mask}
        {\color{importantcolor} $\bar{\boldsymbol{x}}_{1:p} \gets \texttt{roll\_by\_coordinates}(\bar{\boldsymbol{x}}_{1:p}, \boldsymbol{u}_{1:p})$} \tcp*{center foregrounds to their mask centers}
      }
      {\color{importantcolor} $\bar{\boldsymbol{x}}_{1:p} \gets \textsc{StepExceptNoise} ( \bar{\boldsymbol{x}}_{1:p}, \boldsymbol{y}_{1:p}, i ; \boldsymbol{\epsilon}_{\theta}, \alpha, \boldsymbol{t} )$} \tcp*{pre-averaging}
      {\color{importantcolor} 
      \If{$i \leq n_{\text{bstrap}}$}{%
        $\bar{\boldsymbol{x}}_{1:p} \gets \texttt{roll\_by\_coordinates}(\bar{\boldsymbol{x}}_{1:p}, -\boldsymbol{u}_{1:p})$ \tcp*{restore from centering}
      }
      }
      $\tilde{\boldsymbol{x}}[\mathcal{T}_{j}] \gets \tilde{\boldsymbol{x}}[\mathcal{T}_{j}] + \sum_{k = 1}^{p} \bar{\boldsymbol{w}}_{k} \odot \bar{\boldsymbol{x}}_{k}$ \\
      $\tilde{\boldsymbol{w}}[\mathcal{T}_{j}] \gets \tilde{\boldsymbol{w}}[\mathcal{T}_{j}] + \sum_{k = 1}^{p} \bar{\boldsymbol{w}}_{k}$ \\
    }
    $\boldsymbol{x}_{t_{i - 1}} \gets \tilde{\boldsymbol{x}} \odot \tilde{\boldsymbol{w}}^{-1}$ \\
    {\color{importantcolor} $\boldsymbol{x}_{t_{i - 1}} \gets \boldsymbol{x}_{t_{i - 1}} + \eta_{t_{i - 1}} \boldsymbol{\epsilon}\,$, where $\boldsymbol{\epsilon} \sim \mathcal{N}(0, 1)^{H \times W \times D}$} \tcp*{post-addition of noise}
  }
  $\boldsymbol{I} \gets \texttt{dec}(\boldsymbol{x}_{t_{1}})$
\end{algorithm*}

\begin{figure}[tb]
\definecolor{unet}{HTML}{CFE6D7}
\definecolor{decoder}{HTML}{FFD8Cf}
\newcommand{\enccell}{%
\begin{tikzpicture}[scale=.6]%
\draw[black, opacity=.5, line width=0.5] (0,0) -- ++(90:.5) node[midway] (l) {} -- ++(0:.5) node[midway] (t) {} -- ++(-90:.5) node[midway] (r) {} -- cycle node[midway] (b) {};
\draw[opacity=0] (l) -- (r) node[midway, opacity=1] {E};
\end{tikzpicture}
\hspace{-.95em}
}
\newcommand{\deccell}{%
\hspace{-1.2em}
\begin{tikzpicture}[scale=.7]%
\draw[black, opacity=1, fill=decoder, line width=0.5] (0,0) -- ++(90:.35) node[midway] (l) {} -- ++(10:.5) node[midway] (t) {} -- ++(-90:.5) node[midway] (r) {} -- cycle node[midway] (b) {};
\draw[opacity=0] (l) -- (r) node[midway, opacity=1] {D};
\end{tikzpicture}
}
\newcommand{\batchcell}[1]{%
\begin{tikzpicture}[scale=.7]%
\draw[black, opacity=1, fill=unet, line width=0.5] (0,0) -- ++(90:.5) node[midway] (t3) {} -- ++(-10:.25) node(t1) {} -- ++(10:.25) -- ++(-90:.5) node[midway] (t4) {} -- ++(170:.25) node(t2) {} -- cycle;
\draw[opacity=0] (t1) -- (t2) node[midway, opacity=1] {U{\tiny $_{#1}$}};
\end{tikzpicture}
}
\newcommand{\vv}{\\[-1.35em]}
\resizebox{\linewidth}{!}{%
  {
  \scriptsize
  \myfont
    \begin{tabular}{l|l:l:l:l:l:l}
         User Cmd & \hfill t \hfill \phantom{\hspace{-.5em}.} & \hfill t + 1 \hfill \phantom{\hspace{-.5em}.} & \hfill t + 2 \hfill \phantom{\hspace{-.5em}.} & \hfill t + 3 \hfill \phantom{\hspace{-.5em}.} & \hfill t + 4 \hfill \phantom{\hspace{-.5em}.} & \hfill t + 5 \hfill \phantom{\hspace{-.5em}.} \\
         \midrule
         1: initialize & \enccell \batchcell{0} & \batchcell{1} & \batchcell{2} & \batchcell{3} & \batchcell{4} \deccell &\\
         2: no-op & & \batchcell{0} & \batchcell{1} & \batchcell{2} & \batchcell{3} & \batchcell{4} \deccell \\
         3: draw mask & & & \batchcell{0} & \batchcell{1} & \batchcell{2} & \batchcell{3} \\
         4: edit prompt & & & & \enccell \batchcell{0} & \batchcell{1} & \batchcell{2} \\
         5: edit mask & & & & & \batchcell{0} & \batchcell{1} \\
         6: no-op & & & & & & \batchcell{0} \\
        \bottomrule
    \end{tabular}
  }
}
  \caption{%
  Example execution process of \textit{Multi-Prompt Stream Batch} pipeline of \textsc{SemanticDraw}.
  By aggregating latents at different timesteps a single batch, we can maximize throughput by hiding the latency.
  }
  \label{fig:stream}
\end{figure}

%-------------------------------------------------------------------------
\begin{figure*}[t]
  \centering
    \subfloat[Masks]{\includegraphics[width=0.198\linewidth]{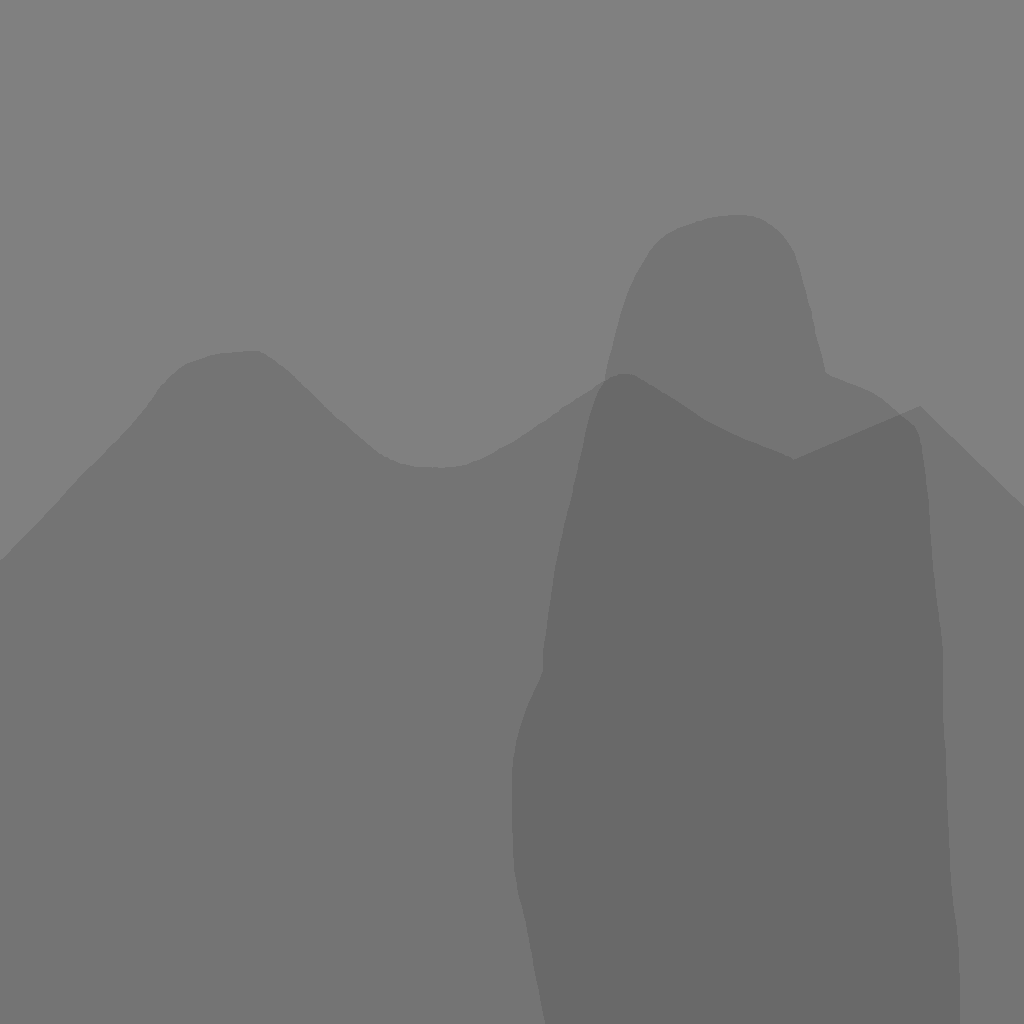}}
    \hfill
    \subfloat[1 step]{\includegraphics[width=0.198\linewidth]{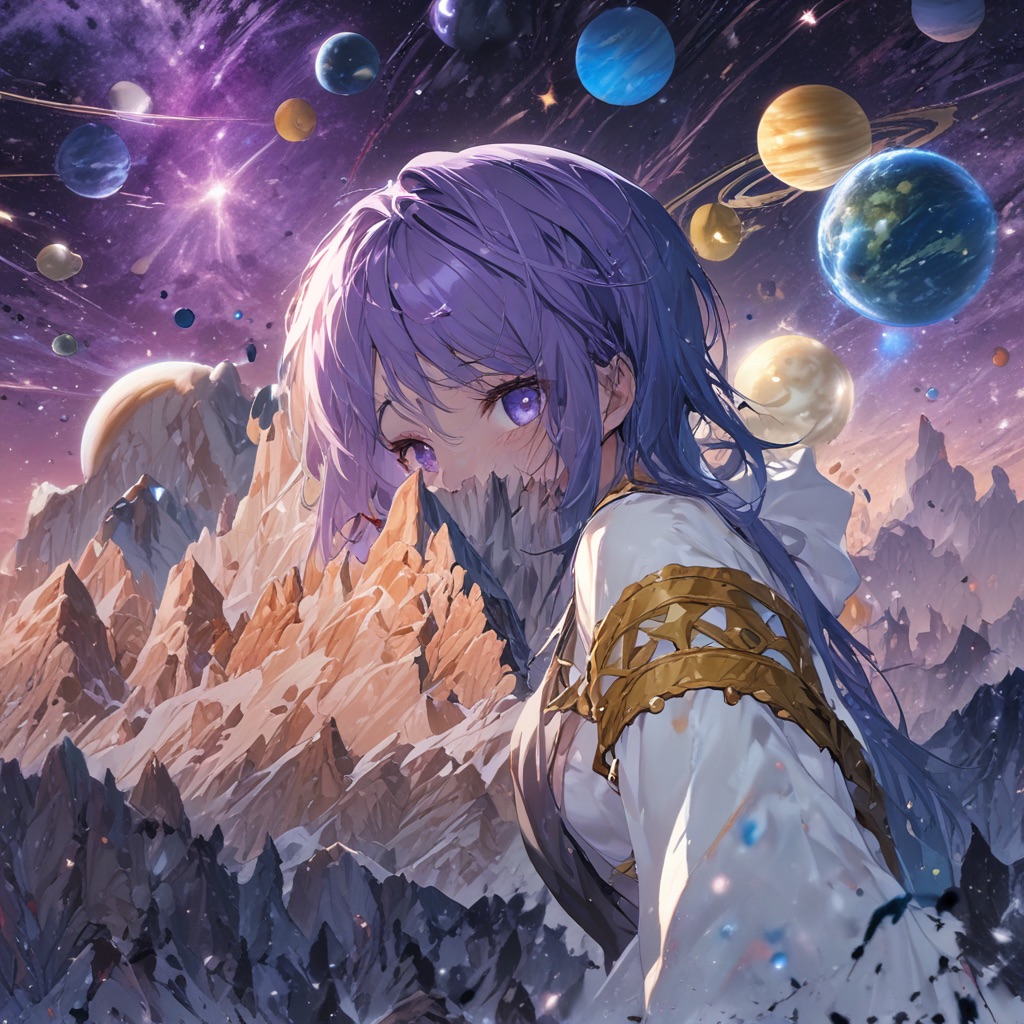}}
    \hfill
    \subfloat[\textbf{2 steps}]{\includegraphics[width=0.198\linewidth]{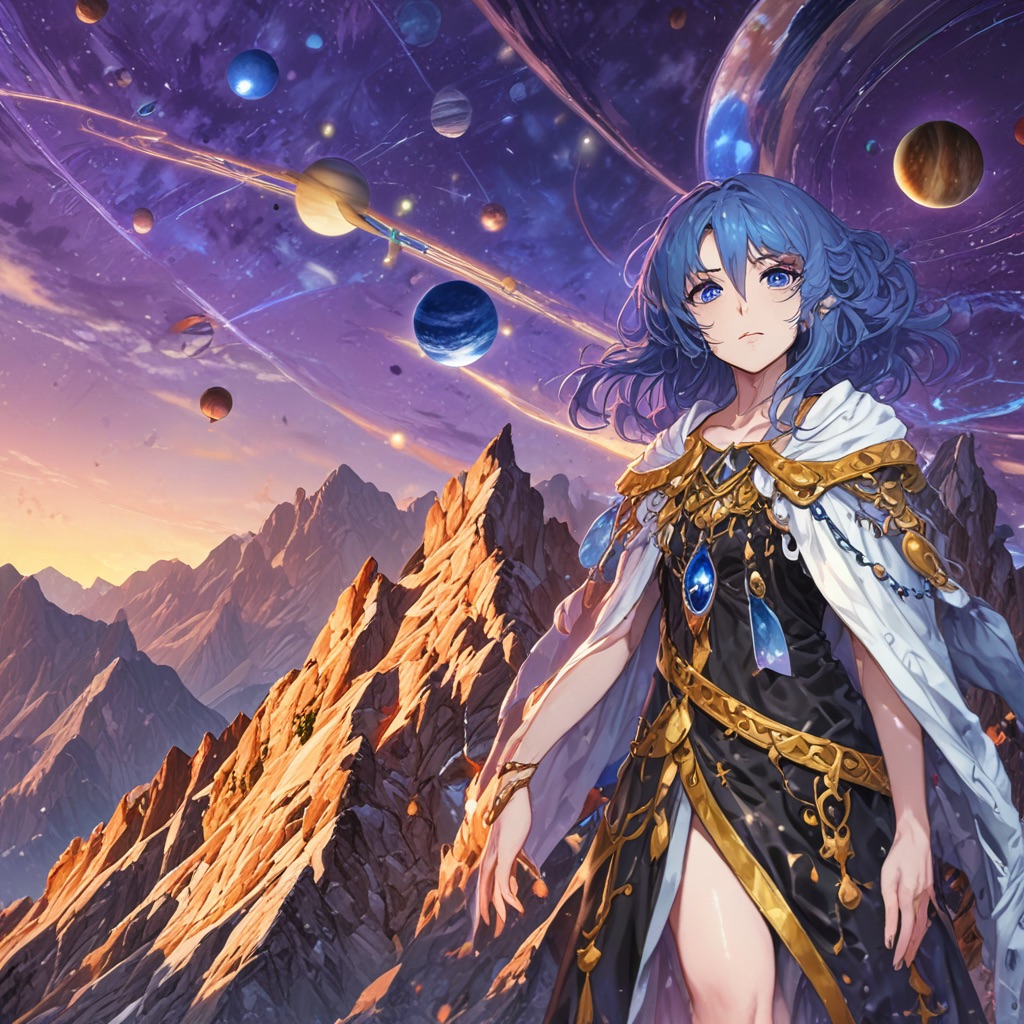}}
    \hfill
    \subfloat[\textbf{3 steps}]{\includegraphics[width=0.198\linewidth]{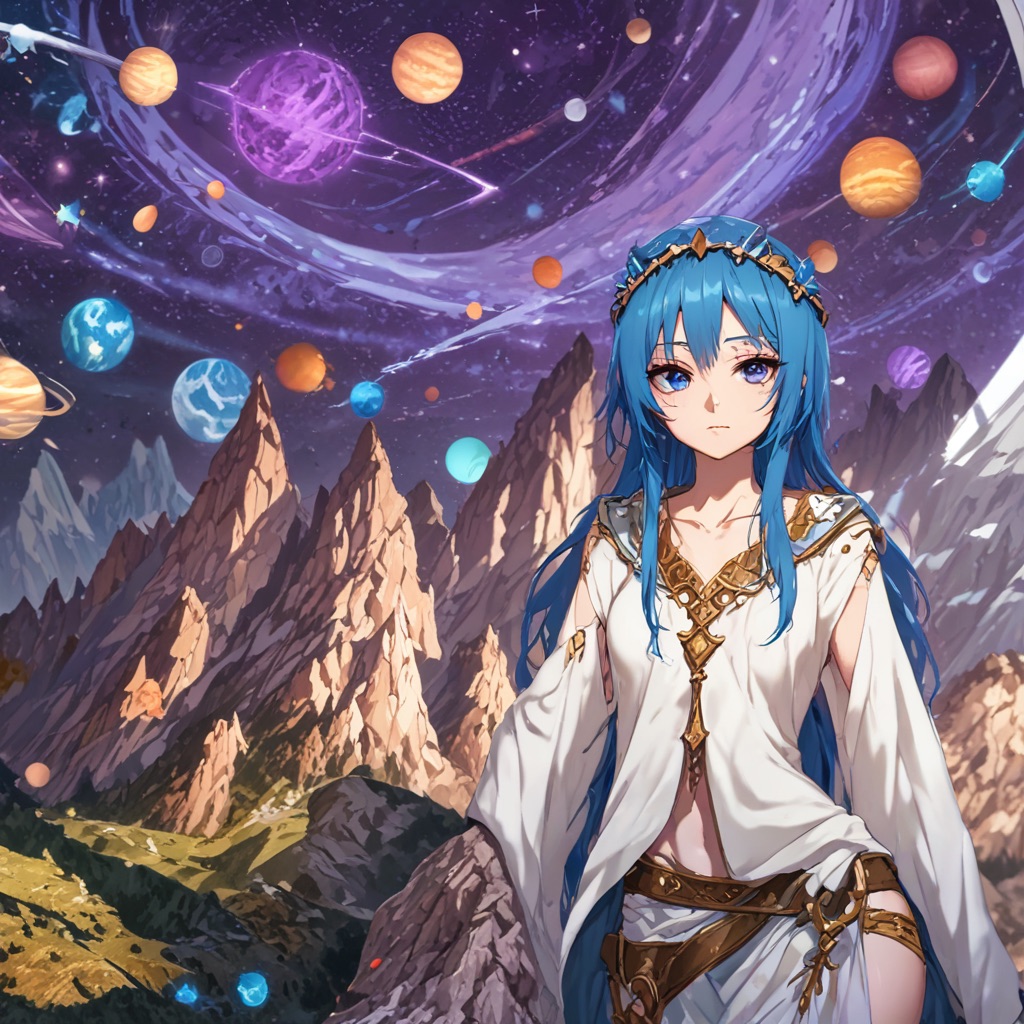}}
    \hfill
    \subfloat[4 steps]{\includegraphics[width=0.198\linewidth]{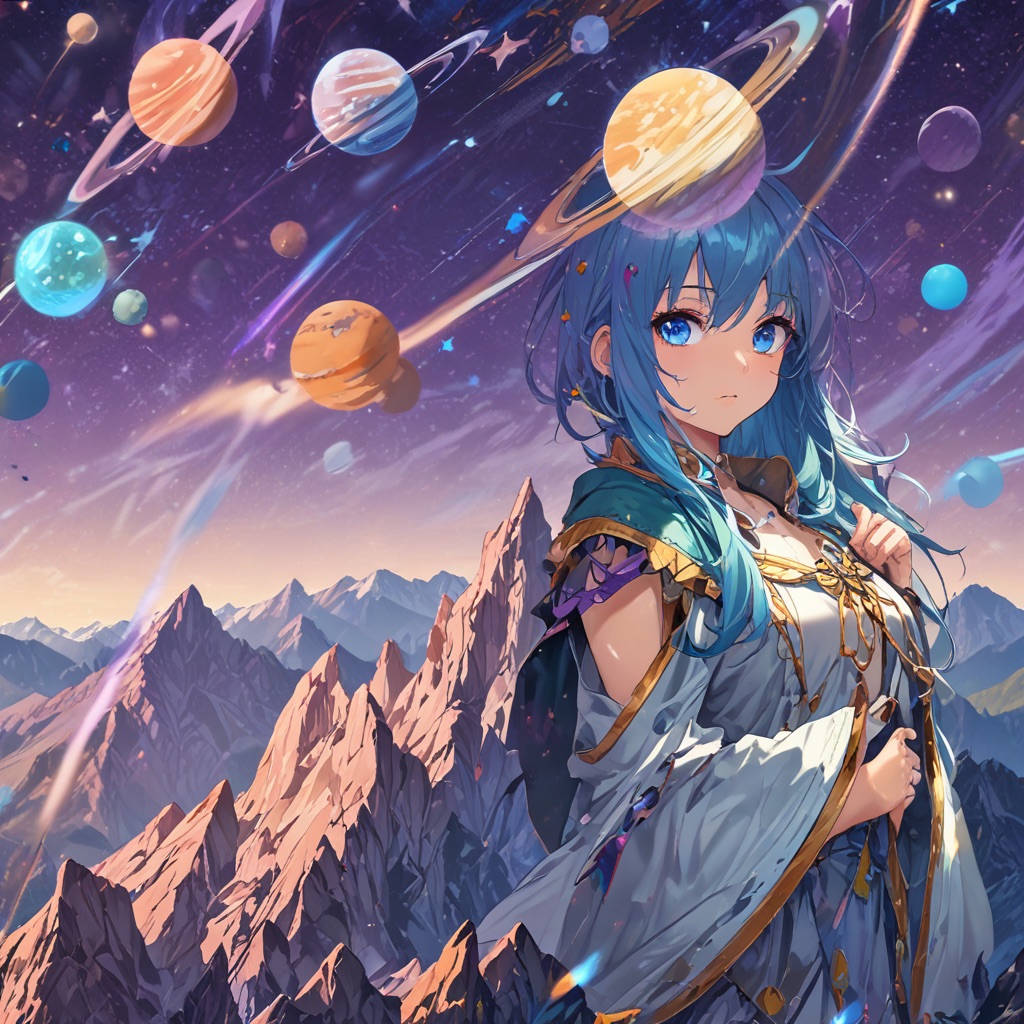}}
    \vspace{-.5em}
  \caption{%
  The number of centering steps effectively trades off centered-bias against overall harmony.
  Composition, harmony, and mask-obedience are achieved in the sweet spot of 2-3 steps.
  }
  \label{fig:tradeoff}
\vspace{-.5em}
\end{figure*}
%-------------------------------------------------------------------------

\begin{figure}[tb]
\newcommand{\figurewidth}{.475\linewidth}
\newcommand{\h}{.475\linewidth}
\newcommand{\vv}{\vspace*{-0.0mm}}
\newcommand{\vvv}{\vspace*{-0.5mm}}
\definecolor{p1color}{HTML}{F89E12}
\definecolor{p2color}{HTML}{F92F6C}
  \centering
  {\myfont
  \makebox[\h][c]{\hspace{-0.\linewidth}\scriptsize{
Background: \textit{``A brick wall''}, \,
  {\color{p2color} Red}: \textit{``A moss''}
  }}\vv\\
\includegraphics[height=\h,width=\figurewidth]{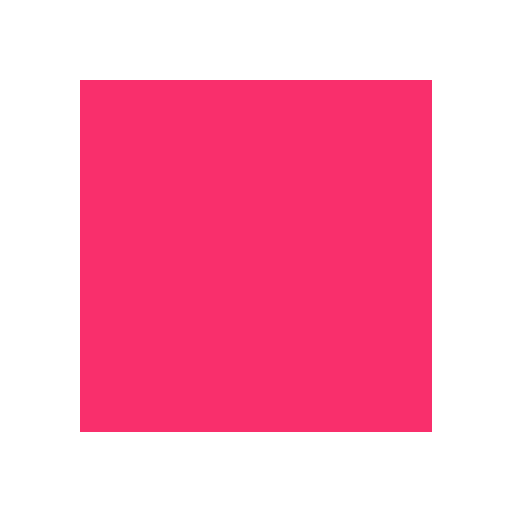}
\includegraphics[height=\h,width=\figurewidth]{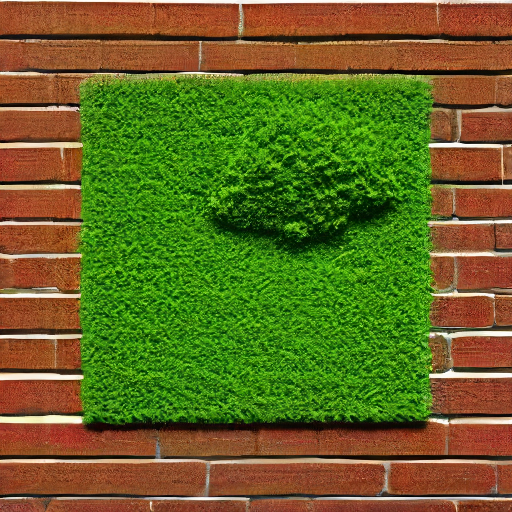}\vvv\\
  \makebox[\h][c]{\hspace{-0.\linewidth}\footnotesize{\textbf{(a)} Prompt mask.}}
  \makebox[\h][c]{\hspace{-0.\linewidth}\footnotesize{\textbf{(b)} $\sigma = 0\,$, \textit{i.e.}, no QMask.}}
\hfill\vv
\includegraphics[height=\h,width=\figurewidth]{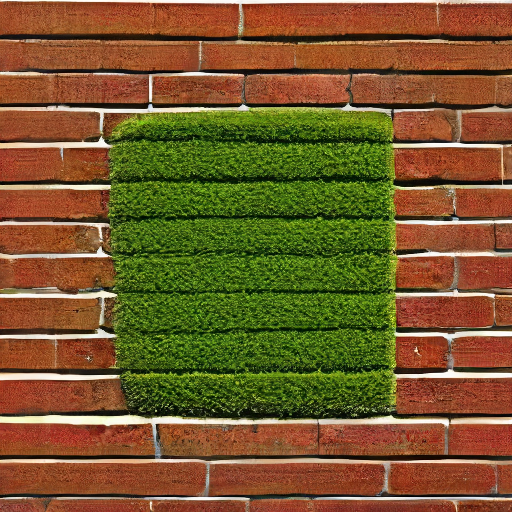}\hfill
\includegraphics[height=\h,width=\figurewidth]{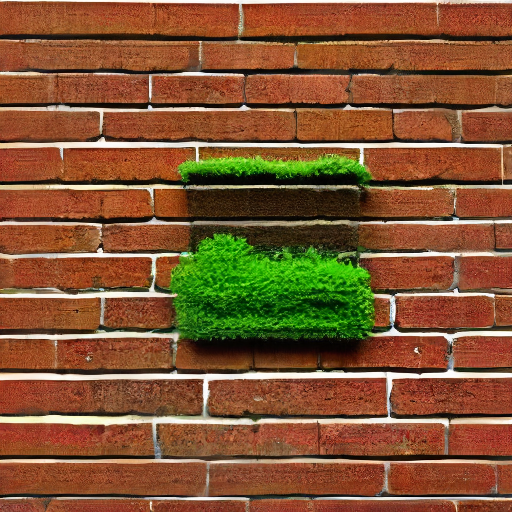}\vvv\\
\hfill
  \makebox[\h][c]{\hspace{-0.\linewidth}\footnotesize{\textbf{(c)} $\sigma = 16\,$.}}\hfill
  \makebox[\h][c]{\hspace{-0.\linewidth}\footnotesize{\textbf{(d)} $\sigma = 32\,$.}}%
\hfill
\vspace{-.1em}
    }
  \caption{%
  Effect of the standard deviation in mask smoothing.
  }
  \label{fig:maskquant:result}
  \vspace{-.5em}
\end{figure}

\subsection{Streaming Pipeline Execution}

Extending Figure~4b of the main manuscript, Figure~\ref{fig:stream} elaborates on the pipelined execution from our \textit{multi-prompt stream batch} architecture for near real-time generation from multiple regionally assigned text prompts.
We have empirically found that the text and image encoders for popular diffusion models take significantly longer latency than the denoising network.
Assuming that users change text prompts and background images less frequently than they change the areas occupied by each semantic masks, such latency can be hidden under the high-throughput streaming generation of images.
Moreover, mask processing takes almost negligible latency compared to image generation or text encoding.
In other words, drawing with semantic masks of pre-encoded text prompts do not affect the generation speed, allowing users to almost seamlessly interact with the generation pipeline by friendly drawing interface.
This user interface of our drawing-based interactive content creation is the same as commercial drawing software with brush tools.
The only difference is that our brush tools apply semantic masks instead of colors or patterns.
This similarity opens up a novel application for diffusion models, \ie, \textsc{SemanticDraw}.

\begin{figure*}[tb]
\newcommand{\figurewidth}{.96\linewidth}
\newcommand{\h}{.384\linewidth}
\newcommand{\hh}{2.5mm}
\newcommand{\vv}{\vspace*{-0.20mm}}
\definecolor{p1color}{HTML}{16C232}
\definecolor{p2color}{HTML}{F92F6C}
\definecolor{p3color}{HTML}{92C6EC}
\definecolor{p4color}{HTML}{FECAC0}
\definecolor{p5color}{HTML}{AC6AEB}
\definecolor{p6color}{HTML}{2692F3}
\definecolor{p7color}{HTML}{F89E12}
\definecolor{p8color}{HTML}{92C62C}
  \centering
{\myfont
    \makebox[\hh]{\rotatebox[origin=l]{90}{\makebox[\h][c]{\hspace{-0.\linewidth}\footnotesize{Ours, Mask Overlay}}}}\hspace{0.mm}%
    \includegraphics[width=\figurewidth]{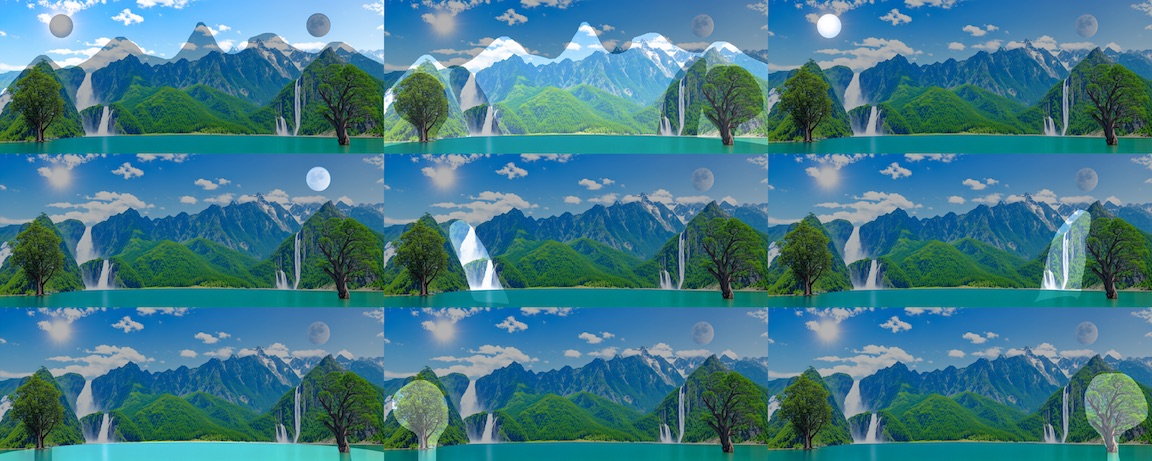}\\[0.3em]
}
{\myfont
  \hfill
  \parbox{.99\linewidth}{\scriptsize{%
  \textbf{Image prompt (row, column):}
  Background (1, 1): \textit{``Clear deep blue sky''}, \,
  {\color{p1color} Green (1, 2)}: \textit{``Summer mountains''}, \,
  {\color{p2color} Red (1, 3)}: \textit{``The Sun''}, \,
  {\color{p3color} Pale Blue (2, 1)}: \textit{``The Moon''}, \,
  {\color{p4color} Light Orange (2, 2)}: \textit{``A giant waterfall''}, \,
  {\color{p5color} Purple (2, 3)}: \textit{``A giant waterfall''}, \,
  {\color{p6color} Blue (3, 1)}: \textit{``Clean deep blue lake''}, \,
  {\color{p7color} Orange (3, 2)}: \textit{``A large tree''}, \,
  {\color{p8color} Light Green (3, 3)}: \textit{``A large tree''}
  }}
}
  \captionof{figure}{%
  Mask overlay images of the generation result in Figure~2 of the main manuscript.
  Generation by our \textsc{SemanticDraw} not only achieves high speed of convergence, but also high mask fidelity in the large-size region-based text-to-image synthesis, compared to the baseline MultiDiffusion~\cite{bar2023multidiffusion}.
  Each cell shows how each mask (including the background one) maps to each generated region of the image, as described in the label below.
  Note that we have \textit{not} provided any additional color or structural control other than our \emph{semantic palette}, which is simply pairs of text prompts and binary masks.
  }
  \label{fig:appx:figure_one_overlay}
\end{figure*}

\subsection{Controlling Fidelity-Harmony Trade-off}
\label{sec:a_alg:tradeoff}

As we have mentioned in Section~3.2, accelerated samplers involving 5 or few steps like in our case rely heavily on the first few steps of inference in determining the structure of the image.
Many diffusion models are trained using \emph{natural} images that place their objects of interest at the center of the canvas.
This makes these diffusion models generate all of their prompt-guided objects at the center of the canvas.
Cropping by masks occasionally leads to destruction of such objects.
Mask-centering bootstrapping is devised in order to alleviate this problem.
However, applying bootstrapping from beginning to the end causes another problem of disharmony in the overall image with multiple region-based prompts.
This can be seen in Figure~\ref{fig:tradeoff}e, where the girls' upper part of head is unnaturally cut.
This problem also caused by the acceleration.
Unlike gradual generation over tens of inference steps, in our accelerated scenario, the later inference steps are responsible for both high quality texture generation \textit{and} boundary creation.
Those quickly generated model-generated boundaries do not align well with the user-given mask inputs, creating unnatural cuts after merging with other prompt-guided subsections of the creation.
We, therefore, provide a simple control handle that trades off mask-fidelity against overall harmony: the number of mask-centering steps in the bootstrapping stage.
The effect of this control handle can be seen in Figure~\ref{fig:tradeoff}.
We have empirically found that 1-3 steps work best, and we have used 2 steps throughout this work.

\subsection{Mask Quantization}
\label{sec:a_alg:maskquant}

To increase harmonization within a created image, we have introduced mask quantization as our final piece of the puzzle in Section~3.2 of the main manuscript.
Mask quantization allows smooth masks with controllable smoothness that resemble soft brush tools in common drawing software.
Therefore, this stage not only increases image fidelity but also enhances user experience in our \textsc{SemanticDraw} application.
This section explains additional technical details of mask quantization.

As Figure~5 of the main manuscript shows, the mask smoothing is an optional preprocessing procedure before generation.
Once users provide a set of masks corresponding to a set of text prompts they want to draw, the binary masks are smoothened with a low-pass filter such as Gaussian blurs.
In order to perform masking with these continuous masks for discrete denoising steps of the accelerated schedulers~\cite{luo2023latent,luo2023lcm,lin2024sdxl,ren2024hyper,chadebec2024flash}, we create a set of binary masks from each of the continuous masks by thresholding with the noise levels predefined by the diffusion scheduler.
For example, Figure~5 of the main manuscript shows five noise levels actually used in generating the results in the main manuscript and throughout this Supplementary Material.
The resulting set of binary masks have monotonically increasing sizes as the corresponding noise levels become lower.
Note that we can interpret a noise level of each generating step as a magnitude of uncertainty during the reverse diffusion process.
Since the boundary of an object is fuzzier than the center of the object of prescribed masked region, the more uncertain boundary regions can be sampled only during the few latest steps where detailed textures dominant over structural development.
Therefore, a natural way of applying these binary masks is in the order of increasing size.
By applying each generated binary mask at the timestep with corresponding noise level, we effectively enlarge the size of the mask of a foreground text prompt as we proceed on the generative denoising steps.

The blurring and quantization of the binary masks have a nice interpretation of a \emph{rough sketch}.
In many cases where users prescribe masks to query for multi-object generation, the exact boundary locations for the best visual construction of an image are not known \textit{a priori}.
In other words, human creation of arts almost always starts with rough sketches.
We can increase or decrease the standard deviation of the blur to control the roughness of the sketch, \ie, the certainty of our designation on the boundary.
This additional control knob is effective in creating AI-driven arts which inherently exploits high randomness in practice.
For reference, Figure~\ref{fig:maskquant:result} shows the effect of increasing the blurriness at the mask proprocessing step.
As the standard deviation of the mask blur increases from 0 to 32, the moss, the content of the mask, {gradually shrinks and semantically blurred} with the brick wall, the background content.
As our supplementary code show, this \textit{semantic mixing effect} of mask blurring and quantization is helpful to harmonize contents in generative editing tasks, \ie, inpainting, where background images are predefined and not fully masked out during generation.

%-------------------------------------------------------------------------
\begin{figure*}[t]
  \centering
  \vspace{-0.5em}
    \subfloat[Masks]{\includegraphics[width=0.245\linewidth]{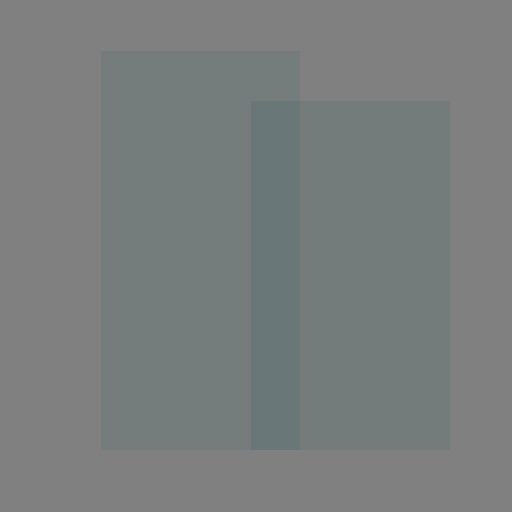}}
    \hfill
    \subfloat[LRDiff (45s)]{\includegraphics[width=0.245\linewidth]{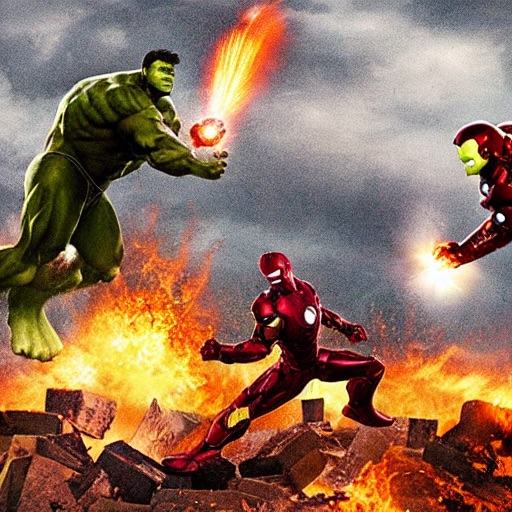}}
    \hfill
    \subfloat[(b)+LCM (1s)]{\includegraphics[width=0.245\linewidth]{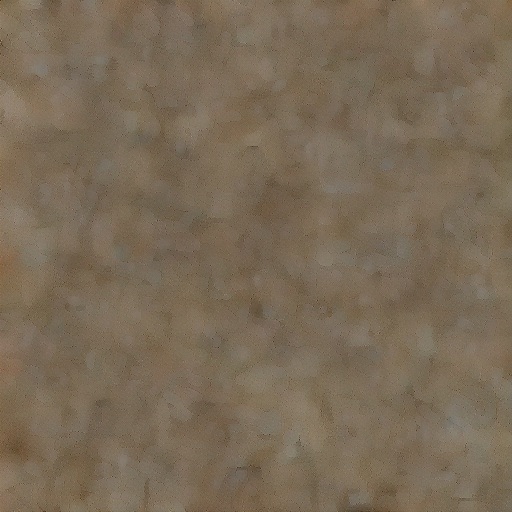}}
    \hfill
    \subfloat[\textbf{Ours} (1s)]{\includegraphics[width=0.245\linewidth]{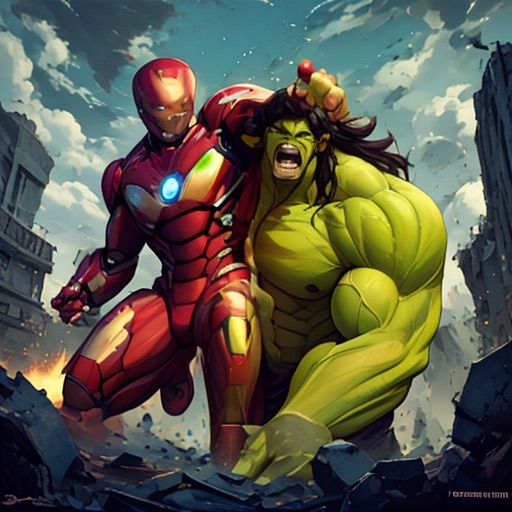}}
    \vspace{-.5em}
  \caption{%
  Qualitative comparison between LRDiff+LCM and ours.
  Background prompt: ``Iron Man and Hulk stand amidst the ruins, engaged in a fierce battle with each other.''
  Left box prompt: ``Iron-man''
  Right box prompt: ``Hulk''
  }
  \label{fig:lrdiff}
\vspace{-.5em}
\end{figure*}
%-------------------------------------------------------------------------

\section{More Results}
\label{sec:b_exp_qual}

In this section, we provide additional visual comparison results between baseline MultiDiffusion~\cite{bar2023multidiffusion}, a simple application of acceleration modules~\cite{luo2023latent,luo2023lcm} to the baseline, and our stabilized Algorithm~\ref{alg:single}.
We show that our algorithm is capable of generating large-scale images from multiple regional prompts with a single commercial off-the-shelf graphics card, \textit{e.g.}, an RTX 2080 Ti GPU.

\begin{figure*}[!htbp]
\newcommand{\h}{\hspace{0.1em}}
\newcommand{\figwidth}{0.204\linewidth}
\newcommand{\figheighta}{0.136\linewidth}
\newcommand{\figheightb}{0.204\linewidth}
\newcommand{\figheight}{0.306\linewidth}
\newcommand{\hh}{20mm}
\newcommand{\vv}{\vspace*{-0.00mm}}
\definecolor{p1color}{HTML}{F89E12}
\definecolor{p2color}{HTML}{F92F6C}
\definecolor{p3color}{HTML}{2692F3}
\centering
{\myfont
  \makebox[\hh][c]{\hspace{.05\linewidth}\scriptsize{%
  Background: \textit{``A cinematic photo of a sunset''}, \,
  {\color{p1color} Yellow}: \textit{``An abandoned castle wall''}, \,
  {\color{p2color} Red}: \textit{``A photo of Alps''}, \,
  {\color{p3color} Blue}: \textit{``A daisy field''}
  }}\vv\\
}
    \subfloat{\includegraphics[width=\figwidth,height=\figheighta]{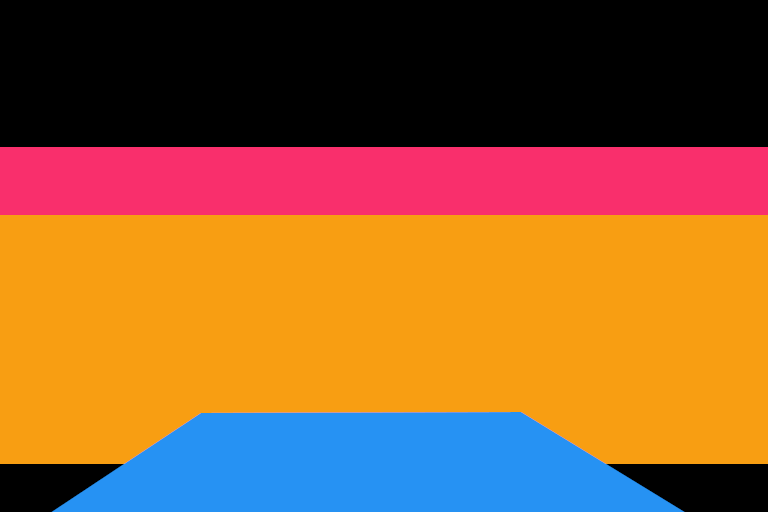}}\h
    \subfloat{\includegraphics[width=\figwidth,height=\figheighta]{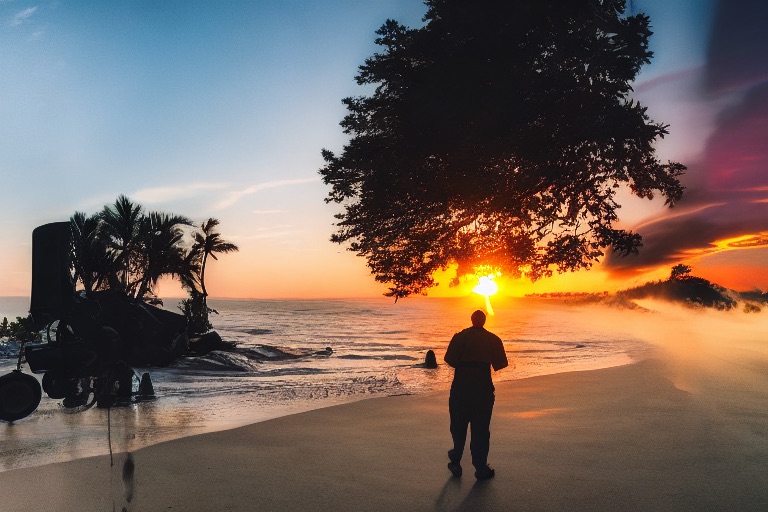}}\h
    \subfloat{\includegraphics[width=\figwidth,height=\figheighta]{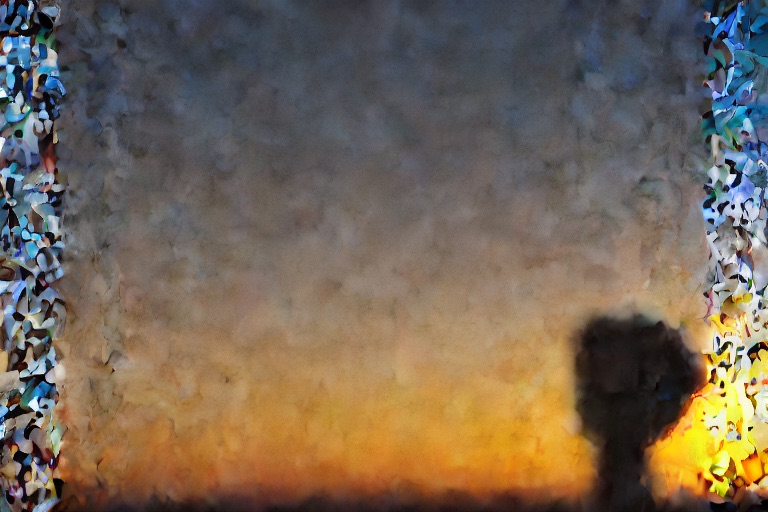}}\h
    \subfloat{\includegraphics[width=\figwidth,height=\figheighta]{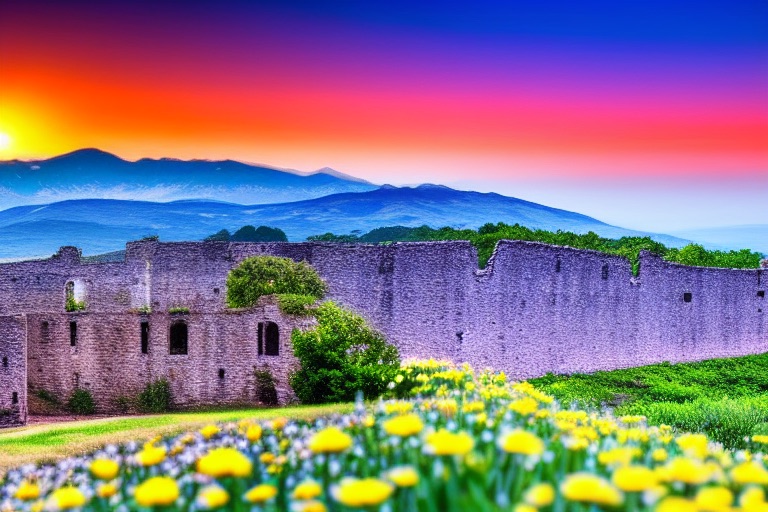}}
    \\[-0.14em]
{\myfont
  \makebox[\hh][c]{\hspace{-0.\linewidth}\scriptsize{%
  Background: \textit{``A photo of outside''}, \,
  {\color{p1color} Yellow}: \textit{``A river''}, \,
  {\color{p2color} Red}: \textit{``A photo of a boy''}, \,
  {\color{p3color} Blue}: \textit{``A purple balloon''}
  }}\vv\\
}
    \subfloat{\includegraphics[width=\figwidth,height=\figheightb]{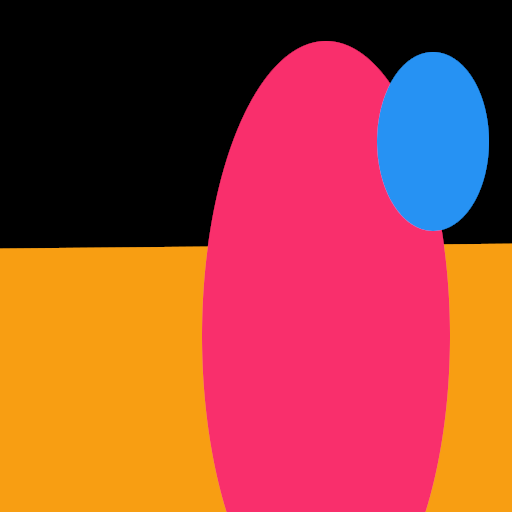}}\h
    \subfloat{\includegraphics[width=\figwidth,height=\figheightb]{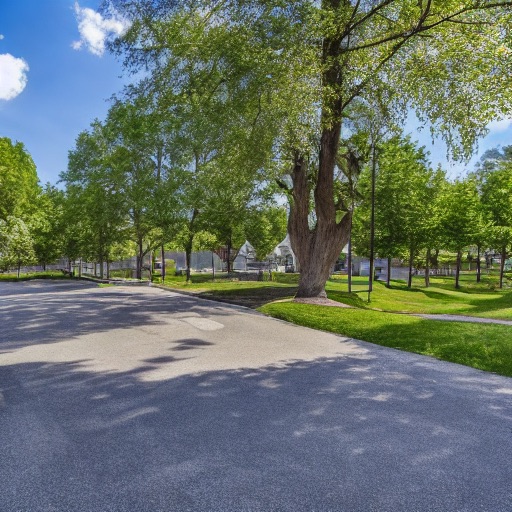}}\h
    \subfloat{\includegraphics[width=\figwidth,height=\figheightb]{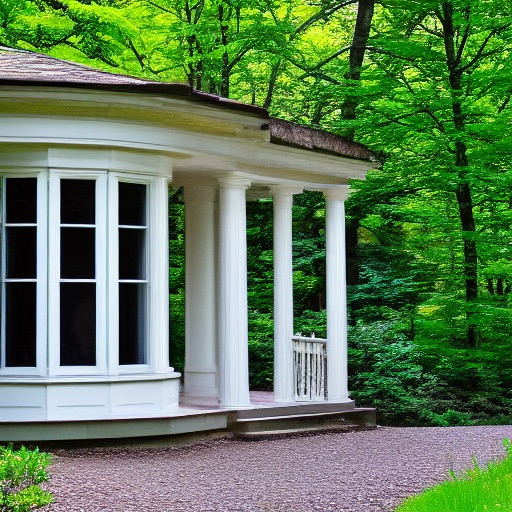}}\h
    \subfloat{\includegraphics[width=\figwidth,height=\figheightb]{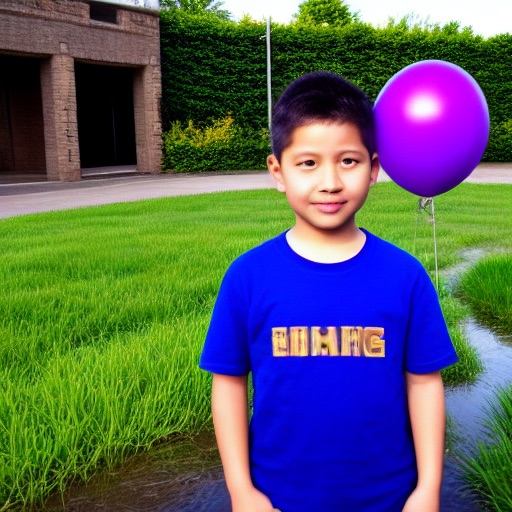}}
    \\[-0.14em]
{\myfont
  \makebox[\hh][c]{\hspace{-0.\linewidth}\scriptsize{%
  Background: \textit{``A grassland''}, \,
  {\color{p1color} Yellow}: \textit{``A tree blossom''}, \,
  {\color{p2color} Red}: \textit{``A photo of small polar bear''}
  }}\vv\\
}
    \subfloat{\includegraphics[width=\figwidth,height=\figheightb]{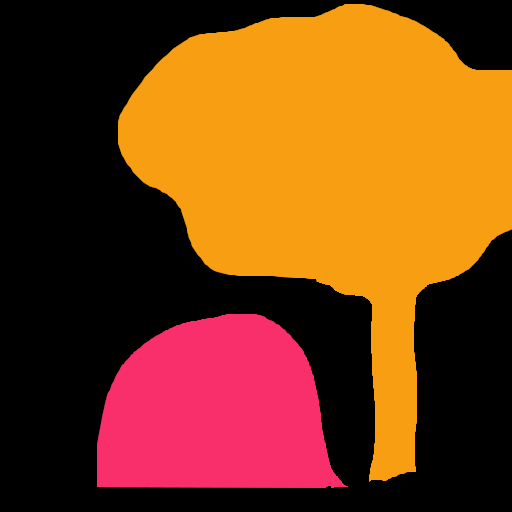}}\h
    \subfloat{\includegraphics[width=\figwidth,height=\figheightb]{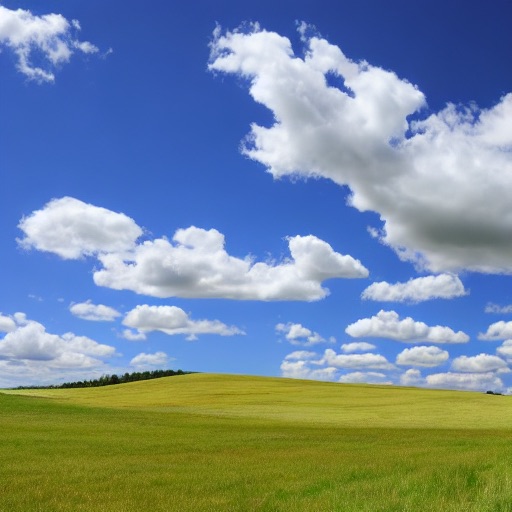}}\h
    \subfloat{\includegraphics[width=\figwidth,height=\figheightb]{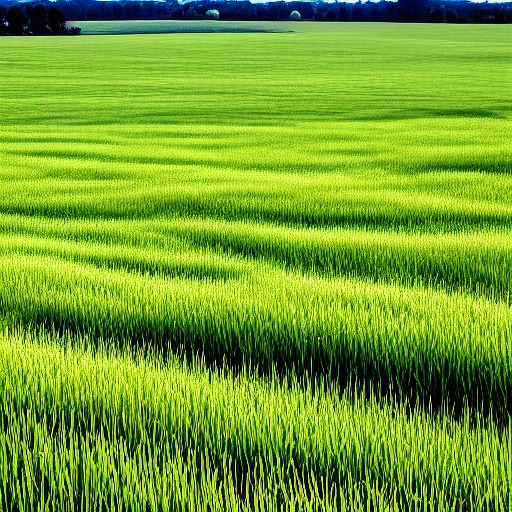}}\h
    \subfloat{\includegraphics[width=\figwidth,height=\figheightb]{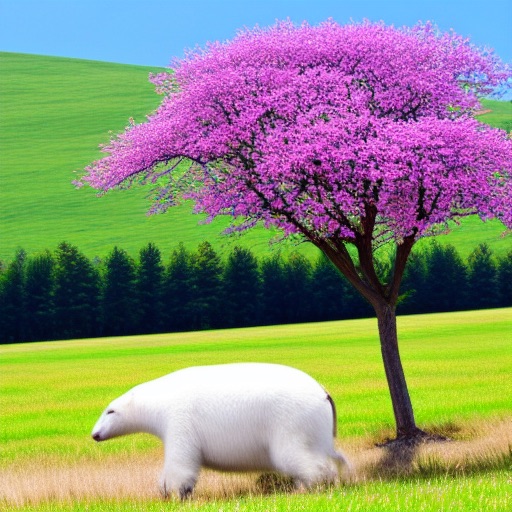}}
    \\[-0.14em]
{\myfont
  \makebox[\hh][c]{\hspace{.05\linewidth}\scriptsize{%
  Background: \textit{``A photo of mountains with lion on the cliff''}, \,
  {\color{p1color} Yellow}: \textit{``A rocky cliff''}, \,
  {\color{p2color} Red}: \textit{``A dense forest''}, \,
  {\color{p3color} Blue}: \textit{``A walking lion''}
  }}\vv\\
}
    \subfloat{\includegraphics[width=\figwidth,height=\figheight]{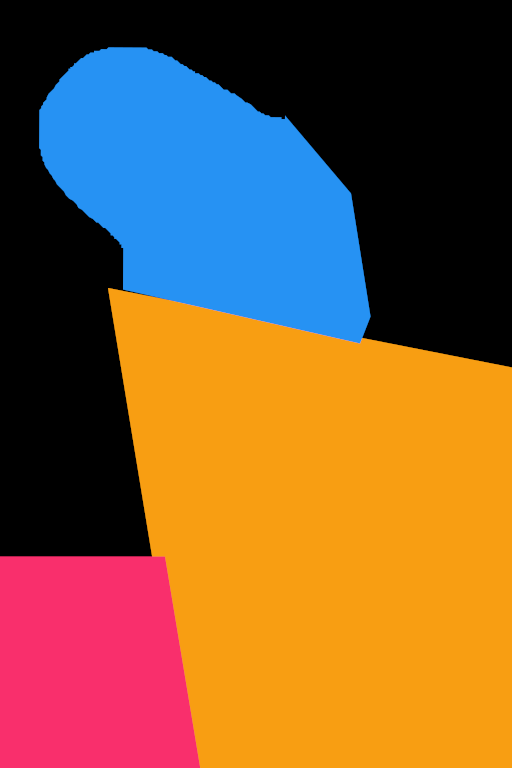}}\h
    \subfloat{\includegraphics[width=\figwidth,height=\figheight]{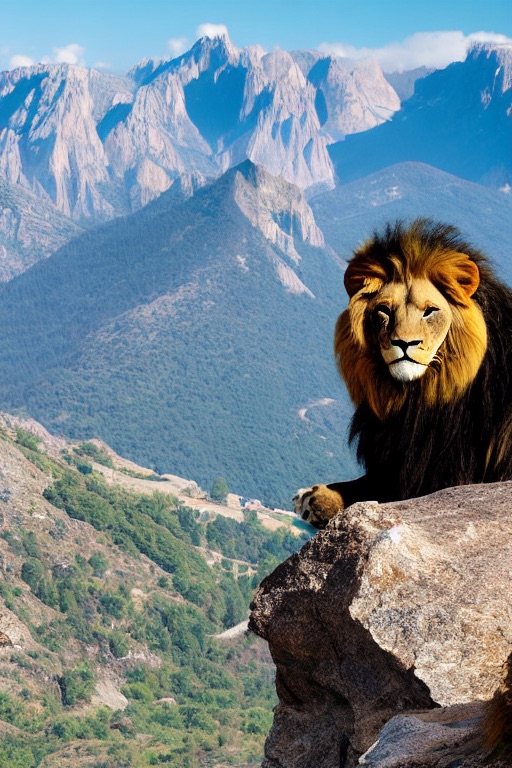}}\h
    \subfloat{\includegraphics[width=\figwidth,height=\figheight]{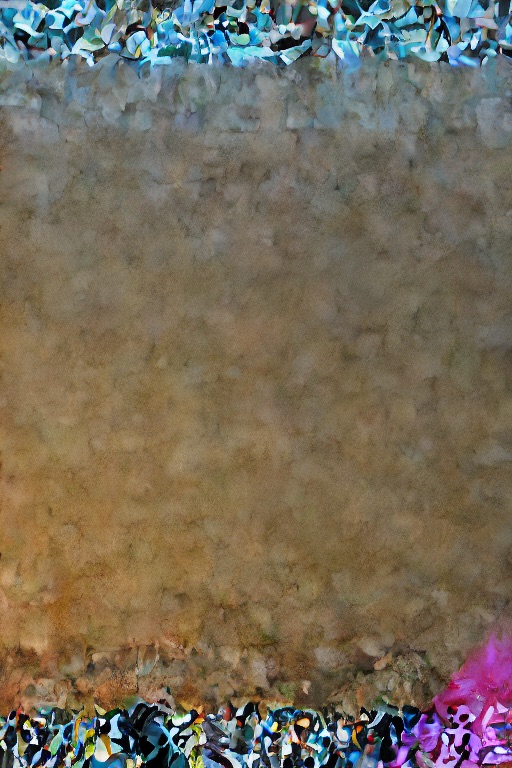}}\h
    \subfloat{\includegraphics[width=\figwidth,height=\figheight]{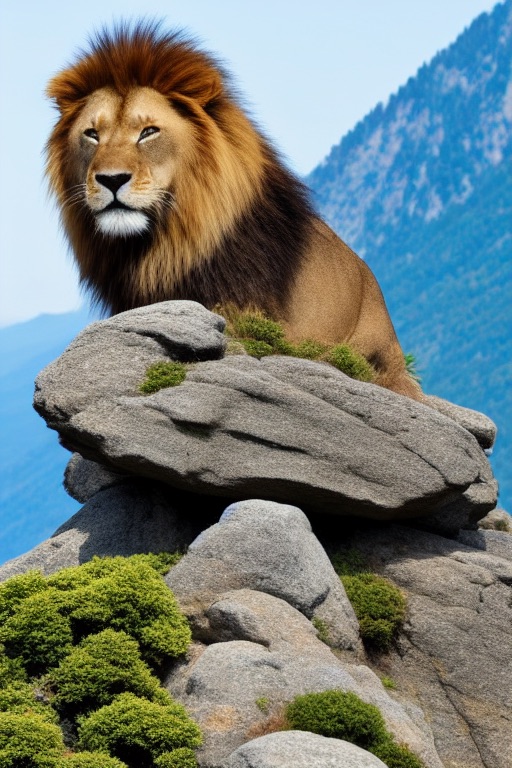}}
    \\[-0.14em]
{\myfont
  \makebox[\hh][c]{\hspace{-0.\linewidth}\scriptsize{%
  Background: \textit{``A photo of the starry sky''}, \,
  {\color{p1color} Yellow}: \textit{``The Earth seen from ISS''}, \,
  {\color{p2color} Red}: \textit{``A photo of a falling asteroid''}
  }}\vv\\
}
    \addtocounter{subfigure}{-16}
    \subfloat[Prompt]{\includegraphics[width=\figwidth,height=\figheight]{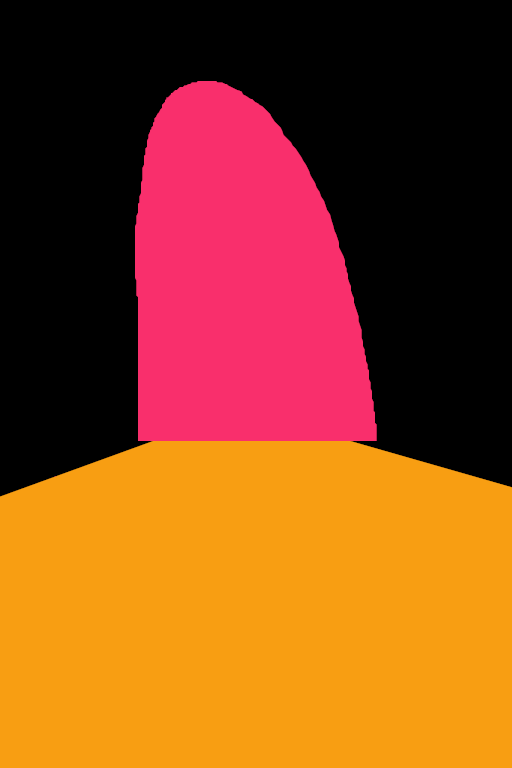}}\h
    \subfloat[MD, 50 steps]{\includegraphics[width=\figwidth,height=\figheight]{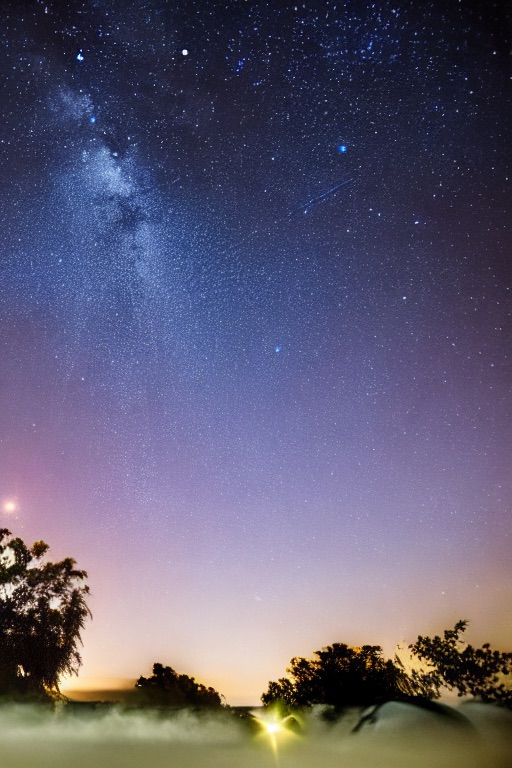}}\h
    \subfloat[MD+LCM, 5 steps]{\includegraphics[width=\figwidth,height=\figheight]{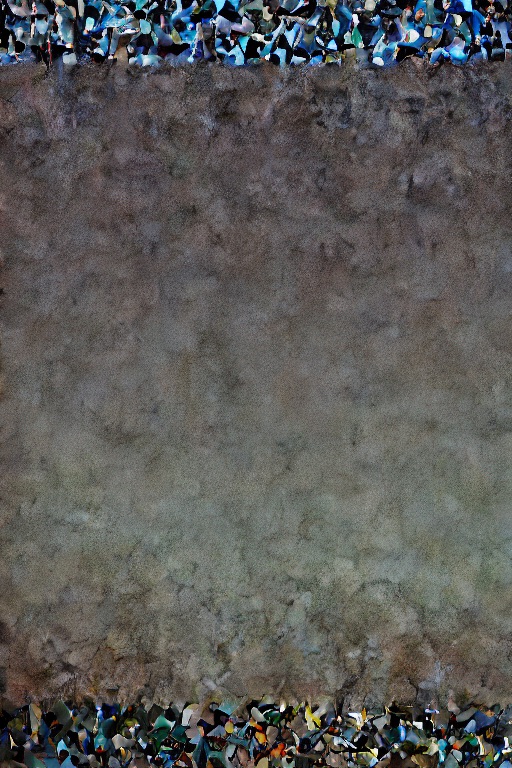}}\h
    \subfloat[{\textbf{Ours, 5 steps}}]{\includegraphics[width=\figwidth,height=\figheight]{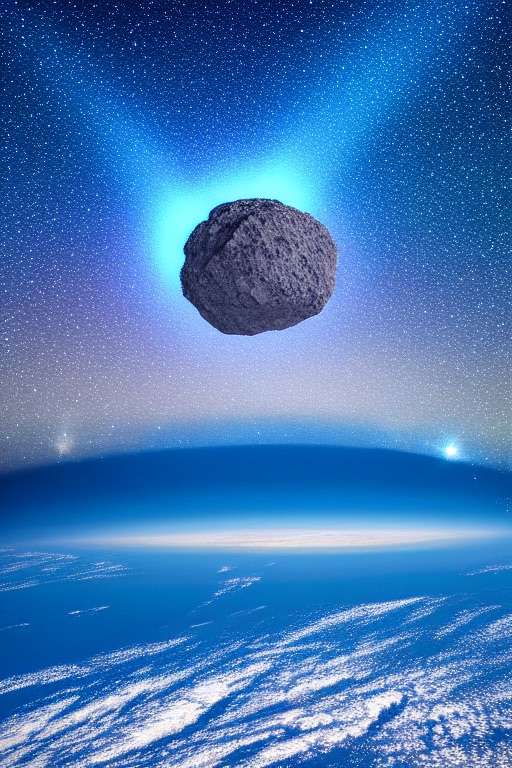}}
\\[-.7em]
  \caption{%
  Additional region-based text-to-image synthesis results.
  Our method accelerates MultiDiffusion~\cite{bar2023multidiffusion} up to $\times$10 while preserving or even boosting mask fidelity.
  }
  \label{fig:appx:region}
\end{figure*}
\begin{figure*}[!htbp]
\newcommand{\figurewidth}{.95\linewidth}
\newcommand{\h}{16.5mm}
\newcommand{\hh}{2.5mm}
\newcommand{\vv}{\vspace*{-0.70mm}}
\definecolor{p1color}{HTML}{16C232}
\definecolor{p2color}{HTML}{F92F6C}
  \centering
  {\myfont
  \makebox[\h][c]{\hspace{-0.\linewidth}\scriptsize{\textit{``A photo of Alps''}}}\vv\\
    \makebox[\hh]{\rotatebox[origin=l]{90}{\makebox[\h][c]{\hspace{-0.\linewidth}\footnotesize{MD ({\color{p2color}\textbf{154s}})}}}}\hspace{0.5mm}%
    \includegraphics[height=\h,width=\figurewidth]{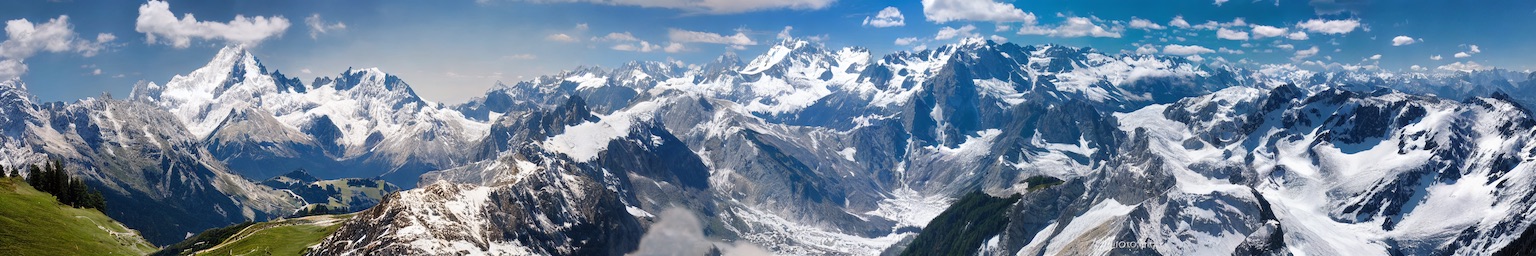}\vv\\
    \makebox[\hh]{\rotatebox[origin=l]{90}{\makebox[\h][c]{\hspace{-0.\linewidth}\footnotesize{MD+LCM ({\color{p1color}10s})}}}}\hspace{0.5mm}%
    \includegraphics[height=\h,width=\figurewidth]{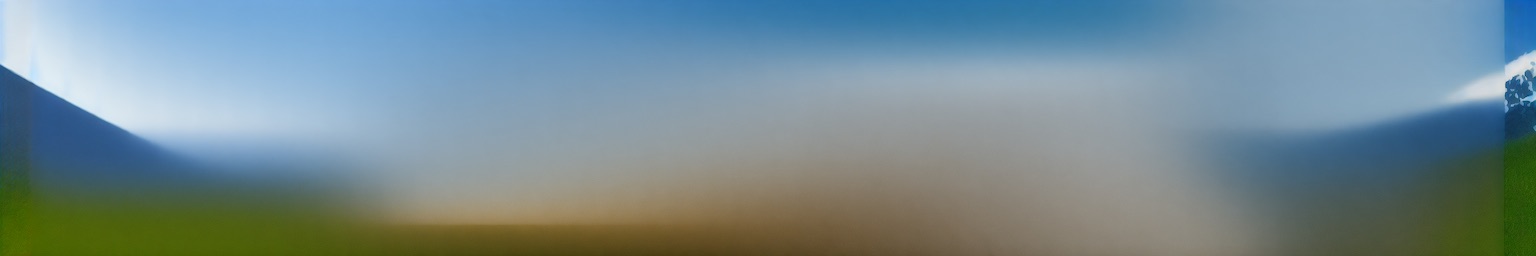}\vv\\
    \makebox[\hh]{\rotatebox[origin=l]{90}{\makebox[\h][c]{\hspace{-0.\linewidth}\footnotesize{\textbf{Ours ({\color{p1color}12s})}}}}}\hspace{0.5mm}%
    \includegraphics[height=\h,width=\figurewidth]{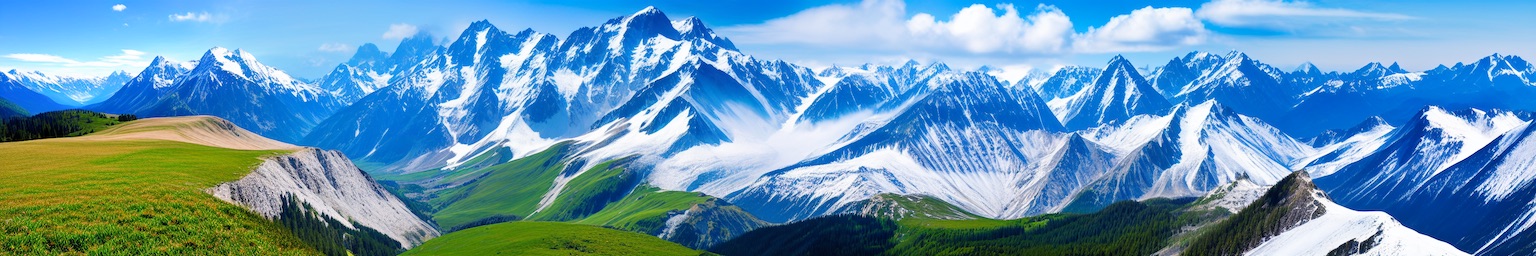}%
    }
  \makebox[\h][c]{\hspace{-0.\linewidth}\scriptsize{\textit{``The battle of Cannae drawn by Hieronymus Bosch''}}}\vv\\
    \makebox[\hh]{\rotatebox[origin=l]{90}{\makebox[\h][c]{\hspace{-0.\linewidth}\footnotesize{MD ({\color{p2color}\textbf{301s}})}}}}\hspace{0.5mm}%
    \includegraphics[height=\h,width=\figurewidth]{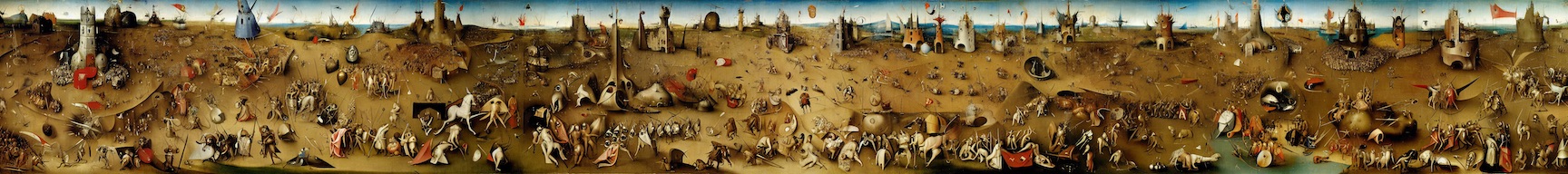}\vv\\
    \makebox[\hh]{\rotatebox[origin=l]{90}{\makebox[\h][c]{\hspace{-0.\linewidth}\footnotesize{MD+LCM ({\color{p1color}17s})}}}}\hspace{0.5mm}%
    \includegraphics[height=\h,width=\figurewidth]{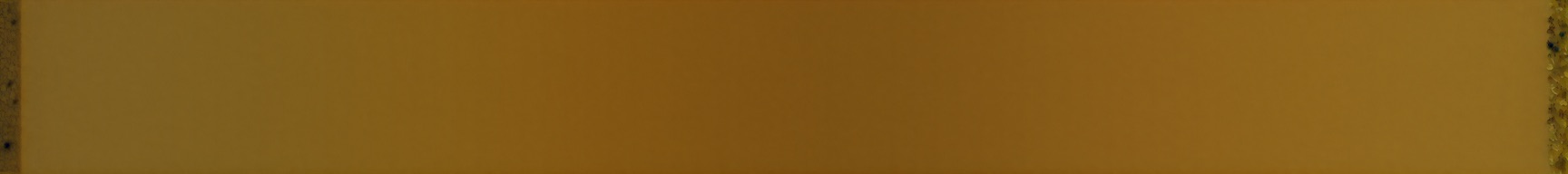}\vv\\
    \makebox[\hh]{\rotatebox[origin=l]{90}{\makebox[\h][c]{\hspace{-0.\linewidth}\footnotesize{\textbf{Ours ({\color{p1color}21s})}}}}}\hspace{0.5mm}%
    \includegraphics[height=\h,width=\figurewidth]{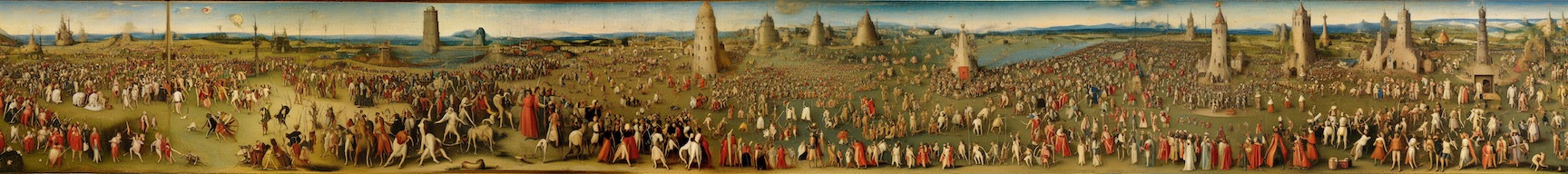}\vv\\%
  {\myfont
  \makebox[\h][c]{\hspace{-0.\linewidth}\scriptsize{\textit{``A photo of a medieval castle in the distance over rocky mountains in winter''}}}\vv\\
    \makebox[\hh]{\rotatebox[origin=l]{90}{\makebox[\h][c]{\hspace{-0.\linewidth}\footnotesize{MD ({\color{p2color}\textbf{296s}})}}}}\hspace{0.5mm}%
    \includegraphics[height=\h,width=\figurewidth]{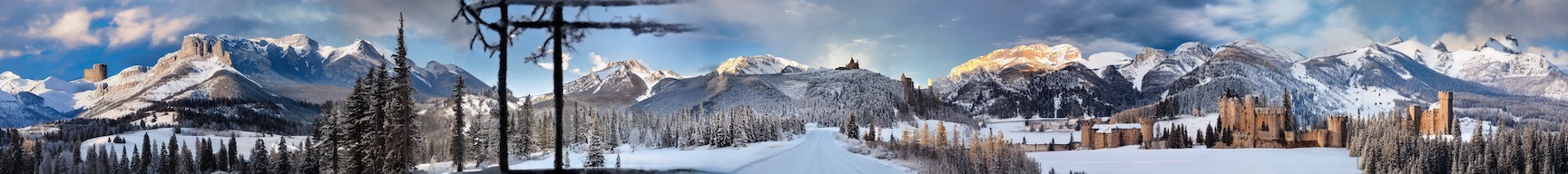}\vv\\
    \makebox[\hh]{\rotatebox[origin=l]{90}{\makebox[\h][c]{\hspace{-0.\linewidth}\footnotesize{MD+LCM ({\color{p1color}19s})}}}}\hspace{0.5mm}%
    \includegraphics[height=\h,width=\figurewidth]{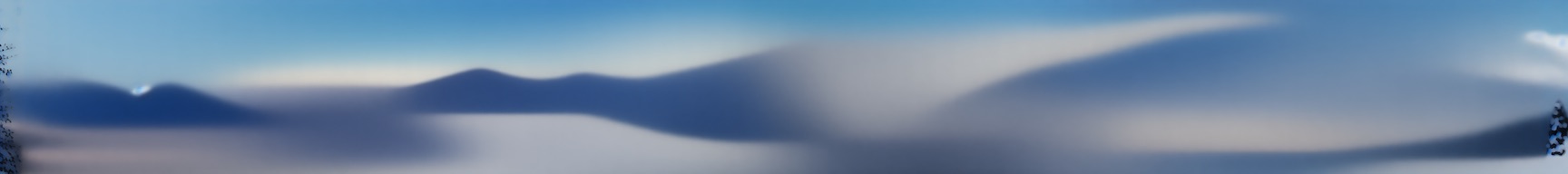}\vv\\
    \makebox[\hh]{\rotatebox[origin=l]{90}{\makebox[\h][c]{\hspace{-0.\linewidth}\footnotesize{\textbf{Ours ({\color{p1color}23s})}}}}}\hspace{0.5mm}%
    \includegraphics[height=\h,width=\figurewidth]{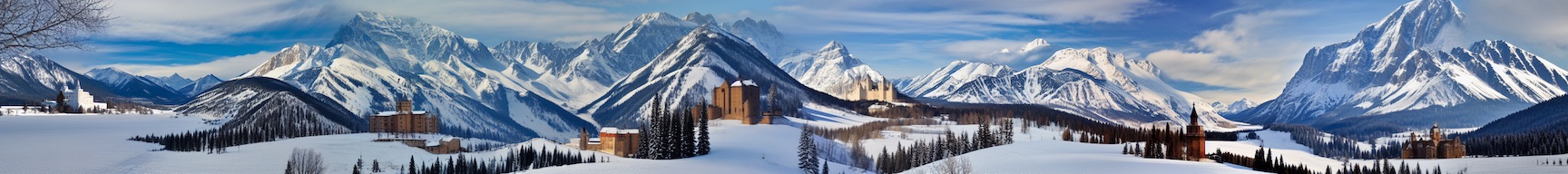}\vv\\%
  \makebox[\h][c]{\hspace{-0.\linewidth}\scriptsize{\textit{``A photo under the deep sea with many sea animals''}}}\vv\\
    \makebox[\hh]{\rotatebox[origin=l]{90}{\makebox[\h][c]{\hspace{-0.\linewidth}\footnotesize{MD ({\color{p2color}\textbf{290s}})}}}}\hspace{0.5mm}%
    \includegraphics[height=\h,width=\figurewidth]{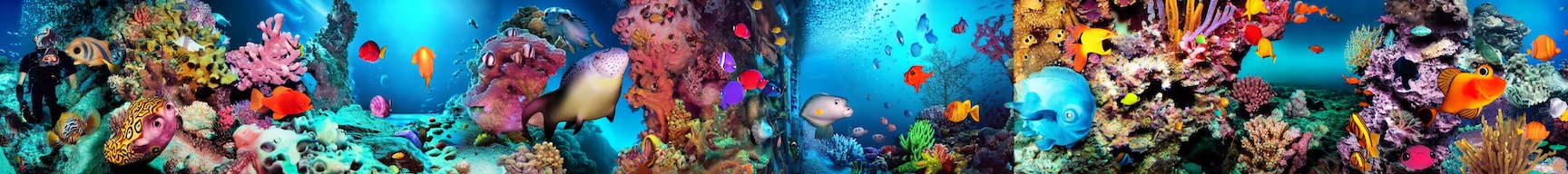}\vv\\
    \makebox[\hh]{\rotatebox[origin=l]{90}{\makebox[\h][c]{\hspace{-0.\linewidth}\footnotesize{MD+LCM ({\color{p1color}18s})}}}}\hspace{0.5mm}%
    \includegraphics[height=\h,width=\figurewidth]{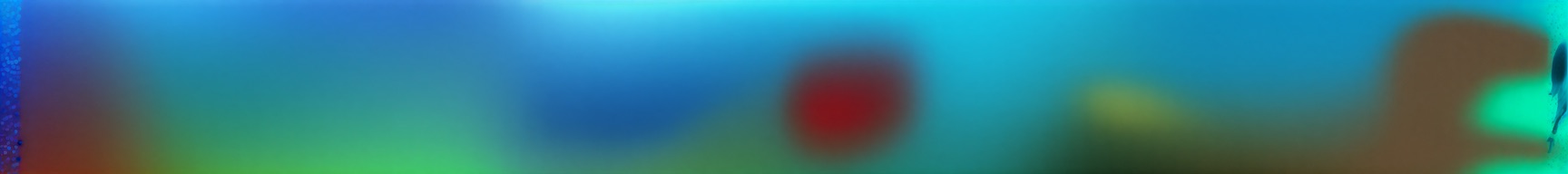}\vv\\
    \makebox[\hh]{\rotatebox[origin=l]{90}{\makebox[\h][c]{\hspace{-0.\linewidth}\footnotesize{\textbf{Ours ({\color{p1color}23s})}}}}}\hspace{0.5mm}%
    \includegraphics[height=\h,width=\figurewidth]{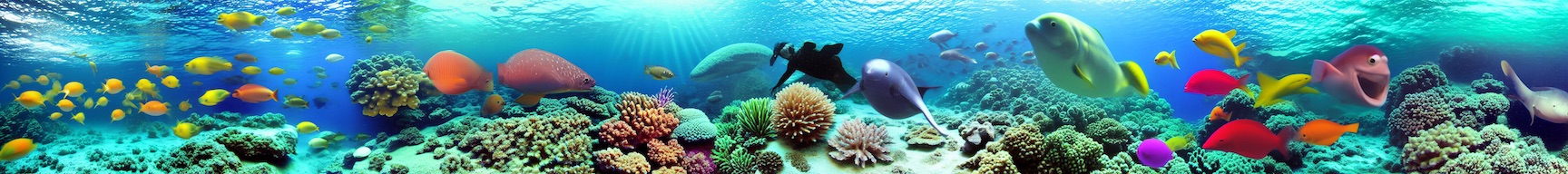}%
    }
  \caption{%
  Additional panorama generation results.
  The images of size $512 \times 4608\,$ are sampled with 50 steps for MD and 4 steps for MD+LCM and Ours.
  Our \textsc{SemanticDraw} can synthesize high-resolution images in seconds.
  We achieve $\times 13$ improvement in inference latency.
  }
  \label{fig:appx:panorama1}
\end{figure*}

\subsection{Region-Based Text-to-Image Generation}
\label{sec:b_exp_qual:region}

We show additional region-based text-to-image generation results in Figure~\ref{fig:appx:region}.
In addition to Figure~6 of the main manuscript, the generated samples show that our method is able to accelerate region-based text-to-image generation consistently by $\times 10$ without compromising the generation quality.
Moreover, Figure~2 of the main manuscript has shown that the benefits from our acceleration method for arbitrary-sized generation and region-based controls are indeed simultaneously enjoyable.
Our acceleration method enables publicly available Stable Diffusion v1.5~\cite{rombach2022high} to generate a $1920 \times 768$ scene from eight hand-drawn masks in 59 seconds, which is $\times 52.5$ faster than the baseline~\cite{bar2023multidiffusion} taking more than 51 minutes to converge into a low-fidelity image.
Figure~\ref{fig:appx:figure_one_overlay} shows mask fidelity of this generation.
We can visualize that even if the generated image has larger dimension than the dimensions the model has been trained for, \textit{i.e.}, $768 \times 768\,$, the mask fidelity is achieved under this accelerated generation.
Locations and sizes of the Sun and the Moon match to the provided masks in near perfection; whereas mountains and waterfalls are harmonized within the overall image, without violating region boundaries.
This shows that the flexibility and the speed of our generation paradigm, \textsc{SemanticDraw}, is also capable of professional usage.

Furthermore, we have found that more recent methods such as LRDiff~\cite{qi2024lrdiff} also suffers the same instability problem when accelerated.
Figure~\ref{fig:lrdiff} shows one example.
In this qualitative results, our method not only achieves faster generation speed ($\times 45$), but also enjoys better mask fidelity and perceptual quality.
This further validates the significance of our strategy in professional interactive content creation.

Regarding that professional image creation process using diffusion models typically involves a multitude of resampling trials with different seeds, the original baseline model's convergence speed of one image per hour severely limits the applicability of the algorithm.
In contrast, our acceleration method enables the same large-size region-based text-to-image synthesis to be done under a minute, making this technology practical to industrial usage.

\begin{figure*}[!htbp]
  \centering
  \begin{subcaptionbox}{
    Screenshot of the application.
    \label{fig:application:screenshot}
  }[0.38\linewidth]
  {
    \includegraphics[width=\linewidth]{figures/app_schematic/app_screenshot1.png}
  }
  \end{subcaptionbox}
  \hfill
  \begin{subcaptionbox}{
    Application design schematics.
    \label{fig:application:schematic}
  }[0.58\linewidth]
  {
    \includegraphics[width=\linewidth]{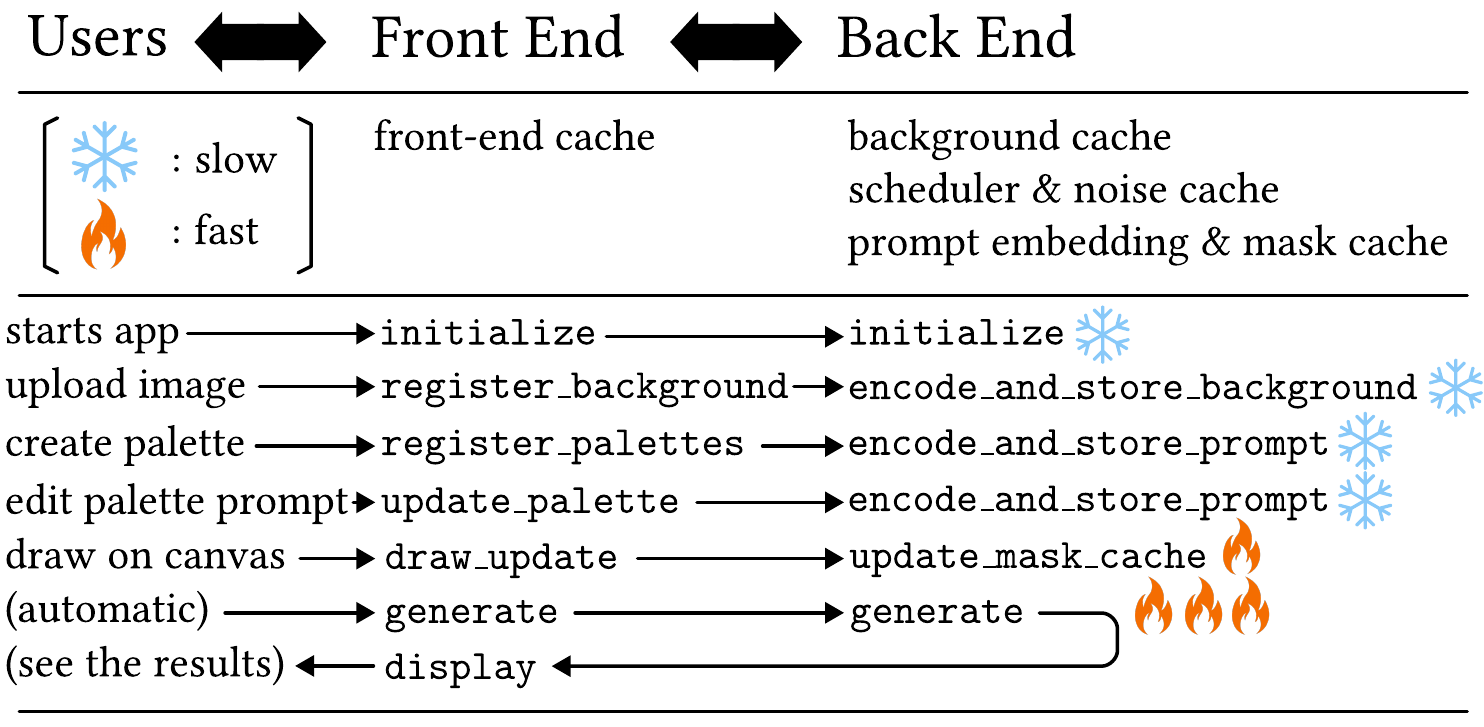}
  }
  \end{subcaptionbox}
  \caption{%
  Sample application demonstrating \emph{semantic palette} enabled by our \textsc{SemanticDraw} algorithm.
  After registering prompts and optional background image, the users can create images in real-time by drawing with text prompts.
  }
  \label{fig:application}
\end{figure*}

\subsection{Panorama Generation}
\label{sec:b_exp_qual:panorama}

We can also visually compare arbitrary-sized image creation with panorama image generation task.
As briefly mentioned in Section~\ref{sec:a_algorithm}, comparing with this task reveals the problem of incompatibility between accelerating schedulers and current region-based multiple text-to-image synthesis pipelines.
Figure~\ref{fig:appx:panorama1} shows the results of large-scale panorama image generation using our method, where we generate $512 \times 4608$ images from a single text prompt.
Na\"ively applying acceleration to existing solution leads to blurry unrealistic generation, enforcing users to resort to more conventional diffusion schedulers that take long time to generate~\cite{song2020denoising}.
Instead, our method is compatible to accelerated samplers~\cite{luo2023latent,luo2023lcm,lin2024sdxl,ren2024hyper,chadebec2024flash}, showing $\times 13$ faster generation of images with sizes much larger than the resolutions of $512 \times 512$ or $768 \times 768\,$, for which the diffusion model~\cite{rombach2022high} is trained.
Combining results from Section~\ref{sec:b_exp_qual:region} and~\ref{sec:b_exp_qual:panorama} our Algorithm~\ref{alg:single} significantly broadens the usability of diffusion models for professional content creators.
This leads to the last section of this Supplementary Material, the description of our submitted demo application.

\section{Sample Application}
\label{sec:e_app_instruction}
This last section elaborates on the design and the example usage of our demo application of \textsc{SemanticDraw}, introduced in Section~5 of the main manuscript.
Starting from the basic description of user interface in Section~\ref{sec:e_app:ui}, we discuss the expected usage of the app in Section~\ref{sec:e_app:basic-usage}.
Our discussion mainly focuses on how real-time interactive content creation is achieved from accelerated region-based text-to-image generation algorithm we have provided.

\subsection{User Interface}
\label{sec:e_app:ui}

As illustrated in Figure~\ref{fig:application:schematic}, user interactions are classified into two groups, \textit{i.e.}, the {\color{importantcolor}{slow}} processes and the {\color{importantcolor}{fast}} processes, based on the latency of response from the model.
Due to the high overhead of text encoder and image encoder, the processes involving these modules are classified as {slow} processes.
However, operations such as preprocessing or saving of mask tensors and sampling of the U-Net for a single step take less than a second.
These processes are, therefore, classified as {fast} ones.
\textsc{SemanticDraw}, our suggested paradigm of image generation, comes from the observation that, if a user first registers text prompts, image generation from user-drawn regions can be done \textit{in real-time}.

\begin{figure*}[tb]
  \centering
  \includegraphics[width=\linewidth]{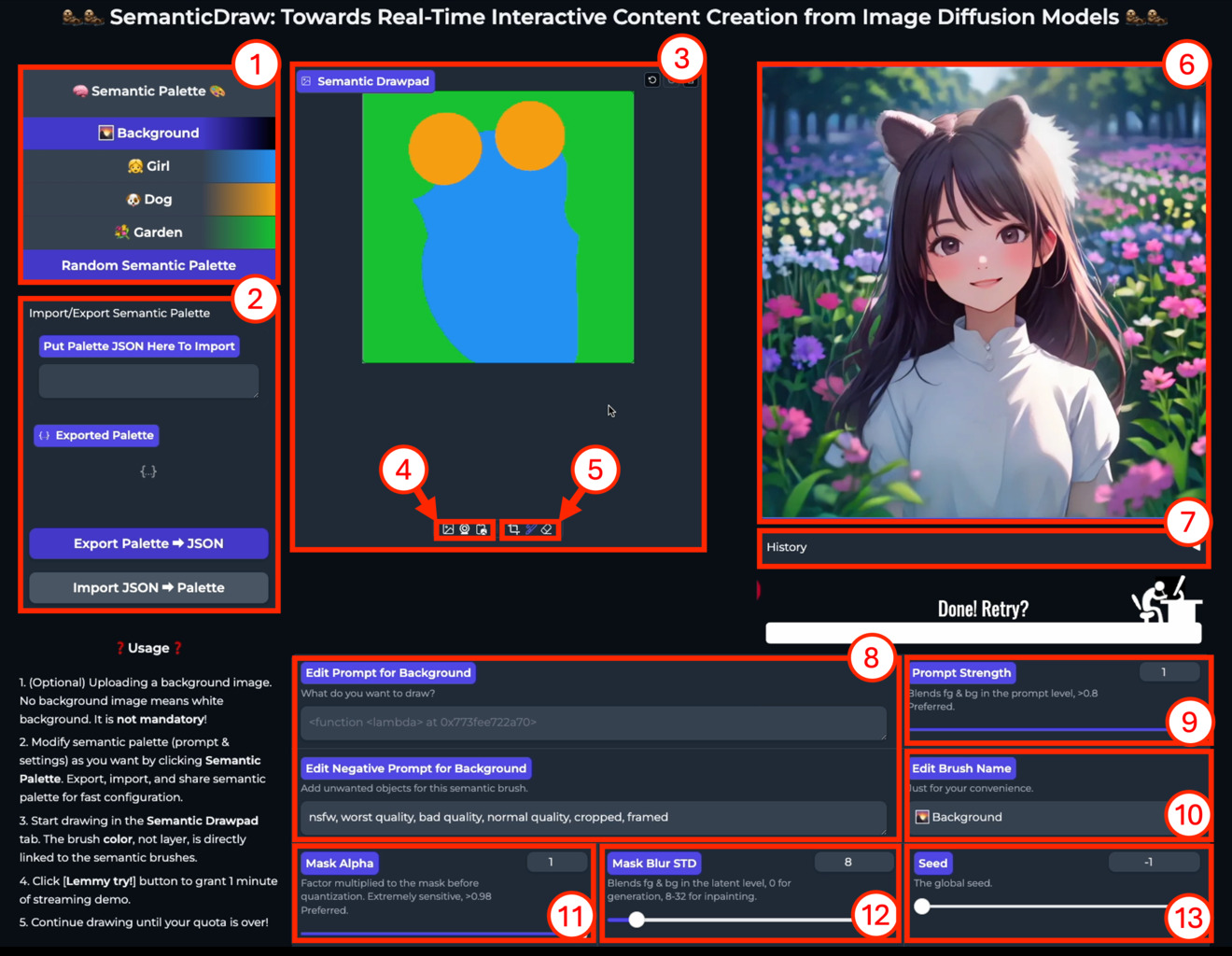}
  \caption{%
  Screenshot of our supplementary demo application.
  Details of the numbered components are elaborated in Table~\ref{tab:appx:controls}.
  }
  \label{fig:appx:demo}
\end{figure*}

The user interface of Figure~\ref{fig:appx:demo} is designed based on the philosophy to maximize user interactions of the {fast} type and to hide the latency of the {slow} type.
Figure~\ref{fig:appx:demo} and Table~\ref{tab:appx:controls} summarize the components of our user interface.
The interface is divided into four compartments:
the (a) {\color{importantcolor} \textit{semantic palette}}, which is a palette of registered text prompts (no. 1-2),
the (b) {\color{importantcolor} drawing screen} (no. 3-5),
the (c) {\color{importantcolor} streaming display and controls} (no. 6-7), and
the (d) {\color{importantcolor} control panel} for the additional controls (no 8-13).
The (a) \textit{semantic palette} manages the number of \textit{semantic brushes} to be used in the generation, which will be further explained below.
Users are expected to interact with the application mainly through (b) drawing screen, where users can upload backgrounds and draw on the screen with selected {semantic brush}.
Then, by turning (c) streaming interface on, the users can receive generated images based on the drawn regional text prompts in real-time.
The attributes of {semantic brushes} are modified through (d) control panel.

Types of the transaction data between application and user are in twofold: a (1) background and a (2) list of text prompt-mask pairs, named {\color{importantcolor} \textit{semantic brushes}}.
The user can register these two types of data to control the generation stream.
Each {semantic brush} consists of two part: (1) {\color{importantcolor} text prompt}, which can be edited in the (d) control panel after clicking on the brush in (a) \textit{semantic palette}, a set of available text prompts to draw with, and (2) {\color{importantcolor} mask}, which can be edited by selecting the corresponding color brush at {\color{importantcolor} drawing tools} (no. 5), and drawing on the {\color{importantcolor} drawing pad} (no. 3) with a brush of any color.
Note that in the released version of our code, the color of semantic brush does not affect generation results.
Its color only separates a semantic brush from another for the users to discern.

\begin{table*}[tb]
  \caption{%
Description of each numbered component in the \textsc{SemanticDraw} demo application of Figure~\ref{fig:appx:demo}.
  }
  \label{tab:appx:controls}
  \centering
  \resizebox{\linewidth}{!}{%
  \begin{tabular}{@{}cll@{}}
    \toprule
    No. & Component Name & Description\\
    \midrule
    1 & \textit{Semantic palette} & Create and manage text prompt-mask pairs. \\
    2 & Import/export semantic palette & Easy management of text prompt sets to draw. \\
    3 & Main drawing pad & User draws at each semantic layers with brush tool. \\
    4 & Background image upload & User uploads background image to start drawing. \\
    5 & Drawing tools & Using brush and erasers to interactively edit the prompt masks. \\
    6 & Display & Generated images are streamed through this component. \\
    7 & History & Generated images are logged for later reuse. \\
    8 & Prompt edit & User can interactively change the positive/negative prompts at need. \\
    9 & Prompt strength control & Prompt embedding mix ratio between the current \& the background. Helps content blending. \\
    10 & Brush name edit & Adds convenience by changing the name of the brush. Does not affect the generation. \\
    11 & Mask alpha control & Changes the mask alpha value before quantization. Recommended: $>0.95\,$. \\
    12 & Mask blur std. dev. control & Changes the standard deviation of the quantized mask of the current semantic brush. \\
    13 & Seed control & Changes the seed of the application. \\
  \bottomrule
  \end{tabular}
  }
\end{table*}
\begin{figure*}[tb]
  \centering
  \begin{subcaptionbox}{
    Upload a background image.
    \label{fig:appx:demo_howto:1}
  }[0.49\linewidth]
  {
    \includegraphics[width=\linewidth]{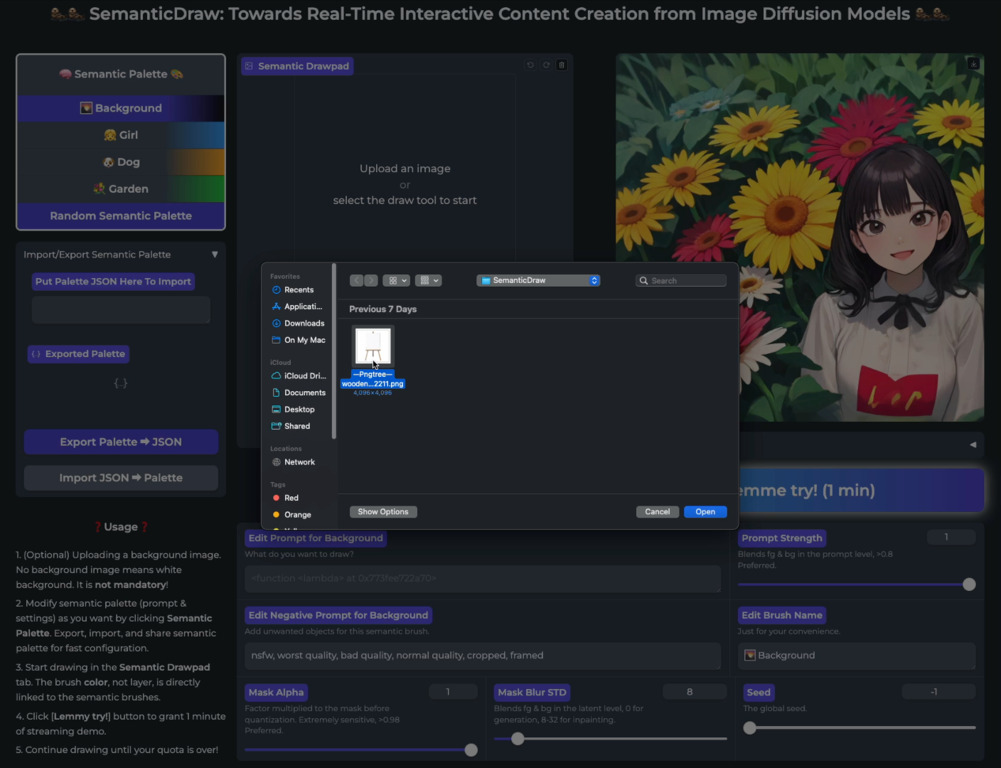}
  }
  \end{subcaptionbox}
  \hfill
  \begin{subcaptionbox}{
    Register \textit{semantic palette}.
    \label{fig:appx:demo_howto:2}
  }[0.49\linewidth]
  {
    \includegraphics[width=\linewidth]{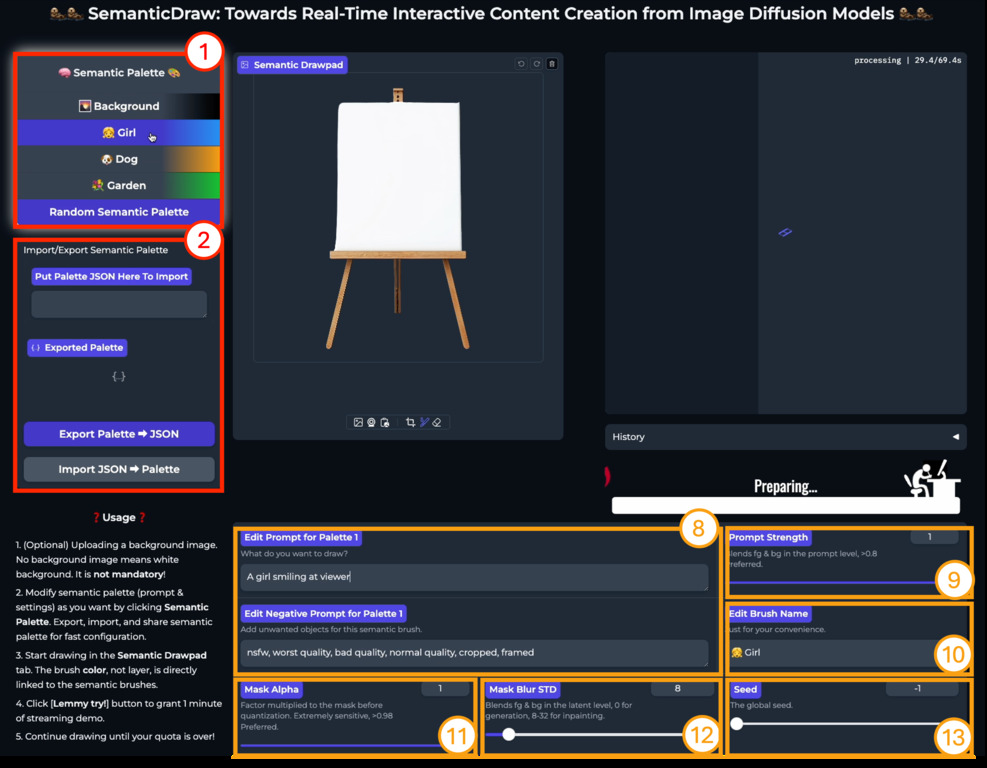}
  }
  \end{subcaptionbox} \\
  \begin{subcaptionbox}{
    Draw with semantic brushes.
    \label{fig:appx:demo_howto:3}
  }[0.49\linewidth]
  {
    \includegraphics[width=\linewidth]{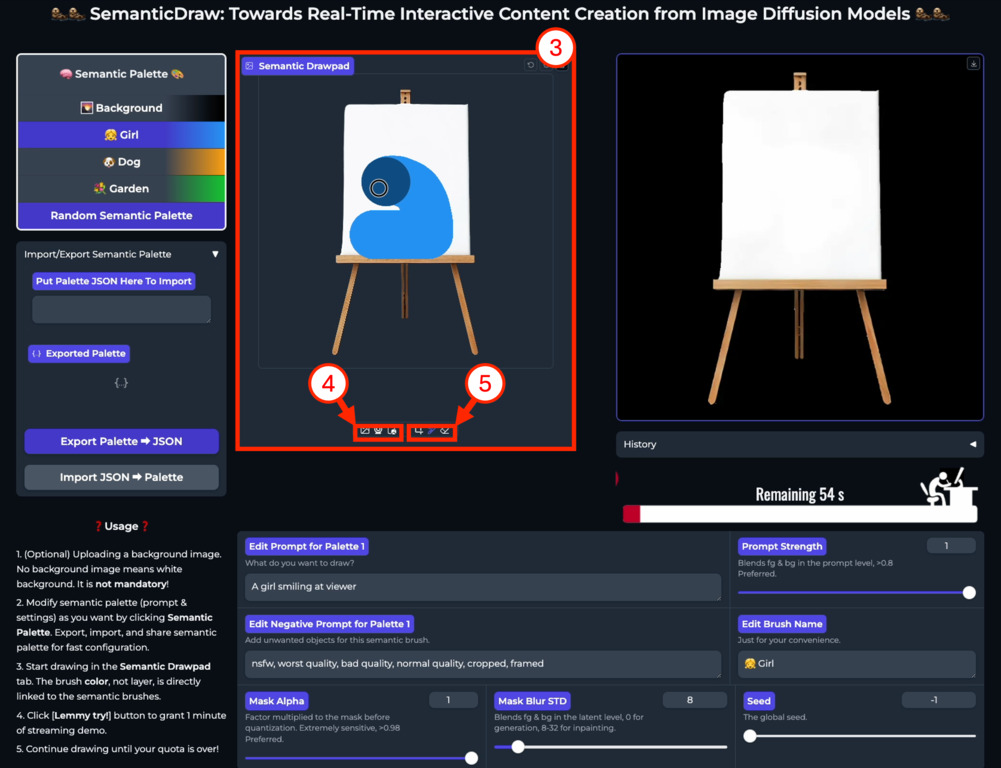}
  }
  \end{subcaptionbox}
  \hfill
  \begin{subcaptionbox}{
    Play the stream and interact.
    \label{fig:appx:demo_howto:4}
  }[0.49\linewidth]
  {
    \includegraphics[width=\linewidth]{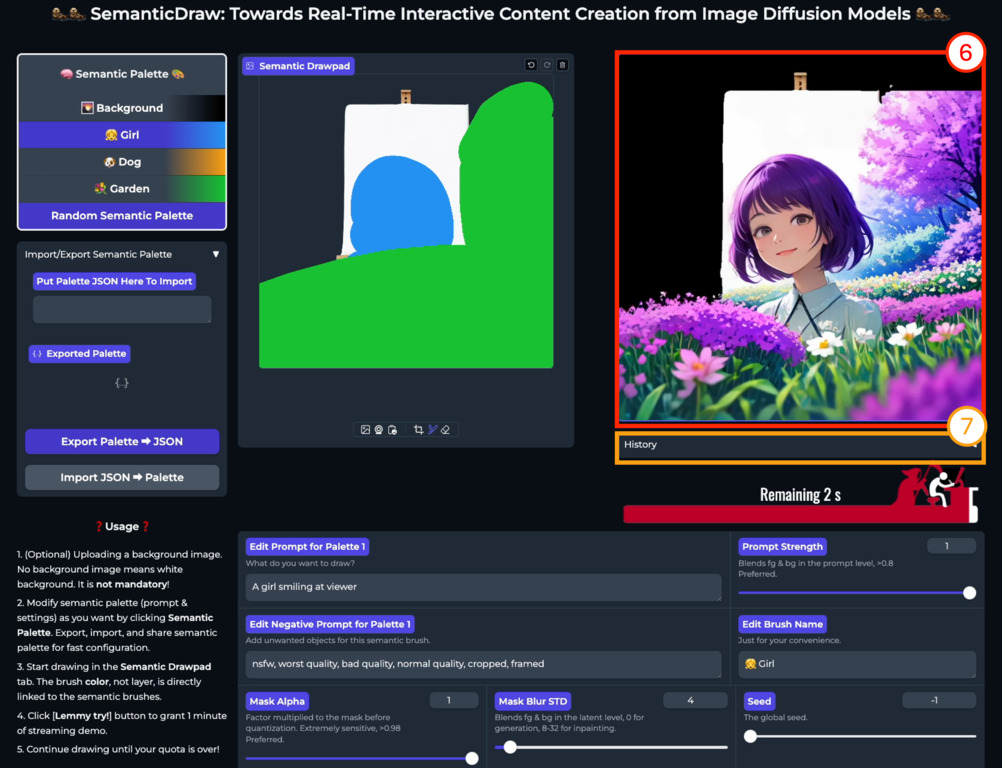}
  }
  \end{subcaptionbox}
  \caption{%
  Illustrated usage guide of our demo application of \textsc{SemanticDraw}.
  }
  \label{fig:appx:demo_howto}
\end{figure*}

As the interface of the (d) control panel implies, our reformulation of MultiDiffusion~\cite{bar2023multidiffusion} provides additional hyperparameters that can be utilized for professional creators to control their creation processes.
The {\color{importantcolor} mask alpha} (no. 11) and the {\color{importantcolor} mask blur std} (no. 12) determine preprocessing attributes of selected {semantic brush}.
Before the mask is quantized into predefined noise levels of scheduler, as elaborated in Section~\ref{sec:a_alg:maskquant}, mask is first multiplied by mask alpha and goes through an isotropic Gaussian blur with a specified standard deviation.
That is, given a mask $\boldsymbol{w}\,$, a mask alpha $a\,$, and the noise level scheduling function $\beta(t) = \sqrt{1 - \alpha(t)}\,$, the resulting quantized mask $\boldsymbol{w}^{(t_{i})}_{1:p}$ is:
\begin{equation}
    \label{eq:appx:maskquant}
    \boldsymbol{w}^{(t_{i})}_{1:p} = \mathbbm{1}[a\boldsymbol{w} > \beta(t_{i})]\,,
\end{equation}
where $\mathbbm{1}[\cdot]$ is an indicator function taking the inequality as a binary operator to make a boolean mask tensor $\boldsymbol{w}^{(t_{i})}_{1:p}$ at time $t_{i}\,$.
The default noise levels $\beta(t)$ of the acceleration modules~\cite{luo2023latent,luo2023lcm,lin2024sdxl,ren2024hyper,chadebec2024flash} are close to one, as shown in Figure~5 of the main manuscript.
This makes mask alpha extremely sensitive.
By changing its value only slightly, \textit{e.g.}, down to 0.98, the corresponding prompt already skips first two sampling steps.
This quickly degenerates the content of the prompt, and therefore, the {\color{importantcolor} mask alpha} (no. 11) should be used in care.
The effect of {\color{importantcolor} mask blur std} (no. 12) is shown in Figure~\ref{fig:maskquant:result}, and will not be further elaborated in this section.
The seed of the system can be tuned by {\color{importantcolor} seed control} (no. 13).
Nonetheless, controlling pseudo-random generator will rarely be needed since the application generates images in an infinite stream.
The {\color{importantcolor} prompt edit} (no. 8) is the main control of {semantic brush}.
The users can change text prompt even when generation is on stream.
It takes exactly the total number of inference steps, \textit{i.e.}, 5 steps, for a change in prompts to take effect.
Further, we provide {\color{importantcolor} prompt strength} (no. 9) as an alternative to highly sensitive {\color{importantcolor} mask alpha} (no. 11) to control the saliency of the target prompt.
Although modifying the alpha channel provides good intuition for graphics designer being already familiar to alpha blending, the noise levels of consistency model~\cite{song2023consistency,luo2023latent,luo2023lcm,lin2024sdxl,ren2024hyper,chadebec2024flash} make the mask alpha value not aligned well with our intuition in alpha blending.
Prompt strength is a mix ratio between the embeddings of the foreground text prompt of given {semantic brush} and background text prompt.
We empirically find that changing the prompt strengths gives smoother control to the foreground-background blending strength than mask alpha.
However, whereas the mask alpha can be applied locally, the prompt strength only globally takes effect.
Therefore, we believe that the two controls are complementary to one another.

%-------------------------------------------------------------------------
\begin{figure*}[t]
  \centering
  \setlength{\tabcolsep}{0pt}  % Remove horizontal spacing between columns
  \begin{tabular}{@{}c@{}c@{}c@{}c@{}c@{}c@{}c@{}}  % Remove padding around columns
    \includegraphics[width=0.14\linewidth,height=0.14\linewidth]{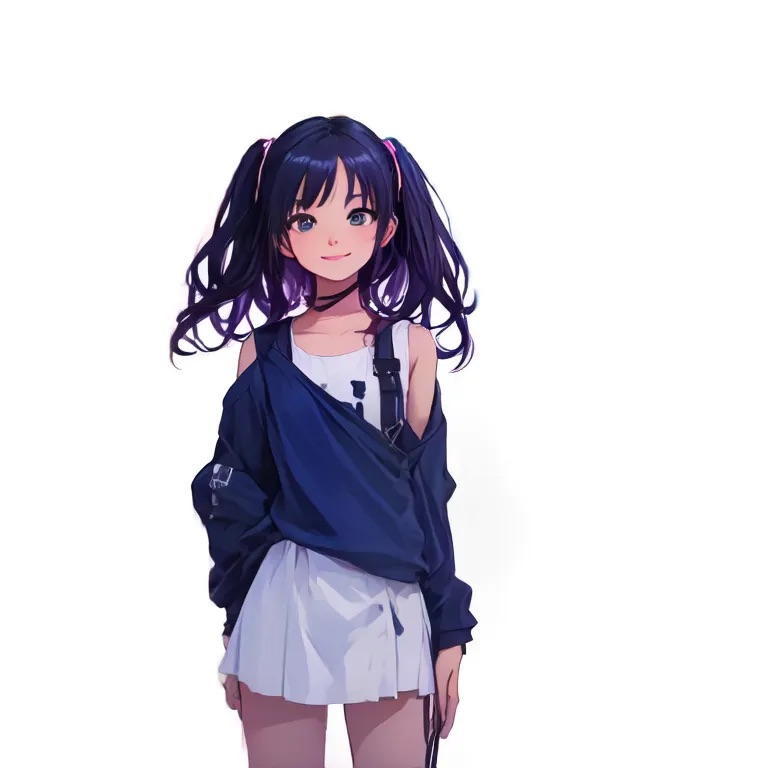} &
    \includegraphics[width=0.14\linewidth,height=0.14\linewidth]{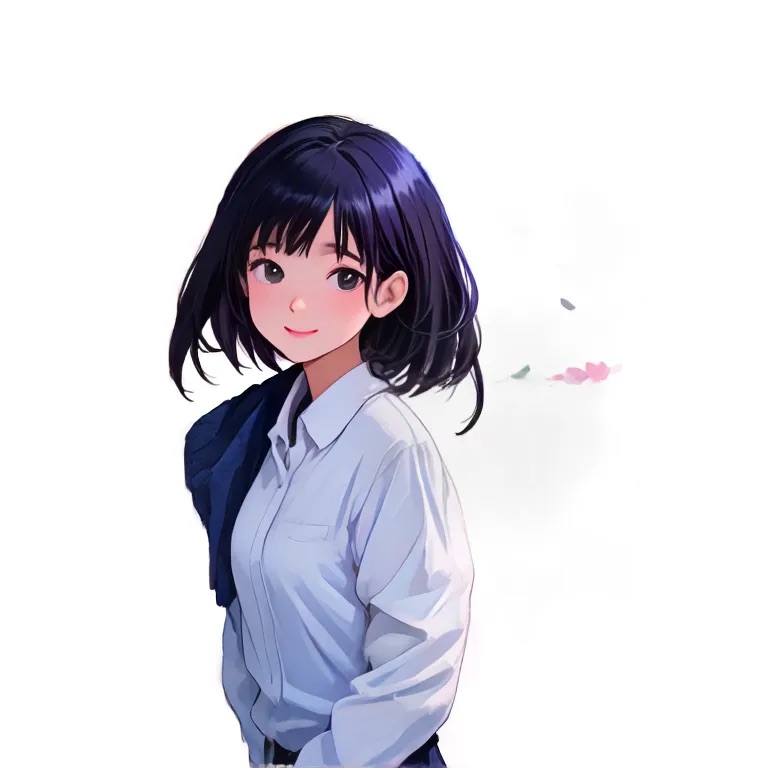} &
    \includegraphics[width=0.14\linewidth,height=0.14\linewidth]{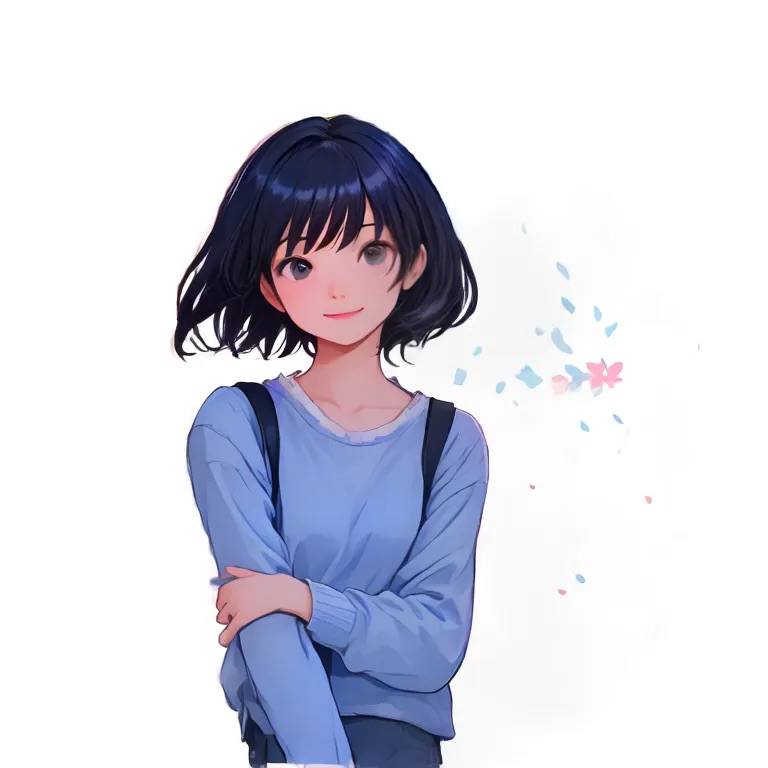} &
    \includegraphics[width=0.14\linewidth,height=0.14\linewidth]{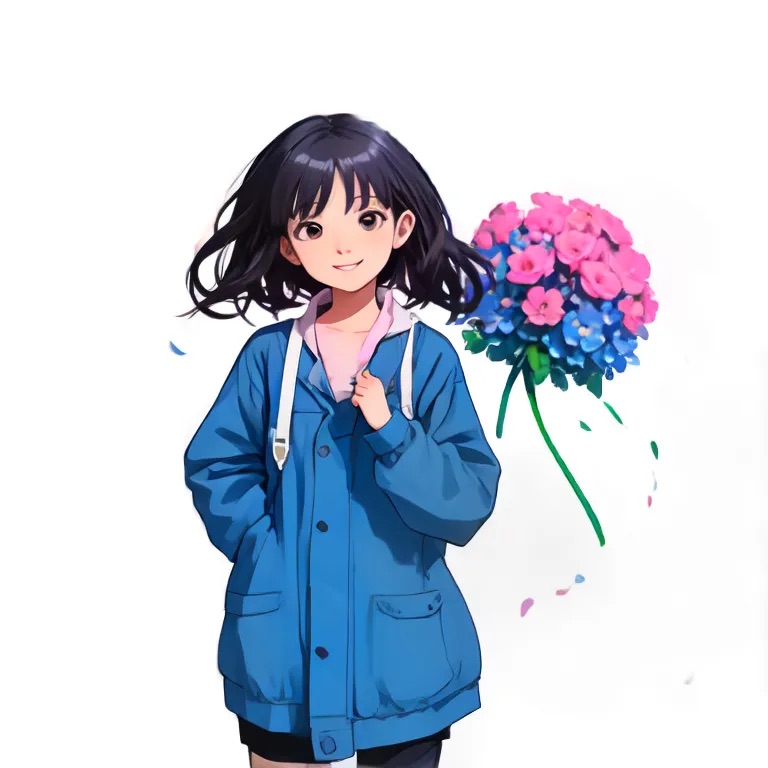} &
    \includegraphics[width=0.14\linewidth,height=0.14\linewidth]{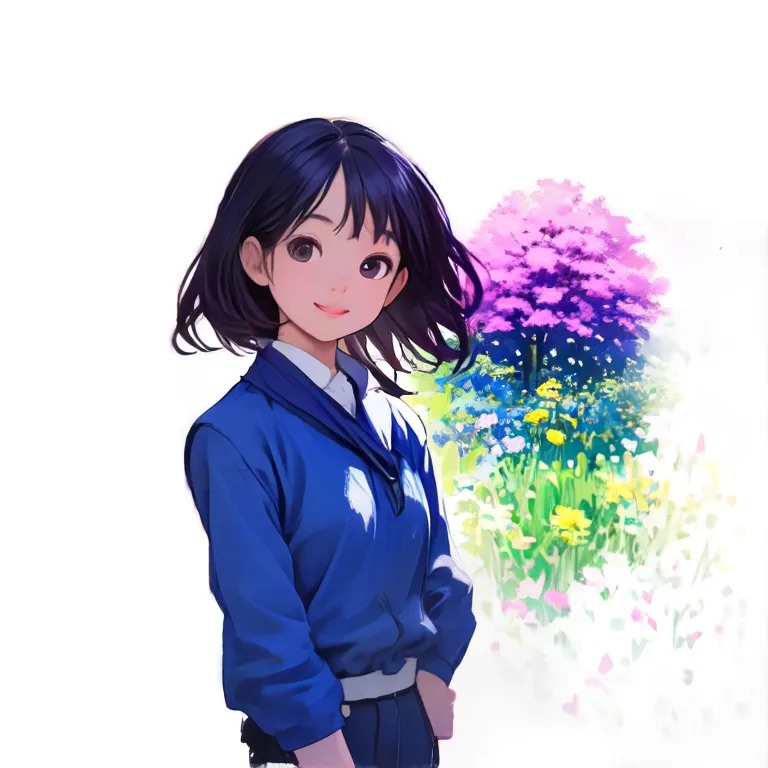} &
    \includegraphics[width=0.14\linewidth,height=0.14\linewidth]{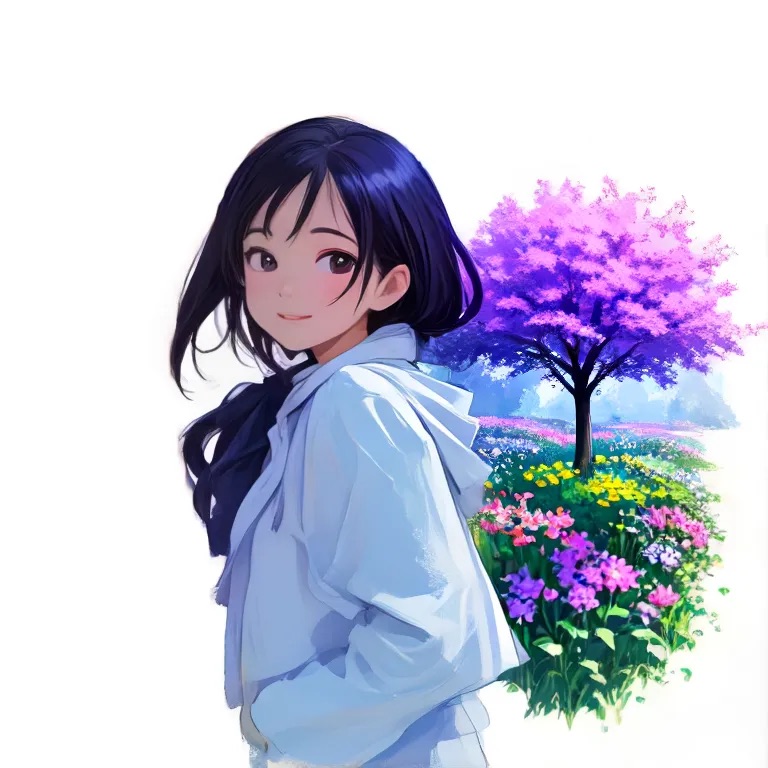} &
    \includegraphics[width=0.14\linewidth,height=0.14\linewidth]{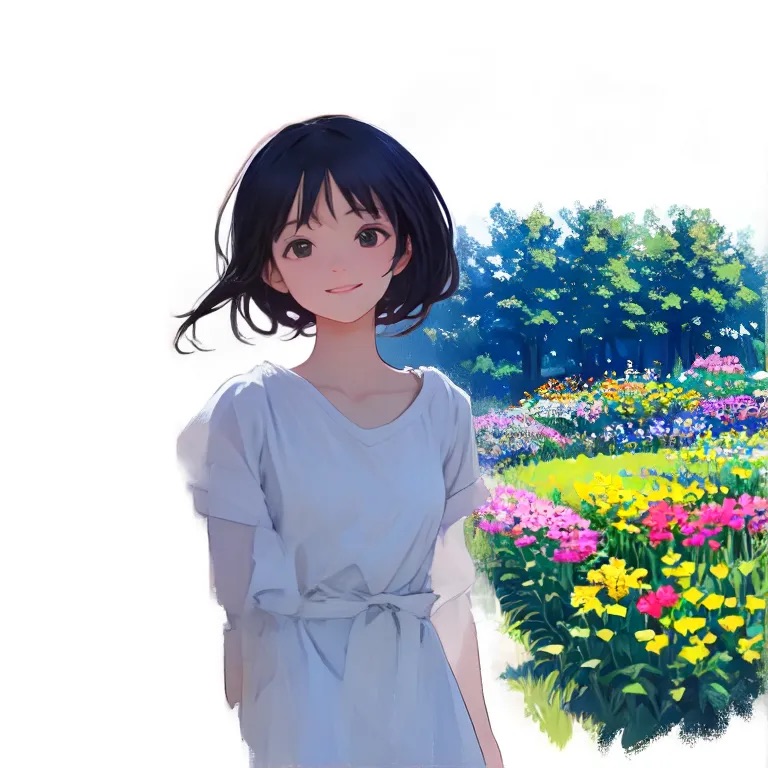} \\
    \includegraphics[width=0.14\linewidth,height=0.14\linewidth]{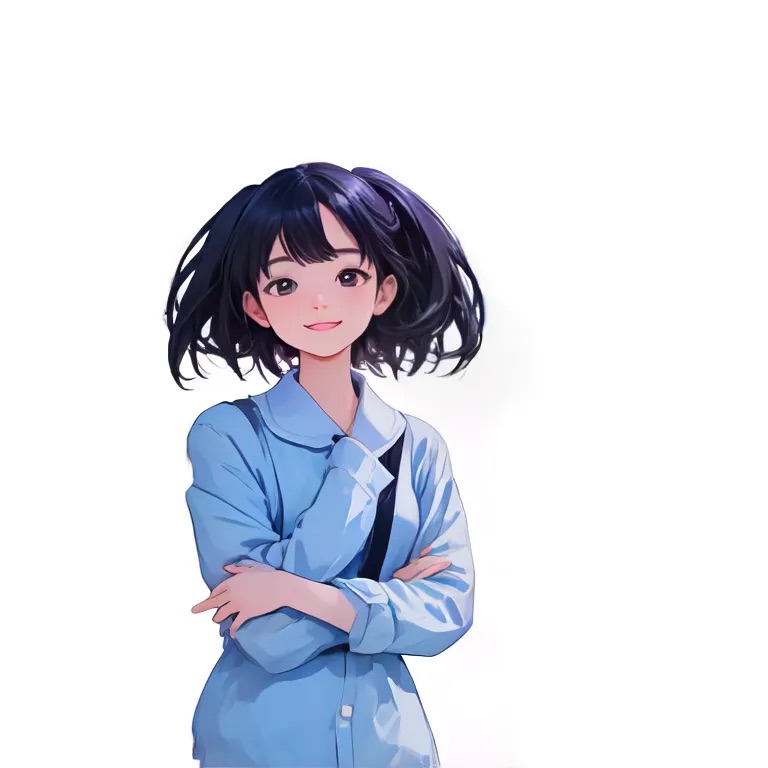} &
    \includegraphics[width=0.14\linewidth,height=0.14\linewidth]{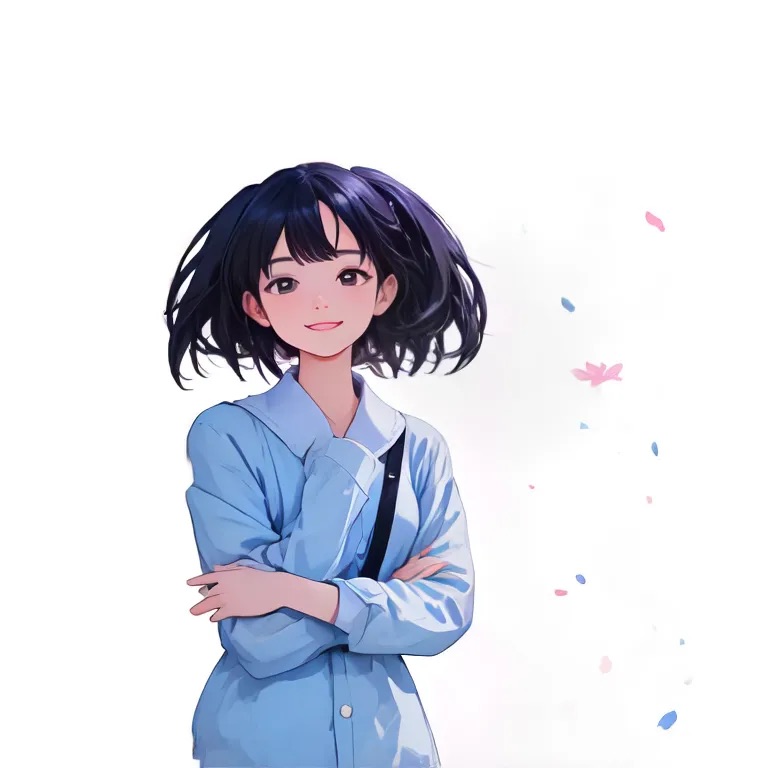} &
    \includegraphics[width=0.14\linewidth,height=0.14\linewidth]{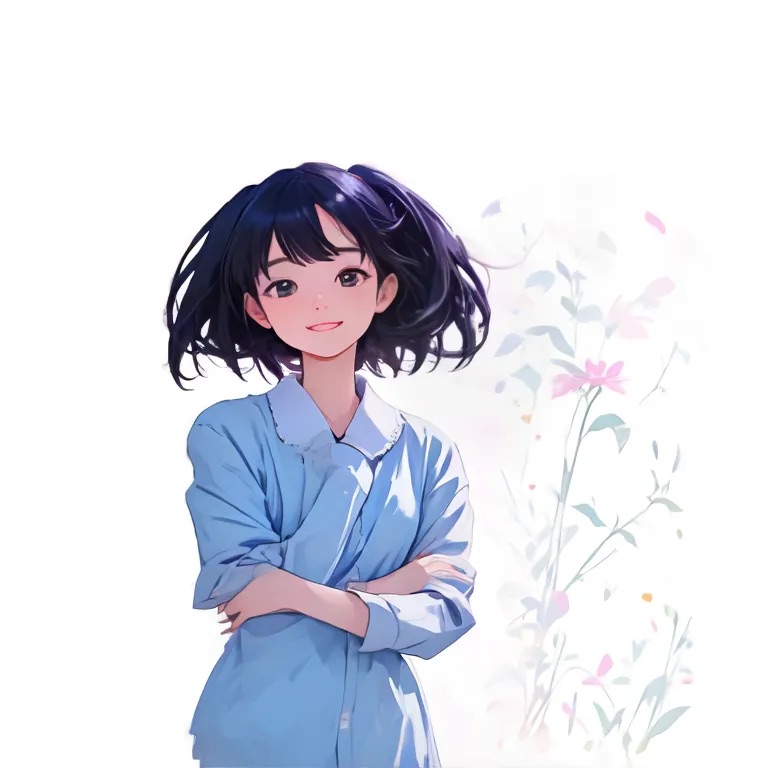} &
    \includegraphics[width=0.14\linewidth,height=0.14\linewidth]{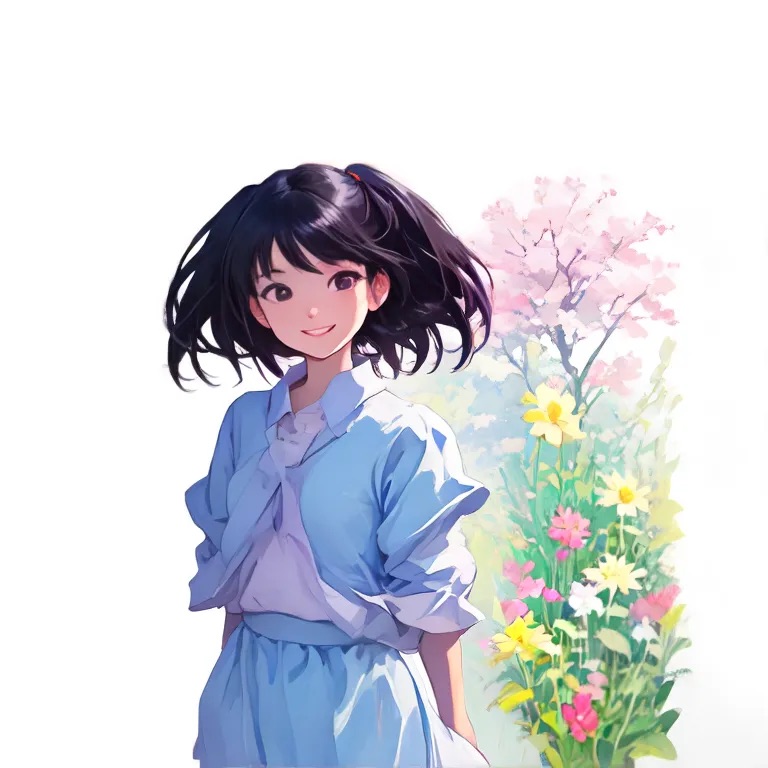} &
    \includegraphics[width=0.14\linewidth,height=0.14\linewidth]{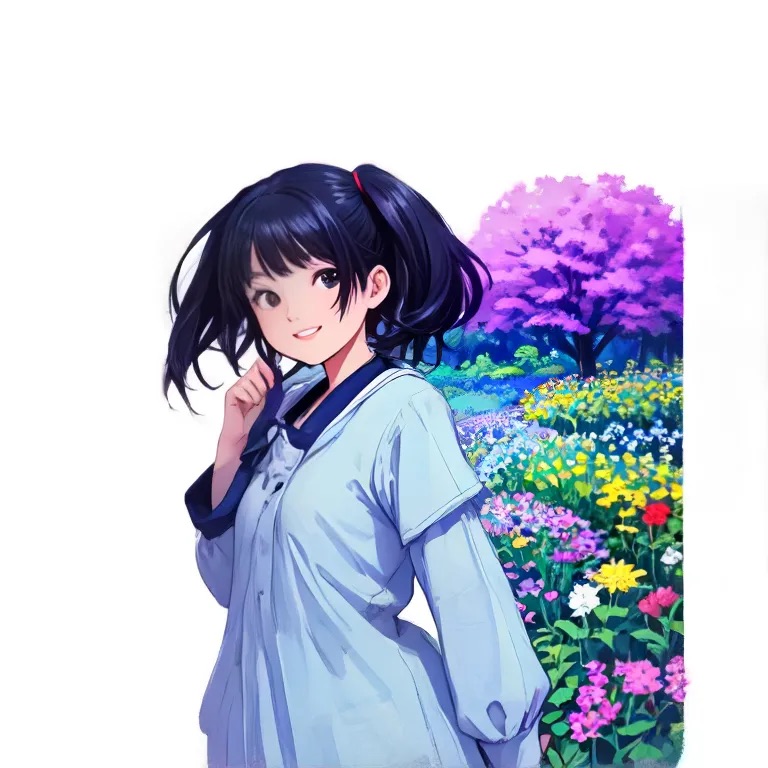} &
    \includegraphics[width=0.14\linewidth,height=0.14\linewidth]{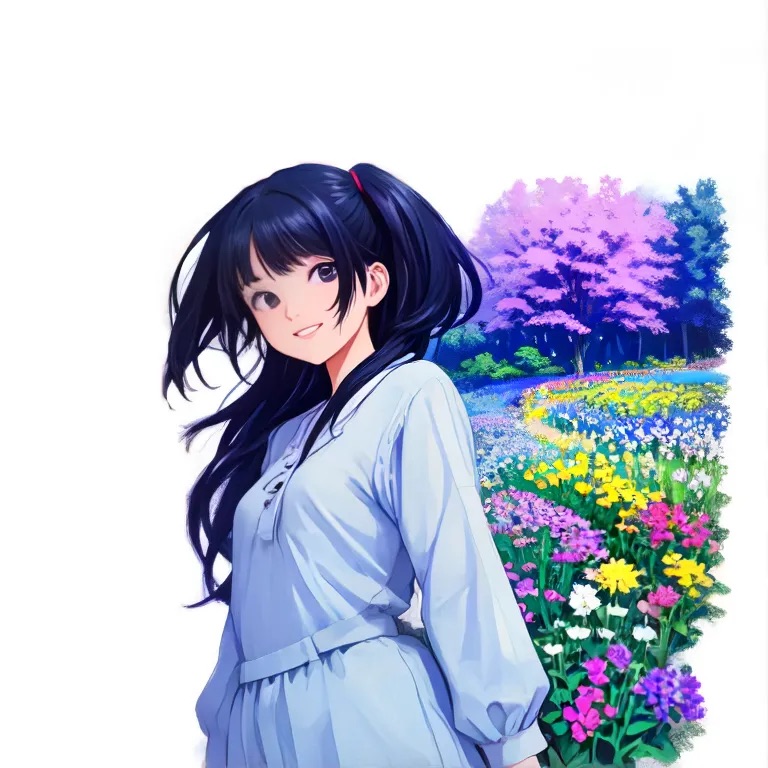} &
    \includegraphics[width=0.14\linewidth,height=0.14\linewidth]{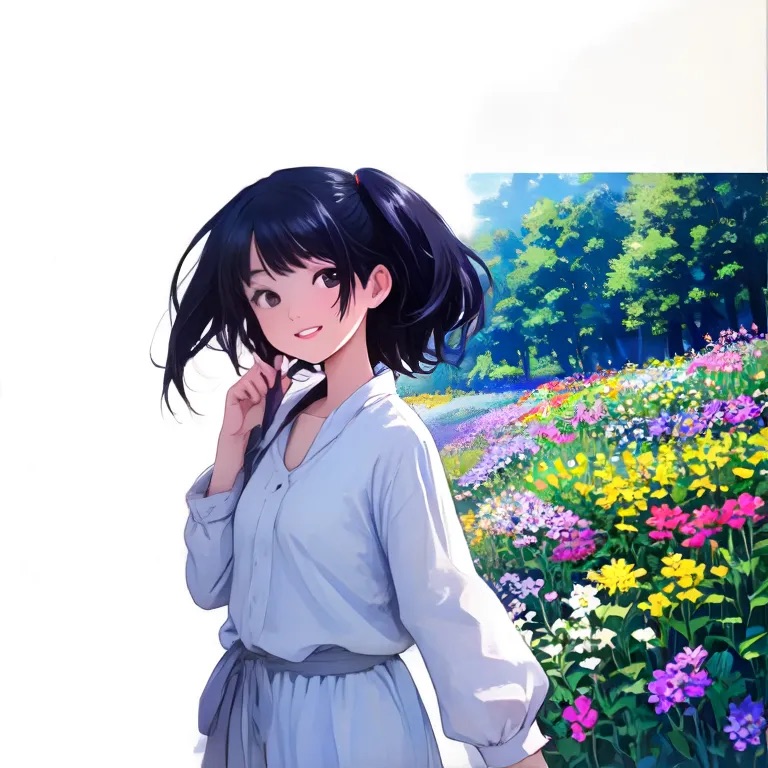}
  \end{tabular}
  \vspace{-1em}
  \caption{
  Sequential generation of frames from real-time drawing of masks.
  Top row: Original without seed-fixing.
  Bottom row: Increased determinism with seed-fixing option.
  A row of images comes sequentially from a single stream of generation given the same sequence of interactive controls (from left to right).
  }
  \label{fig:seed-fixing}
  \vspace{-1.5em}
\end{figure*}
%-------------------------------------------------------------------------

Finally, we provide seed-fixing option that enables incremental generation for drawing-like experience.
The difference between simple streaming generation and streaming generation with our seed-fixing option is elaborated in Figure~\ref{fig:seed-fixing}.
By not only caching the prompt embeddings during streaming but also sharing noise tensors within a stream of generation, we can simply switch into incremental editing in our application.
Therefore, with the seed-fixing option, we can maintain strong consistency across entire stream of generation, which we may call a \textit{session} of content creation.
This enables content creators to switch from random ideation to detailed editing and vice versa, greatly increasing the practicality of the application.
Both options are available in our official code.

\definecolor{importantyellow}{HTML}{F89E12}

\subsection{Basic Usage}
\label{sec:e_app:basic-usage}

We provide the simplest procedure of creating images from \textsc{SemanticDraw} pipeline.
Screenshots in Figure~\ref{fig:appx:demo_howto} illustrate the four-step process.

\paragraph{1. Start the Application.}

After installing the required packages, the user can open the application with the following command prompt:
\begin{lstlisting}[language=bash,firstnumber=1,mathescape=true]
python app.py --model "KBlueLeaf/kohaku-v2.1" --height 512 --width 512
\end{lstlisting}
The application front-end is web-based and can be opened with any web browser through $\texttt{localhost:8000}\,$.
We currently support various baseline architecture including Stable Diffusion 1.5~\cite{rombach2022high}, Stable Diffusion XL~\cite{podellsdxl}, and Stable Diffusion 3~\cite{sauer2024sd3} checkpoints for $\texttt{-{}-model}$ option.
For SD1.5, we support latent consistency model (LCM)~\cite{luo2023latent,luo2023lcm} and Hyper-SD~\cite{ren2024hyper}, for SDXL, we support SDXL-Lightning~\cite{lin2024sdxl}, and for SD3, we support Flash Diffusion~\cite{chadebec2024flash} for acceleration of the generation process.
The height and the width of canvas should be predefined at the startup of the application.

\paragraph{2. Upload Background Image.}

See Figure~\ref{fig:appx:demo_howto:1}.
The first interaction with the application is to upload any image as background by clicking the {\color{importantcolor} background image upload} (no. 4) panel.
The uploaded background image will be resized to match the canvas.
After uploading the image, the background prompt of the uploaded image is automatically generated for the user by pre-trained BLIP-2 model~\cite{li2023blip2}.
The background prompt is used to blend between foreground and background in prompt-level globally, as elaborated in Section~\ref{sec:e_app:ui}.
The interpolation takes place when a foreground text prompt embedding is assigned with a {\color{importantcolor} prompt strength} less than one.
User is always able to change the background prompt like other prompts in the \textit{semantic palette}.

\paragraph{3. Type in Text Prompts.}

See Figure~\ref{fig:appx:demo_howto:2}. 
The next step is to create and manage {semantic brushes} by interacting with the {\color{importantcolor} \textit{semantic palette}} (no. 1).
A minimal required modification is text prompt assignment through the {\color{importantcolor} prompt edit} (no. 8) panel.
The user can additionally modify other options in the control panel marked as {\color{importantyellow} yellow} in Figure~\ref{fig:appx:demo_howto:2}.

\paragraph{4. Draw.}

See Figure~\ref{fig:appx:demo_howto:3}.
A user may start drawing by selecting a brush in {\color{importantcolor} drawing tools} (no. 5) toolbar that matches the user-specified text prompt in the previous step.
Grab a brush, draw, and submit the drawn masks.
After initiating the content creation, the images are streamed through the {\color{importantcolor} display} (no. 6) in real-time from dynamically changing user inputs.
The past generations are saved in {\color{importantcolor} history} (no. 7).

\end{document}